\documentclass[11pt]{article}

\usepackage[a4paper,margin=1in]{geometry}
\usepackage{setspace}
\usepackage[T1]{fontenc}
\usepackage[utf8]{inputenc}
\usepackage{lmodern}
\usepackage{microtype}

\usepackage{amsmath}
\usepackage{amssymb}
\usepackage{amsfonts}
\usepackage{amsthm}
\usepackage{mathrsfs}
\usepackage{bbm}

\usepackage{graphicx}
\graphicspath{{./}{figure/}{figures/}}
\usepackage{float}      
\usepackage{xcolor}

\usepackage{booktabs}
\usepackage{tabularx}
\usepackage{array}
\usepackage{multirow}
\usepackage{makecell}
\usepackage{diagbox}
\usepackage{colortbl}
\newcolumntype{P}[1]{>{\raggedright\arraybackslash}p{#1}}

\usepackage{caption}
\usepackage{longtable}

\usepackage{enumitem}

\usepackage{tikz}
\usetikzlibrary{arrows.meta,calc,positioning}

\usepackage[colorlinks=true,
            linkcolor=blue!50!black,
            citecolor=blue!50!black,
            urlcolor=blue!50!black]{hyperref}
\usepackage[capitalise,nameinlink]{cleveref}
\usepackage{url}

\usepackage[round,sort&compress,authoryear]{natbib}
\bibpunct{(}{)}{;}{a}{,}{,}

\usepackage{titlesec}

\usepackage{authblk}

\providecommand{\keywords}[1]{%
  \par\noindent\textbf{Keywords:}\ #1\par\medskip
}

\title{Bias in Large Language Models: Origin, Evaluation, and Mitigation}

\author[1,*]{Yufei Guo}
\author[1]{Muzhe Guo}
\author[1]{Juntao Su}
\author[1,*]{Zhou Yang}
\author[1]{Mengqiu Zhu}
\author[2]{Hongfei Li}
\author[3]{Mengyang Qiu}
\author[4]{Shuo Shuo Liu}

\affil[1]{Department of Statistics, George Washington University, Washington, DC 20052, USA}
\affil[2]{Department of Statistics, University of Connecticut, Storrs, CT 06269, USA}
\affil[3]{Department of Speech-Language Pathology, Saint Elizabeth University, Morristown, NJ 07960, USA}
\affil[4]{The Pennsylvania State University, University Park, PA 16802, USA}
\affil[*]{Corresponding authors: \textup{\texttt{guoyufei@gwu.edu}} (Y.G.); \textup{\texttt{zhou\_yang@gwu.edu}} (Z.Y.)}

\date{\normalsize Published in \emph{Electronics} \textbf{15}(9), 1824 (2026).
\\
\href{https://doi.org/10.3390/electronics15091824}{https://doi.org/10.3390/electronics15091824}}

\begin{document}

\maketitle

\begin{abstract}
Large language models (LLMs) have revolutionized natural language processing,
but their susceptibility to biases poses significant challenges. This
comprehensive review examines the landscape of bias in LLMs, from its origins
to current mitigation strategies. We categorize biases as intrinsic and
extrinsic, analyzing their manifestations in various natural language
processing (NLP) tasks. The review critically assesses a range of bias
evaluation methods, including data-level, model-level, and output-level
approaches, providing researchers with a robust toolkit for bias detection.
We further explore mitigation strategies, categorizing them into pre-model,
intra-model, and post-model techniques, highlighting their effectiveness and
limitations. Ethical and legal implications of biased LLMs are discussed,
emphasizing potential harms in real-world applications such as healthcare and
criminal justice. By synthesizing current knowledge on bias in LLMs, this
review contributes to the ongoing effort to develop fair and responsible
artificial intelligence (AI) systems. Our work serves as a comprehensive
resource for researchers and practitioners working towards understanding,
evaluating, and mitigating bias in LLMs, fostering the development of more
equitable AI technologies.
\end{abstract}

\keywords{bias evaluation; bias mitigation strategies; large language models}

\tableofcontents

\section{Introduction}
\label{sec1}
The rapid development of \textit{{large language models} 
} (LLMs) has transformed the field of \textit{{natural language processing}} (NLP), introducing new possibilities and applications across various domains, including healthcare \citep{keskar2019ctrl}, finance \citep{yang2020finbert}, education \citep{okonkwo2023role}, and entertainment \citep{jiang2020smart}. These models, such as GPT-3 \citep{brown2020language}, BERT \citep{devlin2018bert}, and others, are designed to understand and generate human-like text by learning from vast corpora of text data. 
These models have become the backbone for tasks such as machine translation \citep{lewis2020bart}, text summarization \citep{nallapati2016abstractive}, sentiment analysis \citep{zhang2018deep}, and automated question answering \citep{raffel2020exploring}, reflecting their deep integration into various industries and day-to-day applications, and making them indispensable tools in the modern AI landscape.

Despite their immense potential and utility, LLMs have raised concerns due to their inherent biases that reflect societal prejudices present in their training data \citep{bender2021dangers, blodgett2020language}. These biases, which manifest as gender, racial, cultural, and socioeconomic stereotypes, pose serious ethical and practical challenges, especially when LLMs are deployed in critical decision-making environments such as healthcare diagnostics \citep{rajkomar2018ensuring}, legal judgments \citep{angwin2016machine}, and hiring processes \citep{chen2018my}. 
Such biases can lead to unequal treatment or skewed results that disproportionately affect marginalized groups, potentially exacerbating existing inequalities. 
\citet{caliskan2017semantics} have noted that LLMs often mirror historical biases present in human language, risking the amplification of these biases within automated systems.

Understanding bias within LLMs requires insight into how biases traditionally arise in statistical models. In classical statistics, biases have been well studied in models for continuous, binary, and time-to-event endpoints \citep{Hastie2009}. These biases manifest due to several issues, such as selection bias \citep{Akaike1974, Heckman1979, Vardi1982}, overfitting \citep{Heckman1979}, handling of covariates \citep{Akaike1974, Tibshirani1996}, limited sample sizes \citep{Nemes2009}, and even the randomization ratio \citep{li2024bias}, which can all lead to incorrect estimates and predictions. These statistical biases share important parallels with those encountered in LLMs. 
For instance, LLMs must handle incomplete or unbalanced datasets carefully to prevent skewed outputs. Similarly, the challenges of selection bias in classical models echo the biases found in LLMs trained on non-representative data sources, which can lead to harmful disparities in model predictions, especially for under-represented groups. 
Addressing these statistical biases requires a range of mitigation strategies, including a robust causal inference framework \cite{Rubin1974}. Similarly, bias mitigation in LLMs involves data-level interventions such as resampling and augmentation, model-level adjustments with fairness constraints, and post-processing corrections to fine-tune outputs \citep{schick2021self}.

LLMs also exhibit biases similar to those observed in traditional statistical models. Bias in LLMs can be broadly categorized into two types: \textit{{intrinsic bias}} and \textit{{extrinsic bias}}. 
\textit{{Intrinsic biases}} originate from the training data, as well as the architecture and underlying assumptions made during model design \citep{sun2019mitigating}. 
LLMs, trained on vast datasets, often sourced from internet and textual repositories, inevitably inherit biases present in these sources. 
For example, language models may reinforce gender stereotypes in occupations, associating ``doctor'' with men and ``nurse'' with women \citep{bolukbasi2016man, zhao2018gender}. 
Similarly, certain demographic groups may be either under-represented or misrepresented within training datasets, further compounding model biases \citep{hovy2021five}. This becomes particularly problematic when LLMs are applied in sensitive domains where accurate and unbiased information is crucial for decision making \citep{bender2021dangers}.

\textit{{Extrinsic biases}}, on the other hand, emerge during the application of LLMs in real-world tasks. These biases are often more subtle as they manifest in the model outputs during specific tasks, such as sentiment analysis, content moderation, or automated decision-making systems. 
For example, \citet{sap2019risk} found that LLMs used in hate speech detection could disproportionately flag certain dialects or vernaculars, such as African-American Vernacular English (AAVE), as more offensive compared to standardized English, reinforcing societal stereotypes. 
Similarly, \citet{kiritchenko2018examining} demonstrated that sentiment analysis systems could produce biased results when analyzing texts associated with different demographic groups, thereby perpetuating harmful stereotypes in their predictions. 

The implications of biases in LLMs are far-reaching. 
In healthcare, for instance, biased models can lead to inappropriate treatment recommendations, potentially worsening existing health inequalities between demographic groups \citep{obermeyer2019dissecting}. 
In legal settings, reliance on biased language models for risk assessment or sentencing decisions can result in discriminatory practices \citep{angwin2016machine}. 
Moreover, biases embedded in LLMs affect everyday applications, such as search engines and social media platforms, where they shape public discourse, potentially reinforcing echo chambers and marginalizing minority voices \citep{pariser2011filter, sap2019social}. 

These biases not only perpetuate inequality but also raise significant ethical and legal questions regarding the use of LLMs in decision-making processes. Given the critical nature of these challenges, there is an urgent need for comprehensive evaluation frameworks to detect, quantify, and mitigate these biases, ensuring that LLMs function as fair and equitable tools for all users. 
\citet{sheng2021societal} provide an overview of these efforts, noting that bias evaluation typically involves analyzing models at various stages of their lifecycle—from data preprocessing to model training and output generation. Common techniques include the use of benchmark datasets designed to highlight biases \citep{nadeem2021stereoset}, probing models to understand their internal representations \citep{hewitt2019structural}, and conducting user studies to assess the real-world impact of biased outputs \citep{sap2019social}.

In terms of bias mitigation, strategies are typically categorized into three broad approaches: data-level interventions, model-level techniques, and post-processing adjustments. 
Data-level interventions involve curating and balancing the training datasets to reduce the prevalence of biased content, often through data augmentation, filtering, or resampling techniques \citep{zhao2018learning}. However, this approach is limited by the vast scale and inherent diversity of LLM training datasets, making it difficult to eliminate biases entirely.
Model-level techniques may include altering the training objectives, incorporating fairness constraints, or modifying model architectures to minimize biased learning \citep{liang2020towards}. 
These methods often trade off some degree of model accuracy for fairness, presenting challenges in real-world deployment. 
Post-processing methods focus on adjusting the outputs of LLMs after generation, applying debiasing algorithms, controlled text generation, or reinforcement learning frameworks to reduce biased content \citep{schick2021self}. While this approach offers flexibility, it may require substantial computational overhead and may not fully address the root causes of bias.

Despite these advancements, fully addressing bias in LLMs remains a complex and ongoing challenge. The complexity of language, the vastness of training datasets, and the massive architectures of modern LLMs---often comprising billions of parameters---make it difficult to identify and correct all instances of bias. 
Furthermore, as~\citet{mehrabi2021survey} suggest, biases are often context-dependent and may evolve over time as societal norms and values change. 
This dynamic nature of bias necessitates continuous monitoring, evaluation, and refinement of both models and the methodologies used to assess their~outputs.

To boost the reproducibility of this review and allow readers to audit the review's coverage, the literature collection process is documented as follows. Specifically, we searched ACL Anthology, arXiv (cs.CL, cs.AI, cs.LG), Google Scholar, Semantic Scholar, and ACM Digital Library, covering major NLP and AI venues, including ACL, EMNLP, NAACL, NeurIPS, ICML, AAAI, and FAccT. Primary query strings included combinations of (``large language model'' OR ``LLM'' OR ``language model'') AND (``bias'' OR ``fairness'' OR ``stereotype'' OR ``debiasing''). The search covered publications from January 2016 through August 2024. Inclusion criteria required papers to directly address bias origins, evaluation, or mitigation in language models, be published in peer-reviewed venues or as substantive preprints, and be written in English. Approximately 200 papers were included in the final synthesis after title, abstract, and full-text screening.

\begin{figure}[H]
    \includegraphics[width=0.98\textwidth]{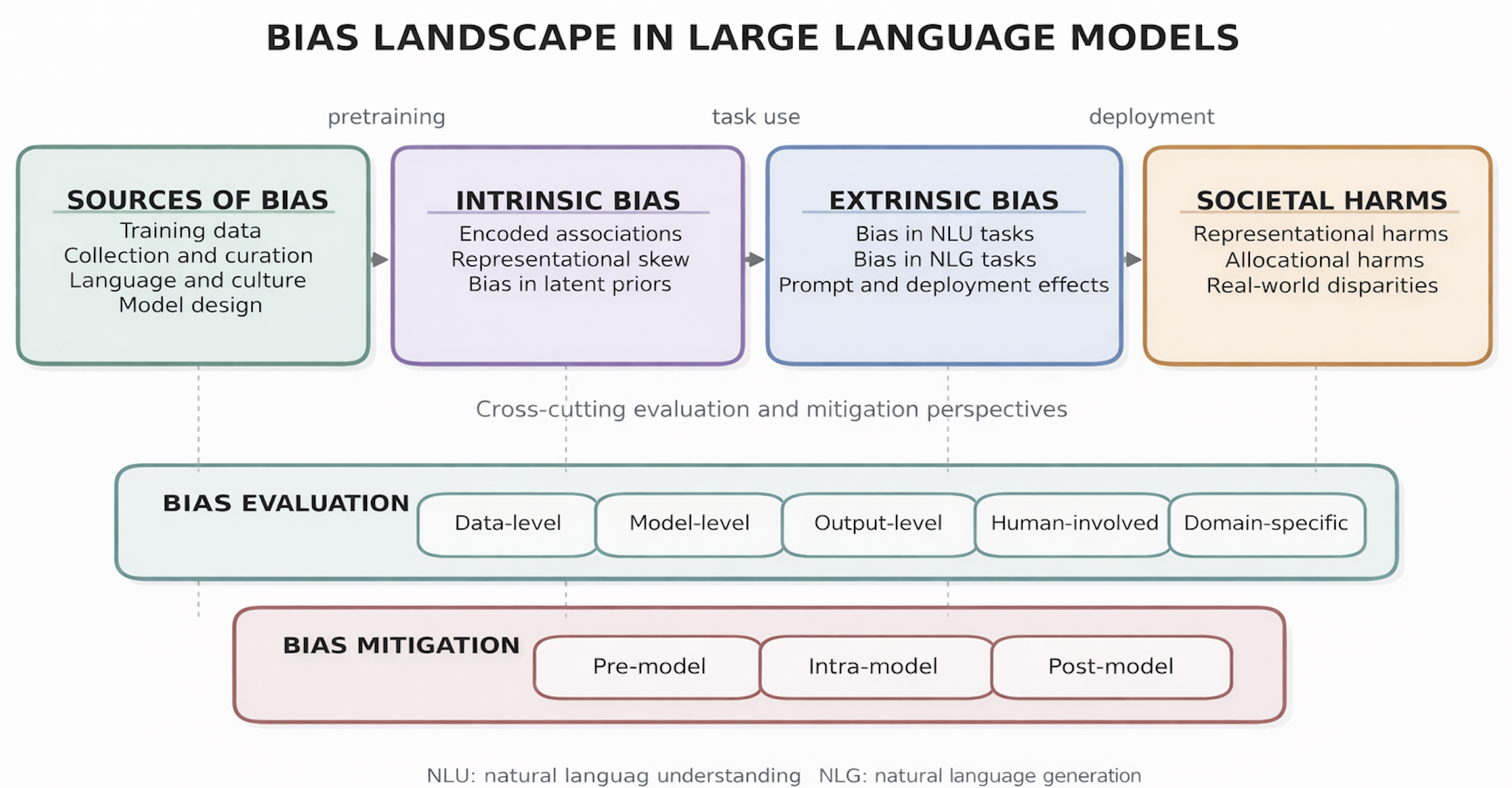}
  \caption{{Overview} of the bias landscape in large language models.}
 \label{fig:1p1_landscape}
\end{figure}

This article provides a structured synthesis of current research on bias in LLMs, covering its origins, manifestations, evaluation, mitigation, and broader societal implications. While prior reviews such as Mehrabi et al.~\cite{mehrabi2021survey}, Navigli et al.~\cite{navigli2023biases}, and Gallegos et al.~\cite{gallegos2024bias} have made important contributions to bias phenomena and mitigation methods, the present review distinguishes itself through five specific contributions. First, we propose and operationalize a precise two-level taxonomic framework distinguishing \textit{{intrinsic bias}}, which is encoded within model representations during pretraining, from \textit{{extrinsic bias}}, which emerges in downstream tasks and generated outputs, providing clearer conceptual boundaries than existing categorizations. Second, we review bias evaluation methods across data-, model-, and output-level analyses, while also discussing human-involved and domain-specific evaluation perspectives. Third, we provide a comparative synthesis of pre-model, intra-model, and post-model mitigation strategies, explicitly examining trade-offs among computational cost, generalizability, and fairness that have received less attention in prior surveys. Fourth, we examine the ethical and legal implications of biased LLMs through a structured taxonomy of representational and allocational harms, grounding abstract fairness concerns in documented real-world consequences. Fifth, we highlight cross-cultural and multilingual gaps that remain under-represented in the current literature, documenting bias phenomena and evaluation gaps across East Asian, Arabic, South Asian, African, and Latin American linguistic and cultural contexts. Figure~\ref{fig:1p1_landscape} provides a high-level overview of the bias landscape in LLMs considered in this review, linking upstream sources of bias to intrinsic bias, extrinsic bias in downstream tasks, cross-cutting evaluation and mitigation perspectives, and broader societal harms.

The remainder of this paper is organized as follows. Sections \ref{sec2} and \ref{sec3} discuss the sources and types of intrinsic and extrinsic bias, drawing on examples from various NLP tasks to illustrate how biases arise and persist. Section \ref{sec4} reviews the methodologies used to detect and measure biases, including both quantitative and qualitative approaches. Section \ref{sec5} compares mitigation strategies across different stages of the model development pipeline, assessing their effectiveness and limitations. Section \ref{sec6} discusses the ethical considerations and legal implications of deploying biased AI systems, underscoring the importance of fairness, transparency, and accountability in AI development. Through these discussions, we aim to contribute to the broader discourse on responsible AI by offering insights and recommendations for researchers, practitioners, and policymakers working to ensure that LLMs serve as equitable and trustworthy tools in society.

\section{Intrinsic {Bias} 
 \label{sec2}}

The bias presented in large language models (LLMs) can be broadly categorized into intrinsic bias and extrinsic bias based on the different stages at which the biases manifest within the model's lifecycle and the type of bias being measured~\citep{doan2024fairness}. Intrinsic bias refers to biases that are inherently within the internal representations or outputs of a trained or pretrained LLM and are independent of any specific downstream tasks. Extrinsic bias refers to the biases that manifest during the model's performance on specific downstream tasks after training or fine-tuning. Intrinsic biases are typically encoded during the training or pretraining phase when the LLM learns from large-scale corpora.
For instance, if a pretrained LLM consistently associates certain professions with specific genders (e.g., assuming all doctors to be male and assuming all nurses to be female) or racial stereotypes~\citep{may2019measuring}, this bias would be considered intrinsic bias. This type of bias is tied to the way the model has encoded relationships and patterns in its internal representations. In this section, we introduce the possible sources of intrinsic bias and present the manifestations of intrinsic bias.

Intrinsic biases are primarily introduced during the training/pretraining stages, where models learn patterns and representations from vast corpora. These biases can be traced to several aspects of data and model design, which are outlined below.

\subsection{Bias in Training Data}
Language models, especially LLMs that are trained on a large corpus, usually have intrinsic bias issues, since the training corpora often contain societal biases that are built into the model~\citep{pagano2023bias, ray2023chatgpt, goldfarb2024fairness, goldman2024statistical}. These biases can arise in several ways, as described below. 
\begin{itemize}
    \item Bias from over-representativeness and under-representativeness: Certain groups, such as gender, age,  race, religion, ethnicity, culture, political, and socioeconomic classes, may be under-represented or over-represented in the corpus~\citep{das2024under, alvero2024large}. For example, men might be over-represented in datasets about leadership or science, while women may be more frequently mentioned in caregiving roles~\citep{into2024systematic}, thus leading to biased associations with demographic information. Another example is that the Latinx population, especially Mexican-American students, remains under-represented in U.S. higher education~\citep{alvero2024large}. 
    \item Spatial and temporal bias: LLMs trained predominantly on a corpus from certain countries or geographic locations may absorb the cultural norms and values, hence building biases into the underlying LLMs. For instance, a model trained on Western-centric data may have a skewed understanding of non-Western cultures~\citep{ahmadbias, talat2022you, mukherjee2024global, lee2024neural, shi2024culturebank}. Prior studies have shown such effects in Arabic contexts, where models may prefer Western-associated entities over Arab ones~\cite{naous2024having}, and in broader multilingual settings, where performance and cultural knowledge vary substantially across under-represented languages and regions~\cite{wang2024not}. Similarly, data collected over different periods may reflect outdated societal norms and values. LLMs trained on these data may be biased by those outdated norms and values. For instance, historical texts may exhibit racist or sexist terminology, which the model may absorb as part of its internal representations~\citep{salinas2023not, bussaja2024evaluating}.
\end{itemize}
\subsection{Bias in Data Collection Methods}
The intrinsic biases can also stem from the methods that are used to collect the training corpus. A large training corpus of LLMs often comprises sub-datasets that are collected from different sources, including web scraping, social media and forums, books and literature, code repositories, scientific and academic papers, legal and finance documents, dialogue and transcriptions, government and public records, custom and proprietary databases, and model-generated datasets. These different types of sources have their own biases. For instance, internet-based datasets can include unmoderated, biased, or hateful content. Academic papers tend to be more neutral, but they often lack the diversity of language found in social media or everyday conversations~\citep{gallegos2024bias, kuntz2023authors, qu2024performance}.
Choosing which texts to include or exclude during the data collection phase can also lead to biases~\citep{navigli2023biases}. For instance, excluding certain groups or viewpoints, intentionally or unintentionally, can introduce selection bias, which can skew the model’s outputs toward the dominant narrative in the training data.
Even with careful curation by LLM creators, biases can still be embedded in the model’s learning, as it is nearly impossible to fully eliminate inappropriate content given the vast scale of the training corpus.

\subsection{Bias in Language Contexts}
Biases can also stem from the context of language. Human languages often contain ambiguities (including lexical ambiguity, syntax ambiguity, semantic ambiguity, and context-dependent meaning) due to the flexible nature of words, syntax, and context, which can lead to model biases. For example, gender-neutral pronouns may  be associated with one gender due to the patterns in the training dataset~\citep{kotek2023gender, dwivedi2023breaking}. Certain linguistic constructions, such as figures of speech and rhetoric, add another layer of complexity and ambiguity to human language by using words and expressions in non-literal or persuasive ways and hence introduce biases~\citep{de2024semantic}. 

\subsection{Bias in Tokenization and Word Embeddings}

Tokenization and the use of word embeddings can further introduce biases into LMs. Some tokenization techniques, such as WordPiece or Byte-Pair Encoding (BPE), may disproportionately split less frequently occurring words into smaller units. This splitting policy can result in fragmented representations of under-represented entities, names, or terminology, particularly affecting minority languages or groups. For instance, Petrov et~al. show that biases can arise at the tokenization stage and that the same content can require substantially different token lengths across languages~\cite{petrov2023language}. Ahia et al. demonstrate large cross-lingual token-count disparities with practical fairness consequences~\cite{ahia2023all}. Ovalle et al.~\cite{ovalle2024tokenization} discover that Byte-Pair Encoding (BPE) tokenization, the tokenizer powering many popular LLMs, influences LLM misgendering significantly. Similarly, biases can also arise from word embedding, where many language models are initialized with pretrained word embeddings~\cite{bolukbasi2016man, caliskan2017semantics, garg2018word, nadeem2021stereoset, rakshit2025prejudice}. 

\section{Extrinsic Bias\label{sec3}}

Extrinsic bias refers to disparities in a model's performance across different downstream tasks, also known as downstream bias or prediction bias \citep{doan2024fairness, gallegos2024bias, li2023survey}. This type of bias emerges when a model's effectiveness varies among tasks or demographic groups, potentially leading to unequal outcomes in practical applications.

Depending on the downstream tasks, extrinsic bias manifests differently. We categorize these tasks into two main groups: natural language understanding (NLU) tasks and natural language generation (NLG) tasks. In NLU tasks, extrinsic bias affects how the model comprehends and interprets input text. In NLG tasks, it can result in the generation of biased or stereotypical language.

\subsection{Natural Language Understanding (NLU) Tasks}

NLU tasks aim to improve a model's comprehension of input sequences beyond the literal meaning of words \citep{chang2024survey}. Extrinsic bias in these tasks can cause the model to misinterpret or unfairly process input text based on biased associations learned during~training.

Common manifestations of bias in NLU tasks include the following:

\begin{itemize}
    \item \textbf{{Gender bias:} 
} Models may incorrectly associate certain professions or roles with a specific gender, leading to errors in tasks like coreference resolution \citep{zhao2018gender}. For example, assuming a doctor is male and a nurse is female, regardless of context.
    \item \textbf{{Age bias:}} Models might make assumptions about individuals based on age stereotypes~\citep{caliskan2017semantics}. For instance, associating technological proficiency only with younger people.
    \item \textbf{{Cultural or regional bias:}} Models could misinterpret idioms or expressions from different cultures or fail to recognize regional language variants \citep{vulic2013cross}. This can result in misunderstandings in tasks like semantic textual similarity or natural language inference.
\end{itemize}

These biases can lead to unfair or inaccurate outcomes in various NLU tasks, such as coreference resolution, semantic textual similarity, natural language inference, classification, reading comprehension, and sentiment analysis. Recurrent issues include reinforcing stereotypes, misclassifying text due to dialects or language variants, and making incorrect inferences based on biased associations.

\subsection{Natural Language Generation (NLG) Tasks}

NLG tasks involve generating coherent and contextually relevant text based on input or instructions \citep{chang2024survey}. Extrinsic bias in these tasks can cause models to produce biased language or reinforce stereotypes in generated text.

Common manifestations of bias in NLG tasks include the following:

\begin{itemize}
    \item \textbf{{Gender bias:}} Models might generate responses that align with gender stereotypes, such as using male pronouns for leaders and female pronouns for nurturing roles \citep{zhao2018gender}.
    \item \textbf{{Age bias:}} Models may produce content that reflects age-related stereotypes, like suggesting only sedentary activities for older adults \citep{caliskan2017semantics}.
    \item \textbf{{Cultural or regional bias:}} Models could favor content from dominant cultures or misrepresent cultural practices, leading to inappropriate or insensitive responses \citep{hendricks2018women}.
\end{itemize}

These biases affect tasks like question answering, sentence completion, conversational agents, recommender systems, machine translation, and summarization. Recurrent issues include reinforcing harmful stereotypes, under-representing minority cultures, and providing biased recommendations or translations.

In general, extrinsic biases in LLMs impact both understanding and generation tasks, leading to unfair or discriminatory outcomes. Addressing these biases is essential to ensure models perform fairly and accurately in real-world applications. An overview of these biases in NLU tasks is provided in Table~\ref{tab:nlu-biases}, and in NLG tasks in Table~\ref{tab:nlg-biases}. Detailed examples can be found in Appendix~\ref{appd_biases}.

{\footnotesize
\renewcommand{\arraystretch}{1.15}
\begin{longtable}{
  >{\raggedright\arraybackslash}p{0.16\textwidth}
  >{\raggedright\arraybackslash}p{0.24\textwidth}
  >{\raggedright\arraybackslash}p{0.24\textwidth}
  >{\raggedright\arraybackslash}p{0.24\textwidth}}
\caption{Overview of extrinsic biases in natural language understanding (NLU) tasks. Each row corresponds to an NLU task, and each column summarizes representative examples of gender, age, and cultural or regional bias. The examples highlight typical patterns rather than exhaustive cases. This table can be used to compare how different types of bias manifest across tasks.}
\label{tab:nlu-biases}\\
\toprule
\textbf{NLU Task} & \textbf{Gender Bias} & \textbf{Age Bias} & \textbf{Cultural or Regional Bias} \\
\midrule
\endfirsthead

\multicolumn{4}{c}{\tablename\ \thetable\ -- \textit{continued from previous page}} \\
\toprule
\textbf{NLU Task} & \textbf{Gender Bias} & \textbf{Age Bias} & \textbf{Cultural or Regional Bias} \\
\midrule
\endhead

\midrule
\multicolumn{4}{r}{\textit{continued on next page}} \\
\endfoot

\bottomrule
\endlastfoot

\textbf{Coreference resolution}
\newline
{Identifying instances where different expressions refer to the same entity \citep{lee2017end}} & \textbf{(1) Stereotypical occupation associations:} Mislinking pronouns based on gender stereotypes in professions, e.g., doctor → male \citep{zhao2018gender}
\newline
\textbf{(2) Gendered pronoun resolution:} Incorrectly resolving pronouns for gender-neutral names, reflecting underlying gender biases \citep{rudinger2018gender} & \textbf{(1) Assumptions about technological proficiency:} Attributing technological skill to certain ages during pronoun resolution \citep{hovy2021importance}
\newline
\textbf{(2) Bias in linking pronouns to age-related roles:} Misassigning actions to younger or older individuals based on stereotypes \citep{garg2018word} & \textbf{(1) Regional variants and pronoun use:} Difficulty resolving pronouns in pro-drop languages due to training on non-pro-drop data \citep{hovy2021importance}
\newline
\textbf{(2) Cultural context in family roles:} Misresolving pronouns based on cultural stereotypes about family responsibilities \citep{rudinger2018gender} \\
\midrule
\textbf{Semantic textual similarity}
\newline
{Evaluating how similar the meanings of two texts are \citep{cer2017semeval}} & \textbf{(1) Gendered language and pronoun resolution:} Lower similarity scores for sentences differing only by gendered pronouns \citep{zhao2017men}
\newline
\textbf{(2) Gender-stereotyped professions:} Overestimating similarity when sentences align with gender stereotypes \citep{bolukbasi2016man} & \textbf{(1) Age-based expectations in language use:} Bias in similarity assessments due to age-related stereotypes \citep{caliskan2017semantics}
\newline
\textbf{(2) Age stereotypes in sentiment and perception:} Overestimating similarity based on stereotypes about age and wisdom \citep{diaz2016query} & \textbf{(1) Cultural idioms and expressions:} Underestimating similarity between culturally equivalent idioms \citep{vulic2013cross}
\newline
\textbf{(2) Regional dialects and variants:} Misjudging similarity due to differences in regional vocabulary \citep{tan2011building} \\
\midrule
\textbf{Natural language inference}
\newline
{Determining the relationship between a premise and a hypothesis \citep{bowman2015large}} & \textbf{(1) Stereotypical gender roles in professions:} Incorrect inferences based on gender stereotypes in job roles \citep{rudinger2018gender}
\newline
\textbf{(2) Gendered language and assumptions:} Bias in inference due to gender assumptions with neutral names \citep{zhao2017men} & \textbf{(1) Bias toward younger individuals in dynamic roles:} Overemphasizing youth in roles like entrepreneurship \citep{caliskan2017semantics}
\newline
\textbf{(2) Age-related stereotypes in professional contexts:} Assuming leadership roles are held by older individuals \citep{zhao2018gender} & \textbf{(1) Cultural norms and social roles:} Inferences influenced by cultural stereotypes about professions \citep{rudinger2018gender}
\newline
\textbf{(2) Regional political and historical contexts:} Difficulty with inferences requiring diverse historical knowledge \citep{chen2016enhanced} \\
\midrule
\textbf{Classification}
\newline
{Assigning categories or labels to input text \citep{de2019bias, meng2020text}} & \textbf{(1) Gendered language in job title classification:} Associating certain professions with specific genders \citep{bolukbasi2016man}
\newline
\textbf{(2) Gendered pronouns and name classification:} Misclassifying names or pronouns based on gender stereotypes \citep{zhao2017men} & \textbf{(1) Sentiment classification bias based on age:} Misclassifying sentiment due to age-related stereotypes \citep{diaz2018addressing}
\newline
\textbf{(2) Bias in job application classification:} Discriminating against older applicants in job screening \citep{de2019bias} & \textbf{(1) Misclassification due to cultural language variants:} Incorrectly classifying non-standard dialects as negative \citep{sap2019risk}
\newline
\textbf{(2) Regional bias in political text classification:} Bias toward dominant regional political ideologies \citep{bender2021dangers} \\
\midrule
\textbf{Reading comprehension}
\newline
{Answering questions based on a given text \citep{rajpurkar2018know}} & \textbf{(1) Stereotypes in gendered activities:} Assuming gender based on activities, e.g., cooking → female \citep{webster2018mind}
\newline
\textbf{(2) Assumptions about gender roles in family settings:} Inferring roles based on traditional gender norms \citep{caliskan2017semantics} & \textbf{(1) Bias in health-related content:} Associating certain health issues with age stereotypes \citep{bolukbasi2016man}
\newline
\textbf{(2) Misinterpretation of age-related roles:} Incorrectly assuming roles based on age \citep{bender2018data} & \textbf{(1) Cultural context misinterpretation:} Misunderstanding culturally specific practices in texts \citep{bender2018data}
\newline
\textbf{(2) Regional language varieties and dialects:} Struggling with comprehension of non-standard dialects \citep{blodgett2020language} \\
\midrule
\textbf{Sentiment analysis}
\newline
{Identifying the emotional tone in text \citep{pang2008opinion}} & \textbf{(1) Bias in sentiment toward gendered products or topics:} Skewed sentiment analysis for gender-specific items \citep{zhao2017men}
\newline
\textbf{(2) Sentiment analysis in gendered contexts:} Misclassifying sentiment in discussions challenging gender norms \citep{misiunas2019density} & \textbf{(1) Sentiment analysis on age-related topics:} Assuming negative sentiment in texts by older individuals \citep{burnap2015cyber}
\newline
\textbf{(2) Stereotyping language use by older adults:} Misinterpreting sentiment due to age-related language stereotypes \citep{hovy2015tagging} & \textbf{(1) Cultural bias in sentiment toward social norms:} Misclassifying sentiment across different cultural norms \citep{blodgett2017racial}
\newline
\textbf{(2) Sentiment in multilingual contexts:} Incorrect sentiment assessment in code-switching texts \citep{hovy2015tagging} \\
\end{longtable}}

{\footnotesize
\renewcommand{\arraystretch}{1.15}
\begin{longtable}{
  >{\raggedright\arraybackslash}p{0.16\textwidth}
  >{\raggedright\arraybackslash}p{0.24\textwidth}
  >{\raggedright\arraybackslash}p{0.24\textwidth}
  >{\raggedright\arraybackslash}p{0.24\textwidth}}
\caption{Overview of extrinsic biases in natural language generation (NLG) tasks. Each row corresponds to an NLG task, while columns summarize representative examples of gender, age, and cultural or regional bias. These examples illustrate common bias patterns and are not exhaustive. Table can be used to compare how different types of bias manifest across tasks.}
\label{tab:nlg-biases}\\
\toprule
\textbf{NLG Task} & \textbf{Gender Bias} & \textbf{Age Bias} & \textbf{Cultural or Regional Bias} \\
\midrule
\endfirsthead

\multicolumn{4}{c}{\tablename\ \thetable\ -- \textit{continued from previous page}} \\
\toprule
\textbf{NLG Task} & \textbf{Gender Bias} & \textbf{Age Bias} & \textbf{Cultural or Regional Bias} \\
\midrule
\endhead

\midrule
\multicolumn{4}{r}{\textit{continued on next page}} \\
\endfoot

\bottomrule
\endlastfoot

\textbf{Question answering}
\newline
{Providing accurate answers based on a given text or knowledge base \citep{rajpurkar2016squad}} & \textbf{(1) Gender bias in answer generation:} When asked, ``What should a good leader do?'', the model might use stereotypically male attributes, implying leadership qualities are inherently male \citep{zhao2018gender}
\newline
\textbf{(2) Bias in answering ambiguous gender questions:} For questions without specified gender, the model might default to male pronouns, reinforcing gender assumptions \citep{webster2018mind} & \textbf{(1) Stereotypical answers about aging:} When asked about activities for elderly people, the model might focus on sedentary activities, neglecting active pursuits \citep{caliskan2017semantics}
\newline
\textbf{(2) Negative bias toward youth:} Suggesting that young people lack experience for roles like managing a company \citep{bolukbasi2016man} & \textbf{(1) Bias in answering culturally specific questions:} Providing answers based on regional popularity rather than global knowledge, e.g., stating ``American football'' as the most popular sport \citep{gardner2018allennlp}
\newline
\textbf{(2) Language and regional bias in answer accuracy:} Better performance on questions related to well-represented regions and languages \citep{kwiatkowski2019natural} \\
\midrule
\textbf{Sentence completion}
\newline
{Predicting and generating the next word or sequence to complete a sentence \citep{radford2019language}} & \textbf{(1) Gender bias in descriptions of physical appearance:} Completing sentences in ways that reinforce stereotypes about women's and men's concerns \citep{caliskan2017semantics}
\newline
\textbf{(2) Gender bias in personal attributes:} Associating women with ``emotional'' and men with ``strong'' in sentence completions \citep{zhao2017men} & \textbf{(1) Activity and lifestyle assumptions:} Completing sentences with stereotypical activities based on age \citep{garg2018word}
\newline
\textbf{(2) Learning and education stereotypes:} Assuming older adults are ``catching up'' on education \citep{sun2019mitigating} & \textbf{(1) Cultural stereotyping in sentence completion:} Overemphasizing specific aspects of a culture, e.g., ``In Japan, people often eat sushi'' \citep{hendricks2018women}
\newline
\textbf{(2) Regional bias in place-based completions:} Reinforcing stereotypes about regions, e.g., ``In Africa, many people live in villages'' \citep{birhane2021large} \\
\midrule
\textbf{Conversational}
\newline
{Generating coherent dialogue and maintaining conversation context \citep{zhang2019dialogpt}} & \textbf{(1) Stereotypical responses based on gendered prompts:} Reinforcing gender stereotypes in descriptions, e.g., nurses as female and caring \citep{bolukbasi2016man}
\newline
\textbf{(2) Bias in gendered interactions:} Responding differently based on assumed user gender \citep{hitsch2010matching} & \textbf{(1) Assumptions about being tech-savvy:} Assuming tech-savvy users are younger and providing simplistic explanations to older users \citep{hovy2015tagging}
\newline
\textbf{(2) Bias in addressing age-related concerns:} Discouraging older individuals from pursuing new careers \citep{wagner2016women} & \textbf{(1) Culturally inappropriate responses:} Failing to understand cultural norms in responses \citep{hovy2016social}
\newline
\textbf{(2) Bias in handling regional topics:} Focusing on negative topics for certain regions \citep{gebru2021datasheets} \\
\midrule
\textbf{Recommender systems}
\newline
{Suggesting personalized content or items based on user data \citep{zhang2019deep}} & \textbf{(1) Product recommendations based on gender stereotypes:} Suggesting products reinforcing traditional gender roles \citep{ekstrand2018all}
\newline
\textbf{(2) Career and education recommendations:} Suggesting STEM careers more to men and arts to women \citep{chen2019correcting} & \textbf{(1) Age-related product recommendations:} Recommending products based on age stereotypes \citep{biega2018equity}
\newline
\textbf{(2) Media and entertainment recommendations:} Assuming preferences based on age, limiting content diversity \citep{burke2018balanced} & \textbf{(1) Regional bias in news and information recommendations:} Under-representing news from minority regions \citep{karimi2018news}
\newline
\textbf{(2) Bias in language and cultural content recommendations:} Prioritizing content in dominant languages \citep{lakew2018multilingual} \\
\midrule
\textbf{Machine translation}
\newline
{Translating text from one language to another \citep{vaswani2017attention}} & \textbf{(1) Gendered language mismatch:} Introducing gender bias when translating from gender-neutral to gendered languages \citep{prates2020assessing}
\newline
\textbf{(2) Gender stereotyping in occupational translations:} Assigning gendered pronouns based on stereotypes \citep{neri2020design} & \textbf{(1) Bias in addressing older adults:} Translations that condescend to older adults \citep{caliskan2017semantics}
\newline
\textbf{(2) Translation of age-related idioms:} Reinforcing negative stereotypes in translations \citep{hovy2015tagging} & \textbf{(1) Cultural nuance loss:} Mistranslating idioms and expressions without cultural context \citep{vanmassenhove2019lost}
\newline
\textbf{(2) Bias toward dominant cultures:} Favoring translations aligning with dominant (e.g., Western) norms \citep{sennrich2016neural} \\
\midrule
\textbf{Summarization}
\newline
{Generating concise summaries of longer texts \citep{nallapati2016abstractive}} & \textbf{(1) Differential emphasis on roles:} Emphasizing traditional gender roles in summaries \citep{bender2018data}
\newline
\textbf{(2) Selective emphasis on gendered information:} Overemphasizing gender-specific details not central to the story \citep{otterbacher2017competent} & \textbf{(1) Bias in summarizing content for different age groups:} Simplifying content in a condescending way \citep{obermeyer2019dissecting}
\newline
\textbf{(2) Omission of contributions based on age:} Highlighting contributions of younger people over older individuals \citep{bender2018data} & \textbf{(1) Omission of culturally significant details:} Omitting culturally important information in summaries \citep{shen2017style}
\newline
\textbf{(2) Bias toward western narratives:} Prioritizing Western perspectives in global news summaries \citep{perez2022models} \\
\end{longtable}}

\section{Bias Evaluation\label{sec4}}

As LLMs become increasingly integrated into various real-world applications, from healthcare decision making to legal judgments and everyday digital interactions, their potential to propagate biases has raised significant ethical and societal concerns \cite{liang2021towards}. These biases, deeply rooted in the data, model architecture, and even in post-processing steps, can lead to discriminatory outcomes against marginalized groups, amplifying existing inequalities \cite{gallegos2024bias}. Therefore, understanding and evaluating these biases are crucial steps in ensuring the fair and responsible deployment of LLMs.

This section focuses on the classification and evaluation of biases in LLMs, providing a comprehensive overview of the methods used to identify and assess biases at different stages of the model lifecycle \cite{gallegos2024bias}. We categorize bias evaluation methods into data-level, model-level, output-level, human-involved, and domain-specific approaches, each addressing distinct sources and manifestations of bias \cite{doan2024fairness}. By systematically exploring these methods, we aim to equip researchers and practitioners with the tools necessary to detect and mitigate biases, ultimately enhancing the fairness and trustworthiness of LLMs in diverse applications \cite{gallegos2024bias}.

Drawing on recent advancements in the field, we highlight key techniques such as demographic distribution analysis, fairness metrics, interpretability tools, and counterfactual fairness, among others \cite{doan2024fairness, liang2021towards}. Additionally, we discuss the role of human judgment and domain-specific evaluations in capturing context-dependent biases that automated methods may overlook \cite{gallegos2024bias}. Through this exploration, we underscore the importance of a multi-faceted approach to bias evaluation, one that combines technical rigor with contextual awareness to address the complex challenges of fairness in LLMs \cite{gallegos2024bias}.

To avoid ambiguity, we distinguish the taxonomy of bias types from the taxonomy of bias evaluation methods. The earlier intrinsic–extrinsic distinction classifies where bias originates or manifests, whereas the present section classifies how bias is evaluated. Within this evaluation taxonomy, data-level, model-level, and output-level methods refer to the primary technical levels at which bias is assessed. In contrast, human-involved and domain-specific evaluations are best understood as cross-cutting evaluation modes: they can be applied to data-, model-, or output-level analyses rather than forming mutually exclusive categories of the same type.

\subsection{Data-Level Bias Evaluation Methods}
\begin{figure}[H]
    \includegraphics[width=0.98\textwidth]{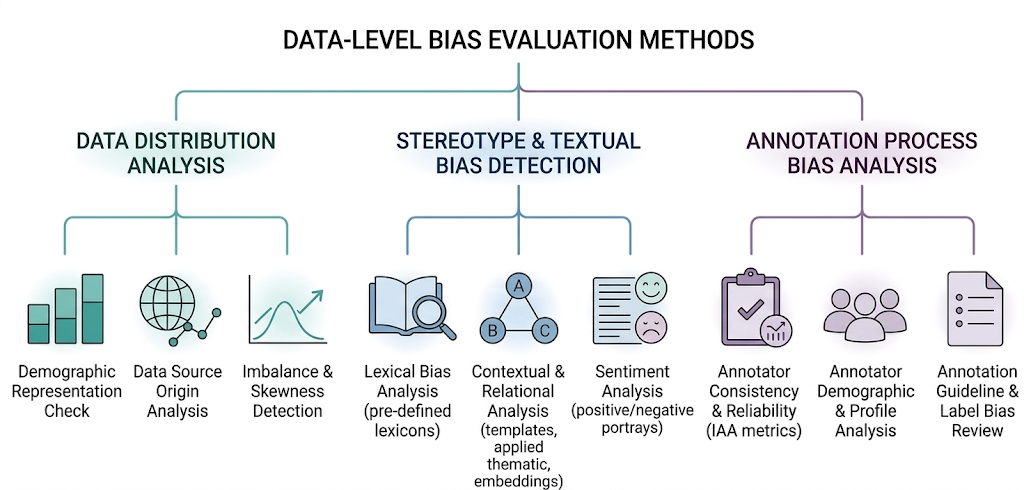}
  \caption{Taxonomy of data-level bias evaluation methods.}
 \label{fig:4p1_data_level_bias}
\end{figure}

Data-level bias evaluation methods focus on identifying and quantifying biases inherent in the training data of LLMs. These biases, rooted in the data, can significantly impact model outputs and exacerbate societal stereotypes. Figure~\ref{fig:4p1_data_level_bias} illustrates the taxonomy for data-level bias evaluation that we discuss in this section.

\vspace{-6pt}

\begin{itemize}
    \item \textbf{{Data distribution analysis:} 
}
    Data distribution analysis is a critical step in identifying and understanding bias at the data level, as it provides insights into how different demographic groups or categories are represented in the dataset used to train LLMs. The primary focus of this analysis is to ensure that the data is balanced and representative, minimizing the risk of perpetuating or amplifying existing biases in the  trained~model.

    \begin{itemize}[leftmargin=8.5mm,labelsep=5.5mm]
    \item \textbf{{Representation across demographics:}} Research has consistently demonstrated that the representation of demographics in training data can significantly influence the outcomes of LLMs. For instance, LLMs trained on datasets with disproportionate representation of certain demographic groups are prone to exhibit biases that favor those groups~\cite{simmons2023large,wang2024large,gorti2024unboxing}.
    For example, an analysis of political biases in LLMs revealed that most conversational LLMs exhibit left-of-center political preferences when probed with politically charged questions or statements~\cite{rozado2024political}. The study demonstrated that LLMs can be steered towards specific political orientations through supervised fine-tuning (SFT) with modest amounts of politically aligned data.

    To assess representation, statistical tools can be employed to measure the distribution of demographic attributes within the dataset. Key metrics might include the relative frequency of each demographic group or the variance in representation across different categories. Visual tools such as histograms, bar charts, or demographic distribution tables can be useful for identifying disparities, providing a clear picture of which groups are over-represented or under-represented in the data.

    \item \textbf{{Imbalance and skewness detection:}} 
    Imbalance and skewness in data distribution can lead to biased model behavior, where the model is overly influenced by the majority class or demographic group. Detecting these issues is crucial for ensuring that the model performs fairly across all segments of the population.

    Imbalances occur when certain classes or demographic groups are significantly more prevalent in the dataset than others. This can result in the model being biased towards the majority group, leading to poor performance for minority groups. Techniques such as Gini coefficients~\cite{khanuja2022evaluating}, entropy measures~\cite{csaky2019improving}, vocabulary usage~\cite{durward2024evaluating}, and frequency analysis can be used to quantify imbalance. Skewness can be visualized using skew and stereotype quantifying metrics~\cite{de2021stereotype, dang2024data}  box plots, histograms, or cumulative distribution functions (CDFs). These tools help identify the extent of skewness in the data and guide the selection of appropriate mitigation strategies.

    \item \textbf{{Data source analysis:}} 
    The quality and bias of a model are heavily influenced by the sources of the data used for training. Biases inherent in the data sources can significantly impact the model's behavior. For instance, if a model is trained primarily on data from Western countries, it may perform poorly when applied in non-Western contexts.

    Evaluating the bias of data sources for LLM training involves analyzing the origins, diversity, and quality of the data, as well as its impact on model performance. This includes cataloging the data sources to ensure they cover a wide range of geographical, cultural, and linguistic contexts, and examining the credibility and inherent biases of each source. For instance, research~\cite{shen2023slimpajama} on the SlimPajama dataset, which includes a rigorously deduplicated combination of web text, Wikipedia, {GitHub,} 
 and books, reveals how different data combinations affect LLM performance. The SlimPajama-DC study highlights two key aspects: the impact of global versus local deduplication on model performance and the importance of data diversity post-deduplication. The findings indicate that models trained with highly deduplicated, diverse datasets outperform those trained on less-refined data, underscoring the significance of comprehensive and balanced data sources. However, duplicate removal is not always neutral with respect to representation. If content from minority, low-resource, or less frequently documented communities is already sparse, aggressive deduplication may disproportionately remove repeated expressions of those voices, thereby reducing their visibility in the corpus and unintentionally increasing representativeness bias. Bias detection tools and adjustments in data weighting can further help in managing these biases, supported by transparent documentation and regular reviews to ensure fairness and accuracy in the model outputs.
    \end{itemize}

    \item \textbf{{Stereotype and bias detection in text data:}} 
    Stereotype and bias detection involves analyzing the content of the training data to identify and quantify biases related to stereotypes, offensive language, or prejudiced statements. This can be achieved through the following.

    \begin{itemize}[leftmargin=8.5mm,labelsep=5.5mm]
    \item \textbf{{Lexical analysis:}} 
    This method focuses on identifying specific words or phrases within the training data that are associated with bias or stereotypes. Lexical analysis relies on predefined lexicons, which are curated lists of terms known to carry biased or stereotypical connotations. Tools like Hatebase, an extensive repository of hate speech terms, provide a valuable resource for identifying harmful language in text data. Hate speech can also be measured by a large-scale study~\cite{silva2016analyzing}. BiCapsHate~\cite{kamal2023bicapshate} is a deep learning model used to detect hate speech in online social media posts.
    Similarly, the Dictionary of Affect in Language (DAL)~\cite{whissell1989dictionary} categorizes words based on their emotional connotations, offering insights into how language can perpetuate stereotypes. 
    By scanning the training data for occurrences of these terms, researchers can quantify the prevalence of biased language and identify specific areas where the data may reinforce harmful stereotypes. For example, a study by~\cite{dev2021measures} demonstrated how biased language in training data could lead to the propagation of racial and gender stereotypes in NLP models.

    \item \textbf{{Contextual analysis:}} While lexical analysis identifies the presence of specific biased terms, contextual analysis delves deeper into how these terms are used within the text. This method employs advanced NLP techniques to examine the context in which potentially biased language appears, allowing for the identification of subtler forms of bias. For instance, the same word might be used neutrally in one context but carry a prejudiced meaning in another. Contextual analysis examines sentence structure, co-occurrence patterns, and the surrounding language to uncover these nuances. 
    A similar study by~\cite{bolukbasi2016man} has shown that even when biased language is not overtly present, underlying patterns can still perpetuate stereotypes, such as gender biases, in word embeddings. 
    This approach is crucial for identifying and mitigating biases that may not be immediately apparent but can significantly impact the behavior of LLMs. {Zhao et al.}~\cite{zhao2019gender} 
    quantified and analyzed the gender bias exhibited in ELMo's contextualized word vectors. Contextual analysis can be implemented by methods like template-based method~\cite{kurita2019measuring}, applied thematic analysis~\cite{mackieson2019increasing}, and contextualized word embedding analysis~\cite{basta2019evaluating}.

    \item \textbf{{Sentiment analysis:}} Sentiment analysis is another vital tool in detecting bias in text data. This technique assesses the sentiment—positive, negative, or neutral—associated with different demographic groups or topics within the training data. By analyzing how different groups are described, researchers can identify patterns of negative or biased portrayals that could influence the model's outputs. For example, if certain demographic groups are consistently associated with negative emotions or connotations, the model might learn to replicate these biases in its predictions or interactions. Research by~\cite{kiritchenko2018examining} has highlighted how sentiment analysis can be used to detect and address such biases in text data. This method provides a quantitative measure of bias, enabling targeted interventions to reduce the impact of these biases on the model's behavior. 
    Sentiment analysis can be implemented using various approaches, including traditional machine learning methods~\cite{wankhade2022survey, saxena2022introduction, hasan2019sentiment}, frameworks based on preprocessed data~\cite{hasan2019sentiment}, and transformer-based methods~\cite{devlin2018bert,naseem2020transformer,abdullah2022deep}. 

    \end{itemize}

    \item \textbf{{Annotation bias analysis:}} 
    Annotation bias analysis is a crucial component in evaluating and mitigating bias in LLMs. This process involves examining the biases introduced during the data annotation phase, where human annotators label or categorize training data. Since annotators bring their own biases and perspectives, their subjective decisions can inadvertently introduce skewed or biased annotations, which in turn affect the performance and fairness of the LLMs. To perform annotation bias analysis, researchers must review the annotation guidelines and the training process for annotators, ensuring that they are designed to minimize bias. Additionally, evaluating the consistency and fairness of annotations across different demographic groups and annotators helps identify any disparities. Tools such as inter-annotator agreement metrics~\cite{artstein2017inter} and statistical analysis~\cite{paun2022statistical} of annotated data can reveal potential biases. For example, research by~\cite{havens2022uncertainty} on gender bias in annotated datasets highlights how annotation practices can reinforce stereotypes and biases, emphasizing the need for rigorous analysis and correction methods. Another way to reduce annotation bias is a human--LLM collaborative approach~\cite{wang2024human}. By addressing annotation bias, researchers can enhance the quality and fairness of training data, leading to more balanced and unbiased LLMs.

\end{itemize}

\subsection{Model-Level Bias Evaluation Methods}
Model-level bias evaluation methods assess biases that arise during the training and prediction stages of LLMs, focusing on whether the model exhibits discriminatory behavior toward specific groups.

\begin{itemize}
    \item \textbf{{Fairness metrics:}} These metrics are pivotal in evaluating the model’s output fairness across different groups. Common metrics include the following.
    \begin{itemize}[leftmargin=8.5mm,labelsep=5.5mm]
        \item \textbf{{Equal opportunity:}} Ensures that the true positive rate (TPR) is similar across groups defined by sensitive attributes such as gender or race. For example, a sentiment analysis model should equally recognize positive sentiments across both male- and female-associated names \cite{liang2021towards, gallegos2024bias}.
        
        \item \textbf{{Predictive parity:}} Focuses on the consistency of prediction accuracy (precision) across groups. This is particularly important in applications like credit scoring, where unequal predictive performance can lead to discriminatory outcomes against minority groups \cite{doan2024fairness, gallegos2024bias}.
        
        \item \textbf{{Calibration:}} This metric checks whether predicted probabilities align with actual occurrences, particularly across different demographic groups, ensuring that the model’s confidence in its predictions is justified \cite{doan2024fairness, esiobu2023robbie}.
    \end{itemize}
    
    \item \textbf{{Interpretability tools:}} Tools such as SHAP (Shapley Additive Explanations) \cite{mosca2022shap} and LIME (Local Interpretable Model-agnostic Explanations) \cite{wu2024auditing} provide insights into how specific features influence model predictions, helping identify biases. For instance, SHAP values can reveal that gendered words heavily influence predictions, indicating potential gender bias in the model \cite{liang2021towards, lin2024towards}. Similarly, LIME approximates complex models locally with interpretable models to highlight the impact of features on individual predictions, which is crucial for diagnosing bias \cite{wu2024auditing}. However, these tools should be interpreted with caution in LLMs with billions of parameters. In such massive models, feature attributions are often approximate and local, and may be unstable across prompt wording, tokenization, sampling settings, or semantically similar inputs. They may also fail to capture complex high-order interactions and therefore should not be treated as definitive causal explanations of bias. For this reason, SHAP- and LIME-based analyses are best used as diagnostic signals that should be corroborated with counterfactual, robustness, and output-level evaluations.
    
    \item \textbf{{Counterfactual fairness:}} This method generates counterfactual examples by altering sensitive attributes (e.g., changing names from traditionally male to female) to see if the model’s outputs remain invariant. A fair model should ideally produce the same outcomes regardless of these attribute changes \cite{doan2024fairness, li2023survey}.
    Here, counterfactual fairness is used in the model-level sense: it examines whether the model’s decision rule or prediction remains invariant when only a sensitive attribute is changed, rather than comparing differences in the wording or tone of free-form generated responses.
    \end{itemize}
    
Table~\ref{tab:model_bias_evaluation_clean} summarizes the three categories of model-level bias evaluation methods discussed above.

{\footnotesize
\renewcommand{\arraystretch}{1.2}
\begin{longtable}{
  >{\raggedright\arraybackslash}p{0.22\textwidth}
  >{\raggedright\arraybackslash}p{0.68\textwidth}}
\caption{Summary of model-level bias evaluation methods.}
\label{tab:model_bias_evaluation_clean}\\
\toprule
\textbf{Method} & \textbf{Description} \\
\midrule
\endfirsthead

\multicolumn{2}{c}{\tablename\ \thetable\ -- \textit{continued from previous page}} \\
\toprule
\textbf{Method} & \textbf{Description} \\
\midrule
\endhead

\midrule
\multicolumn{2}{r}{\textit{continued on next page}} \\
\endfoot

\bottomrule
\endlastfoot

Fairness metrics & \textbf{Equal opportunity:} Ensures similar true positive rates (TPRs) across sensitive groups. \\
\cmidrule{2-2}
& \textbf{Predictive parity:} Checks for consistent prediction accuracy across groups. \\
\cmidrule{2-2}
& \textbf{Calibration:} Aligns predicted probabilities with actual outcomes for fairness. \\
\midrule
Interpretability tools & \textbf{SHAP:} Explains the impact of individual features on model predictions. \\
\cmidrule{2-2}
& \textbf{LIME:} Provides local approximations of complex models for feature impact analysis. \\
\midrule
Counterfactual fairness & \textbf{Scenario testing:} Alters sensitive attributes (e.g., gender) to test output consistency. \\
\cmidrule{2-2}
& \textbf{Equity check:} Verifies that changes do not affect model outcomes unfairly. \\
\end{longtable}}

\subsection{Output-Level Bias Evaluation Methods}
\begin{figure}[H]
    \includegraphics[width=0.97\textwidth]{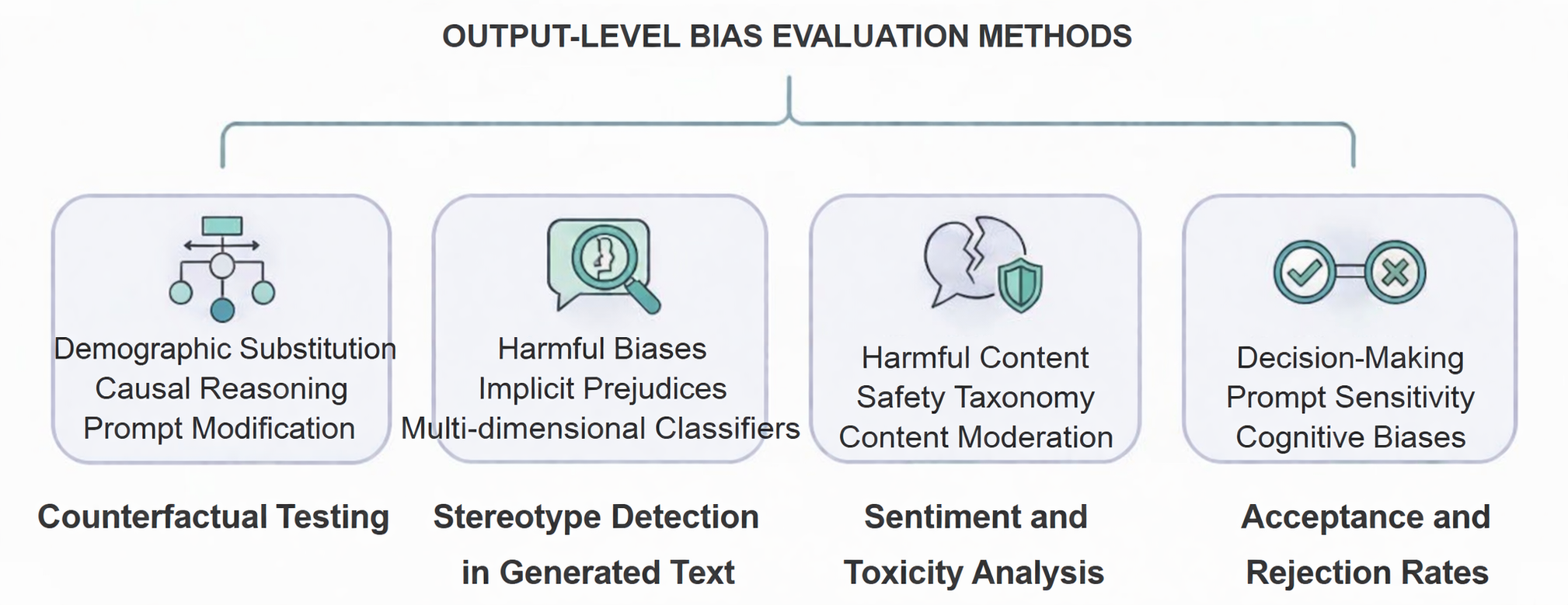}
  \caption{Taxonomy of output-level bias evaluation methods.}
 \label{fig:4p3_output_level_bias}
\end{figure}

Output-level bias evaluation methods assess how LLMs generate responses, especially with regard to fairness and neutrality across different demographic groups. These methods aim to identify whether the model's predictions, recommendations, or generated content reflect biases that may have been learned during training, including those based on race, gender, socioeconomic status, political orientation, and more. Here, we explore four major metrics used to detect bias in LLM-generated text: counterfactual testing, stereotype detection, sentiment and toxicity analysis, and acceptance and rejection rates (see Figure~\ref{fig:4p3_output_level_bias}).
Unlike model-level counterfactual fairness, output-level counterfactual testing compares the generated outputs themselves under minimally modified prompts, focusing on whether the content, framing, sentiment, or recommendation changes across demographic substitutions.

\begin{itemize}
    \item \textbf{{Counterfactual testing:}} 
        Counterfactual testing involves modifying input prompts by altering specific attributes, such as gender, race, or ethnicity, while keeping the rest of the context unchanged. This method evaluates whether the LLM's output varies based on these demographic changes, allowing researchers to isolate the effect of these attributes on the model's behavior. For instance, swapping ``John'' with ``Maria'' in a sentence can reveal if the model responds differently due to the gender of the subject.
        
        Recent work~\cite{wang2024causalbench} emphasizes the importance of robust causal reasoning in LLMs to enhance fairness, arguing that a strong understanding of causal relationships can mitigate biases and reduce hallucinations. Beyond fairness, this connection is also important for understanding hallucinations. When a prompt is ambiguous or underspecified, biased social associations learned from the training corpus may act as anchors, leading the model to generate stereotype-consistent details even when those details are not supported by the input. For example, the model may infer occupations, traits, intentions, or background facts from demographic cues because such associations are statistically frequent in the data, not because they are causally justified in the given context. In this sense, some hallucinations can be viewed not merely as factual errors, but as bias-driven distortions in which stereotype-consistent priors fill evidential gaps. Stronger causal reasoning may reduce this risk by helping the model distinguish socially correlated attributes from causally relevant information. Another study~\cite{banerjee2024all} introduces a dynamic framework that compares outputs across different demographic groups in the same context, improving the fairness of generated text without requiring costly model retraining. Additionally, new tools~\cite{cheng2024interactive} for generating and analyzing counterfactuals allow users to explore LLM behavior interactively, ensuring the counterfactuals are both meaningful and grammatically accurate.
        
        These advancements underscore the power of counterfactual testing in identifying biases in LLMs. By systematically altering demographic features in prompts, researchers can evaluate how models treat different groups and ensure equitable outputs across all demographics.

    \item \textbf{{Stereotype detection in generated text:}}
    Stereotype detection in generated text focuses on identifying and mitigating harmful biases that LLMs may perpetuate in their outputs. Since LLMs are trained on vast amounts of publicly available data, which often contain stereotypical narratives related to race, gender, profession, and religion, the risk of these biases appearing in generated content is significant. This method assesses how models reproduce stereotypes, providing insights into implicit prejudices embedded in their responses.

    One approach~\cite{wu2024auditing} introduced a unified dataset combining multiple stereotype detection datasets. Researchers fine-tuned LLMs on this dataset and found that multi-dimensional classifiers were more effective in identifying stereotypes. This study also highlighted the use of explainable AI tools to ensure models align with human understanding. Another study~\cite{bai2024fairmonitor} developed a dual framework that combines static evaluations with dynamic, real-world scenario simulations. This dynamic aspect is particularly effective in detecting subtle, context-specific biases that static tests might miss. A qualitative method presented in
    ~\cite{babonnaud2024bias} uses prompting techniques to uncover implicit stereotypes in LLM-generated text. This approach, focusing on biases like gender and ethnicity, employs the Tree of Thoughts technique to systematically reveal hidden prejudices and provide a reproducible method for stereotype detection. These methods collectively help researchers detect and mitigate biases in LLM outputs, offering a deeper understanding of how stereotypes manifest in generated text.

    \item \textbf{{Sentiment and toxicity analysis:}}  
    Sentiment and toxicity analysis is crucial for evaluating output-level biases in LLMs, specifically targeting harmful or offensive content and ensuring adherence to ethical standards. This area of evaluation not only addresses toxicity but also helps in identifying subtle biases that may emerge in    generated content.

    Llama Guard~\cite{inan2023llama} introduces a model-based approach to bias evaluation by using a safety risk taxonomy for classifying both prompts and responses in human--AI conversations. This model, fine-tuned on a carefully curated dataset, shows strong performance in detecting various forms of toxicity. It represents a practical application of bias evaluation methods by providing a dynamic tool for assessing and mitigating harmful content in real-world interactions. The definition-based toxicity metric~\cite{koh2024can} offers a flexible solution for bias evaluation by using LLMs to measure toxicity according to predefined criteria. This method outperforms traditional metrics, enhancing the F1 score significantly and addressing the limitations of existing models that rely on dataset-specific definitions. It exemplifies how bias evaluation methods can be refined to improve the detection of nuanced toxic content. Moderation Using LLM Introspection (MULI)~\cite{hu2024toxicity} advances bias evaluation by analyzing internal model responses to detect toxic prompts. This approach leverages patterns in response token logits and refusal behaviors, providing a cost-effective way to assess biases without requiring additional training.

    The OpenAI Moderations Endpoint represents a practical implementation of bias evaluation methods for filtering potentially harmful content. This tool helps developers assess and moderate LLM outputs, ensuring they align with safety and ethical guidelines. It demonstrates how automated tools can be integrated into the bias evaluation process to manage and mitigate risks associated with model-generated text.

    \item \textbf{{Acceptance and rejection rates:}} 
    Evaluating acceptance and rejection rates in LLMs is essential for identifying biases that may arise in decision-making tasks. Recent work has explored how these biases manifest, particularly in contexts such as hiring decisions or moral dilemmas, where the outputs of LLMs could reinforce existing societal inequalities.

    In one study~\cite{an2024large}, researchers assessed how LLMs respond to job applicants based on perceived race, ethnicity, and gender by manipulating first names. They found that acceptance rates were significantly higher for masculine White names compared to masculine Hispanic names. These acceptance and rejection rates were highly prompt-sensitive, meaning that variations in how the question was framed could lead to different outcomes. This suggests that LLMs can subtly reinforce biases depending on how inputs are structured. Another evaluation method~\cite{scherrer2024evaluating} used to study LLM behavior in moral decision making showed that models are consistent in choosing commonsense actions in unambiguous scenarios, while in more ambiguous cases, LLMs exhibited uncertainty and greater variability. Closed-source models often displayed more consistent preferences in these ambiguous scenarios, which suggests that proprietary models might encode specific tendencies in moral decision making.
    BiasBuster~\cite{echterhoff2024cognitive} provides a systematic method for uncovering and evaluating cognitive biases in LLMs, particularly in high-stakes decision making. By using a dataset of 16,800 prompts designed to probe various types of biases (e.g., prompt-induced, sequential), this framework allowed researchers to assess how LLMs handled acceptance and rejection decisions under different bias conditions.
\end{itemize}

\subsection{Human-Involved Bias Evaluation Methods}
Human-involved evaluation incorporates human judgment to identify complex and context-dependent biases in LLM outputs.
Human-involved evaluation is defined by the source of judgment (i.e., human assessors), whereas domain-specific evaluation is defined by application-specific norms, risks, and fairness criteria (e.g., healthcare, law, education); accordingly, a domain-specific evaluation may be human-involved, automated, or both.

\begin{itemize}
    \item \textbf{{Human review and assessment:}} Experts or crowd-sourced reviewers manually assess model outputs for biases. This method is particularly effective in detecting nuanced biases that automated tools might miss, such as subtle stereotypes or culturally specific biases \cite{doan2024fairness}.
    
    \item \textbf{{Qualitative research methods:}} Interviews, focus groups, and case studies offer deep insights into how different communities perceive model outputs, providing a qualitative dimension to bias evaluation that complements quantitative approaches~\citep{liang2021towards}.
\end{itemize}

\begin{figure}[H]
    \includegraphics[width=0.97\textwidth]{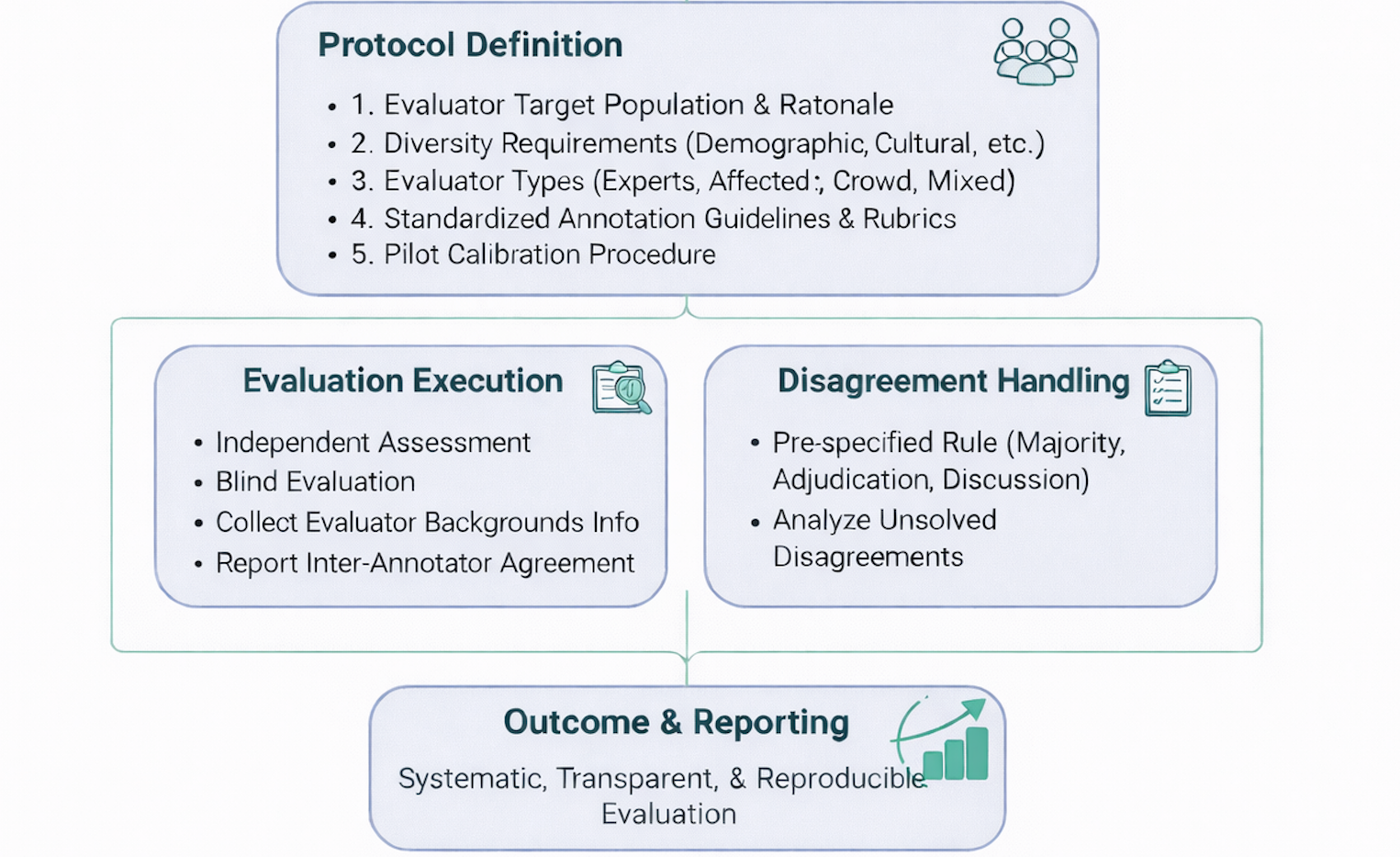}
  \caption{Human evaluation protocol design.}
 \label{fig:4p4_Human_Involved_Bias}
\end{figure}

To reduce subjectivity in human-involved bias evaluation, the evaluation protocol (Figure~\ref{fig:4p4_Human_Involved_Bias}) should be specified explicitly rather than left implicit. At a minimum, such a protocol should define: (1) the target population of evaluators and the rationale for their selection; (2) the diversity requirements of the evaluator pool, including demographic, cultural, linguistic, and professional diversity when the task involves socially situated judgments; (3) whether domain experts, affected-group members, crowd workers, or mixed panels are used; (4) standardized annotation guidelines, examples, and rating rubrics; and (5) the procedure for pilot calibration before formal evaluation. In practice, evaluators should assess outputs independently and, when possible, without access to model identity or study hypotheses, in order to reduce anchoring and confirmation effects. The manuscript should also report the number of evaluators, their relevant background characteristics, the assignment protocol, and inter-annotator agreement statistics when applicable~\cite{artstein2017inter}. For disagreement handling, a pre-specified rule should be provided, such as majority vote for straightforward labels, adjudication by a trained senior reviewer, or discussion-based consensus for context-sensitive cases; unresolved disagreements may also be retained and analyzed separately as indicators of ambiguity in the bias construct itself. Such protocol design helps ensure that human evaluation is systematic, transparent, and reproducible, rather than relying solely on subjective impressions.

\vspace{6pt}

\subsection{Conclusion}
Bias evaluation is critical for the responsible use of LLMs. By systematically categorizing and applying various evaluation methods, researchers can better understand and mitigate the biases that LLMs may propagate, thereby enhancing their fairness and societal impact.

\section{Bias Mitigation\label{sec5}}

Biases are often present in unstructured data, and LLMs trained on such data can not only learn these biases but sometimes amplify them as well \citep{bolukbasi2016man, sheng2019woman}.
Various methods of bias mitigation exist under different contexts, such as social bias \citep{garimella2021he, garimella2022demographic}, political bias~\citep{liu2022quantifying}, and stereotype bias \citep{nie-etal-2024-multilingual}.
Instead of reviewing these methods for bias mitigation by bias contexts, we categorize them by the timing of debiasing.
More specifically, methods for mitigating biases can be categorized into three distinct approaches: (1) preprocessing the input data prior to training LLMs; (2) changing the model's architecture and/or
picking the best model based on the predefined bias evaluation metrics; and (3) improving unbiasedness by calibrating the outputs at the decision-making stage.
We call these three approaches pre-model debiasing, intra-model debiasing, and post-model debiasing, respectively.
Table~\ref{tab: mitigation} provides an overview of the advantages and disadvantages of each approach.

\begin{table}[htbp]
\renewcommand{\arraystretch}{1.2}  
\caption{{Overview} of methods for bias mitigation.}
\footnotesize
\begin{tabularx}{\textwidth}{m{2.3cm}<{\raggedright}m{5.3cm}<{\raggedright}m{5cm}<{\raggedright}}
\hline
\textbf{Debiasing}  & \textbf{Methods} & \textbf{Summary}\\
\hline
Pre-model & Resampling; data augmentation; expert intervention & Time saving without model training; time consuming to annotate bias cases, ineffectiveness with a self-biased model, privacy issue of resampling \\
\hline
Intra-model & Equalization and declustering; movement pruning; transfer learning; dropout regularization;
causal inference & Flexibility in mitigating various types of biases, strong performance in empirical studies; time consuming for modifying and training models \\ \hline
Post-model & Reinforced calibration; Self-Debias; projection-based methods; causal prompting & Time saving without model training; demand of a large amount of data \\ \hline
\end{tabularx}
\label{tab: mitigation}
\end{table}

Figure~\ref{fig:mitigation_pipeline} provides a visual summary of the three-stage debiasing pipeline and the representative techniques within each stage.

\begin{figure}[H]
\resizebox{0.99\textwidth}{!}{
\begin{tikzpicture}[
    stage/.style={
        rectangle, draw, fill=green!15,
        text width=3.5cm, text centered,
        minimum height=1.2cm,
        font=\small\bfseries,
        rounded corners=3pt
    },
    method/.style={
        rectangle, draw, fill=white,
        text width=3.2cm, text centered,
        minimum height=0.7cm,
        font=\footnotesize,
        rounded corners=2pt
    },
    bigarrow/.style={->, very thick, >=stealth, color=gray!70},
    smallarrow/.style={->, thick, >=stealth},
    connector/.style={thick}
]

\node[stage] (pre)   at (0,0)    {Pre-Model\\Debiasing};
\node[stage] (intra) at (6,0)    {Intra-Model\\Debiasing};
\node[stage] (post)  at (12,0)   {Post-Model\\Debiasing};

\draw[bigarrow] (pre) -- node[above, font=\footnotesize\itshape] {Training} (intra);
\draw[bigarrow] (intra) -- node[above, font=\footnotesize\itshape] {Inference} (post);

\node[method, anchor=west] (p1) at (-0.4,-1.9) {CDA/CDS};
\node[method, anchor=west] (p2) at (-0.4,-3.0) {Resampling};
\node[method, anchor=west] (p3) at (-0.4,-4.1) {Expert curation};

\coordinate (preMid)      at ($(pre.south)+(0,-0.45)$);
\coordinate (preTrunkTop) at ([xshift=-0.45cm,yshift=0.28cm]p1.north west);
\coordinate (preTrunkBot) at (preTrunkTop |- p3.west);

\draw[connector] (pre.south) -- (preMid) -- (preTrunkTop);
\draw[connector] (preTrunkTop) -- (preTrunkBot);

\foreach \m in {p1,p2,p3}{
    \draw[smallarrow] (preTrunkTop |- \m.west) -- (\m.west);
}

\node[method, anchor=west] (i1) at (5.2,-1.9) {Equalization loss};
\node[method, anchor=west] (i2) at (5.2,-3.0) {Movement pruning};
\node[method, anchor=west] (i3) at (5.2,-4.1) {Transfer learning};
\node[method, anchor=west] (i4) at (5.2,-5.2) {Dropout/Causal};

\coordinate (intraMid)      at ($(intra.south)+(0,-0.45)$);
\coordinate (intraTrunkTop) at ([xshift=-0.45cm,yshift=0.28cm]i1.north west);
\coordinate (intraTrunkBot) at (intraTrunkTop |- i4.west);

\draw[connector] (intra.south) -- (intraMid) -- (intraTrunkTop);
\draw[connector] (intraTrunkTop) -- (intraTrunkBot);

\foreach \m in {i1,i2,i3,i4}{
    \draw[smallarrow] (intraTrunkTop |- \m.west) -- (\m.west);
}

\node[method, anchor=west] (o1) at (10.8,-1.9) {Self-Debias};
\node[method, anchor=west] (o2) at (10.8,-3.0) {Projection (INLP)};
\node[method, anchor=west] (o3) at (10.8,-4.1) {RL calibration};
\node[method, anchor=west] (o4) at (10.8,-5.2) {Causal prompting};

\coordinate (postMid)      at ($(post.south)+(0,-0.45)$);
\coordinate (postTrunkTop) at ([xshift=-0.45cm,yshift=0.28cm]o1.north west);
\coordinate (postTrunkBot) at (postTrunkTop |- o4.west);

\draw[connector] (post.south) -- (postMid) -- (postTrunkTop);
\draw[connector] (postTrunkTop) -- (postTrunkBot);

\foreach \m in {o1,o2,o3,o4}{
    \draw[smallarrow] (postTrunkTop |- \m.west) -- (\m.west);
}

\end{tikzpicture}
}
\caption{{Overview} 
of the three-stage bias mitigation pipeline. Pre-model methods intervene on training data, intra-model methods modify the training process or model architecture, and post-model methods adjust outputs at inference time. Representative techniques are listed under each stage.}
\label{fig:mitigation_pipeline}
\end{figure}

\subsection{Pre-Model Debiasing}
Pre-model debiasing approaches involve preprocessing the input data before training LLMs, so that the entire training data are balanced and representative of the entire population.
Typical ways of balancing training data include oversampling, undersampling, and synthetic data generation \citep{ferrara2023should}.
We review the commonly used pre-model debiasing methods below.

One of the widely used methods is counterfactual data augmentation (CDA), which is often used to mitigate bias in the literature \citep{zmigrod-etal-2019-counterfactual, webster2020measuring, barikeri-etal-2021-redditbias, mondal2024mitigating}.
Counterfactual data substitution (CDS), a variant of CDA, is also proposed to randomly substitute potentially biased text to avoid duplication \citep{Maudslay2019ItsAI}.
Generally speaking, CDA rebalances a corpus by swapping bias attribute words in a dataset.
For example, we may swap the sentence 
\clearpage
$$\textit{``The professor came to the classroom but he forgot to bring his laptop''}$$
with 
$$\textit{``The professor came to the classroom but she forgot to bring her laptop''}.$$

{Meade et al.} \cite{meade_2022_empirical} further empirically evaluate CDA for mitigating religious bias by swapping religious terms.

{Serouis et al.} \cite{serouis2024exploring} propose an iterative method in which a human 
can oversee the preprocessing step.
That is, before training begins, a dataset report is sent to an LLM. 
An expert then decides the next step: either issuing a command to the LLM to create new training data or starting the training process. 
The decision is made by measuring the quality of the generated data.
If a command is used to generate alternative training data, iterative feedback between the expert and the LLM continues until predefined criteria are satisfied, leading to a decision to proceed with model training. 
After training and experimentation, detailed results concerning protected groups are provided to the LLM for summarization (particularly for multiple groups and subgroups) or directly to the expert. 
Based on these comprehensive reports, the expert may choose to use the LLM to produce a revised dataset version to mitigate potential biases.

\subsection{Intra-Model Debiasing}

Intra-model debiasing works by selecting the model that prioritizes unbiasedness and fairness during the model fitting.
For example, {Ref.} \cite{garimella2021he} proposes a method to mitigate explicit and implicit biases in BERT using existing and newly proposed loss functions.
The proposed work focuses on gender bias and uses a predefined list of target word pairs and a text dataset as inputs. 
It takes the pretrained BERT and further trains it on the given input, while mitigating the existing social biases by equalizing and declustering.
The equalizing stage aims to equalize the associations of neutral words by defining an equalizing loss \citep{qian-etal-2019-reducing}.
Observing that words are stereotypically associated with a group (e.g., {\textit{delicate}, \textit{pink}, \textit{beautiful}),} 
 the authors develop a declustering stage to decluster the implicit clusters by defining a declustering loss.
A similar technique is also applied to mitigate social stereotypes when exposed to different demographic groups \citep{garimella2022demographic}.
Moreover, {Ref.}~\cite{joniak-aizawa-2022-gender} presents an innovative framework for examining bias in pretrained transformer-based language models through movement pruning. 
It identifies a subset of the model that contains less bias than the original by pruning it while fine-tuning on a debiasing objective. 
It explores pruning attention heads, which are crucial components of transformers, by pruning square blocks and introducing a novel method for pruning entire heads. 
An interesting bias--performance trade-off for gender bias from this study is that the level of bias increases as the model's performance improves.

Another line of studies focuses on the technique of transfer learning \citep{zhuang2020comprehensive, azunre2021transfer, ge2024openagi, liu2024unified}.
Transfer learning leverages knowledge learned from a pretrained model to improve the performance of a new model on a relevant task but with fewer data points.
More specifically, it fine-tunes a pretrained model on the new task by updating the weights of neurons or modifying the network architecture.
For example, FairDistillation \citep{delobelle2022fairdistillation}, a cross-lingual method based on knowledge distillation, is used to build smaller language models while managing specific biases.
It leverages existing language models as a teacher, providing a richer training signal without the need for retraining from scratch. 
To avoid transferring learned correlations to new language models, it further substitutes CDA's augmentation strategy with probabilistic rules between tokens.
Furthermore, we notice that a few works utilize multilinguality as bias mitigation \citep{ahn-oh-2021-mitigating,levy-etal-2023-comparing, nie-etal-2024-multilingual}. 
These works advocate multilingual training to mitigate biases.
They demonstrate the effectiveness of multilinguality in reducing bias and boosting prediction accuracy \citep{nie-etal-2024-multilingual}.

Another useful method is dropout regularization, which is studied in 
\citep{srivastava2014dropout, webster2020measuring}.
Briefly, they explore the impact of increasing dropout parameters for BERT's and ALBERT’s attention weights and hidden activations, followed by an additional phase of pretraining. 
Their experiments reveal that dropout regularization reduces gender bias in these models. 
They propose that dropout disrupts the attention mechanisms in BERT and ALBERT, thereby preventing the models from learning undesirable word associations.

Grounded in the theory of causal inference, causality-guided methods have been investigated due to the theoretical guarantees and good generalization  \citep{da2024reducing, cai2024locating}.
For instance, researchers have proposed a bias reduction method for sentiment classification \citep{da2024reducing}.
This framework treats the neural network as a causal directed acyclic graph (DAG).
Borrowing the concept of causal mediation analysis for neural networks \citep{vig2020investigating}, the authors study the direct and indirect effects of individual neurons to identify parts of a model where gender bias is causally implicated. 
Finally, the authors apply counterfactual training to only fine-tune the BERT target layers.
Moreover, Ref. \cite{cai2024locating} uses causal mediation analysis to trace the causal effects of different components’ activation within an LLM.
In this way, they were 
able to investigate the storage locations and generation mechanisms of
gender bias in LLMs.
The least square debias method (LSDM) was proposed 
to modify parameters to mitigate
gender bias in models.

\subsection{Post-Model Debiasing}

Post-model debiasing mitigates bias by adjusting the output of LLMs.
The benefits of post-model debiasing are that there are no changes in the training data and no changes in the architecture of the LLM.
For example, reinforced calibration can be used to mitigate political bias in GPT-2~\citep{liu2021mitigating, liu2022quantifying}.
It keeps the architecture of GPT-2 unchanged but calibrates bias during the generation.
More specifically, researchers have presented a reinforcement learning-based framework for political biases in two modes: word embedding debias and classifier-guided debias.
For the word embedding debias, the researchers pick the unbiased token at each generation step by forcing neutral words to maintain equal distances from groups of sensitive words (e.g., male and female) in the embedding space \citep{bolukbasi2016man, park-etal-2018-reducing, zhao2018learning}.
The classifier-guided debiasing works when the bias is not purely word-level \citep{bordia-bowman-2019-identifying}.

Self-Debias is another work that leverages the model's internal knowledge to generate unbiased prompts \citep{schick2021self}.
The necessary condition for this approach is that if a model is able to discern the presence of undesired biases, it should be able to avoid them at test time.
Self-Debias first detects when the models output undesirable attributes based on internal knowledge.
It then reduces the probability of producing a biased test (such as \textit{{sexist}}, \textit{{racist}}, and \textit{{violent}}) using a decoding algorithm.
This has been empirically evaluated as a strong debiasing technique \citep{meade_2022_empirical}.

Projection-based methods have also been shown to be useful for mitigating biases.
Notable methods such as SENT-DEBIAS \citep{liang2020towards} and INLP \citep{ravfogel-etal-2020-null} work by projecting a model's representations onto a subspace.
More specifically, SENT-DEBIAS involves four steps:    (1) identifying words that display biased attributes (e.g., \textit{{he}} and \textit{{she}}); (2) embedding these words within biased attribute sentences to generate their sentence representations by CDA; (3) estimating the bias subspace within these sentence representations by principle component analysis (PCA); and (4) debiasing general sentences by removing their projection onto this bias subspace.
INLP roughly debiases a model’s representations by training a linear classifier to predict the protected attribute we want to remove (e.g., \textit{{gender}}) from the representations. 
The representations are then debiased by projecting them into the nullspace of the trained classifier’s weight matrix, effectively eliminating all the information the classifier used to predict the protected attribute from the representation. 
This process can be applied iteratively to further debias the representation.

In alignment with causal methods in intra-model debiasing, researchers have also developed methods based on causal prompting \citep{li2024steering, zhang2024causal}.
Prompting methods have gained in popularity since the general public has no access to the model's internal structure due to business interests.
Specifically, {Ref.} \cite{li2024steering} proposes causal prompting based on front-door adjustment \citep{pearl2016causal}.
The proposed method modifies prompts without access to the parameters and logits of LLMs.
First, it queries LLMs to generate chains of thought (CoTs) 
$m$ times with the input prompt (demonstration examples and a question of the test example).
An encoder-based clustering algorithm is applied to these CoTs and the top $K$ representative CoTs are selected.
Next, it retrieves the optimal demonstration examples for each representative.
Finally, the LLMs are queried $T$ times to obtain $T$ answers for each representative CoT, and the final answer is obtained by a weighted voting.
{Ref.}~\cite{zhang2024causal} constructs detailed causal models for both the underlying data-generating process of the training corpus and the LLM's reasoning process. 
They recognize that selection mechanisms are crucial in determining how the LLM's output can be influenced by different prompts.
Additionally, they propose a causality-guided debiasing framework. 
By gaining causal insights into the data-generating processes, the necessary conditions that prompt-design strategies must meet to achieve principled and effective debiasing are identified.

\subsection{Critical Comparison of Mitigation Strategies}
\label{sec:mitigation_comparison}

Having reviewed the three categories of debiasing approaches, we now offer a critical comparison of their effectiveness, computational requirements, and practical trade-offs. Table~\ref{tab:mitigation_comparison} provides an estimated comparison of the computational costs across representative methods.

{\footnotesize
\renewcommand{\arraystretch}{1.22}
\begin{longtable}{
  >{\raggedright\arraybackslash}p{0.13\textwidth}
  >{\raggedright\arraybackslash}p{0.19\textwidth}
  >{\raggedright\arraybackslash}p{0.21\textwidth}
  >{\raggedright\arraybackslash}p{0.14\textwidth}
  >{\raggedright\arraybackslash}p{0.18\textwidth}}
\caption{Estimated computational cost comparison of representative bias mitigation strategies. Costs are approximate and depend on model size and implementation.}
\label{tab:mitigation_comparison}\\
\toprule
\textbf{Category} & \textbf{Method} & \textbf{Key Mechanism} & \textbf{Relative Cost} & \textbf{GPU Requirement} \\
\midrule
\endfirsthead

\multicolumn{5}{c}{\tablename\ \thetable\ -- \textit{continued from previous page}} \\
\toprule
\textbf{Category} & \textbf{Method} & \textbf{Key Mechanism} & \textbf{Relative Cost} & \textbf{GPU Requirement} \\
\midrule
\endhead

\midrule
\multicolumn{5}{r}{\textit{continued on next page}} \\
\endfoot

\bottomrule
\endlastfoot

Pre-model & CDA/CDS & Augmenting training data & Low & Minimal \\
Pre-model & Expert curation & Human review of data & Medium & None \\
Intra-model & Equalization loss & Modified training objective & High & Full training \\
Intra-model & Movement pruning & Pruning biased subnetworks & High & Full training \\
Intra-model & FairDistillation & Knowledge distillation & Medium--high & Distillation run \\
Intra-model & Dropout regularization & Additional pretraining & Medium & Pretraining pass \\
Post-model & Self-Debias & Modified decoding & Low & Inference only \\
Post-model & SENT-DEBIAS/INLP & Subspace projection & Low--medium & Inference + PCA \\
Post-model & Reinforced calibration & RL-based generation & Medium & RL training \\
Post-model & Causal prompting & Prompt modification & Low--medium & Multiple queries \\
\end{longtable}}

Several key trade-offs emerge from this comparison. Pre-model methods are conceptually straightforward and computationally inexpensive but cannot address biases that arise from model architecture or training dynamics. They are also limited in their ability to handle biases that are not explicitly encoded in individual words or phrases but emerge from complex distributional patterns.
Intra-model methods offer the most direct intervention into the bias-producing mechanisms but require full model retraining or fine-tuning, which is computationally expensive and may not be feasible for the largest contemporary LLMs. 
Furthermore, intra-model methods are model-specific and may not generalize across architectures.
Post-model methods are attractive for their flexibility and low computational overhead, as they do not require modifying the training data or model parameters. However, they address only the symptoms of bias rather than its root causes, and their effectiveness depends on the ability to detect bias at inference time. Self-Debias, for instance, relies on the assumption that the model can recognize its own biases, which may not hold for all bias types, especially subtle or intersectional biases.

In practice, a combination of methods across all three stages is likely to yield the most robust debiasing outcomes. Pre-model data curation can reduce the most egregious data-level biases, intra-model techniques can address biases encoded in model representations, and post-model methods can serve as a final safety net during deployment. Future research should focus on developing integrated debiasing pipelines that coordinate interventions across these stages.

\section{Ethical Concerns and Legal Challenges\label{sec6}}
Due to rapid advancements and successful integration of LLMs into systems that are widespread throughout our society, more and more concerns are being raised that LLMs can learn and amplify societal biases (gender, race, religion, age, etc.) embedded in the enormous Internet data they are trained on \citep{gallegos2024bias}. As a result, LLMs can potentially generate biased outputs among different social groups, which falls under the ethical concerns of AI. Furthermore, disparate treatment caused by biased outputs in decision-making processes can result in legal challenges under anti-discrimination laws. 

We categorize biased outputs and disparate treatment according to a pre-established taxonomy: representational harms and allocational harms \citep{blodgett2020language}. Representational harms refer to when an automated system exhibits negative attitudes or misrepresents, even ignores existence of certain social groups. Allocational harms refer to when an automated system allocate resources in favor of some social groups over others. 

After reviewing many types of representational harms described in the literature, we constrained our discussion to four categories that can cover most representational harms or ethical concerns:
\begin{itemize} 
    \item Stereotyping.
    \begin{itemize}[leftmargin=8.5mm,labelsep=5.5mm]
        \item Definition: Negative abstractions about a labeled social group.
        \item Result 1: Reinforced social bias against certain social groups.
        \item Result 2: Toxic language towards certain social groups.
    \end{itemize} 
    \item Misrepresentation. 
    \begin{itemize}[leftmargin=8.5mm,labelsep=5.5mm]
        \item Definition: An incomplete sample generalized to a social group. 
        \item Result 1: Reinforced normativity of the dominant social group and implicit exclusion or devaluation of other groups.
        \item Result 2: Biased outputs that mischaracterize certain social groups.
    \end{itemize}   
    \item Disparate system performance. 
    \begin{itemize}[leftmargin=8.5mm,labelsep=5.5mm]
        \item Definition: Degraded understanding or model performance in LLM language processing or generation between social groups.
        \item Result 1: Allocational harms.
    \end{itemize}
\end{itemize}

Stereotyping: LLMs can reinforce over-simplified and often negative labels attached to certain social groups. \citet{abid2021antimuslim} demonstrate that LLMs capture Muslim-violence bias in various use cases and the Muslim-violence bias is more severe than commonly seen biases for other religious groups. \citet{bender2021dangers} describe how LLMs tend to refer 
to ``women doctors'' since even in today's world, doctor is a male-dominant occupation. Stereotypes perpetuated and even exacerbated by LLMs can further translate into offensive language that targets a social group. \citet{gehman2020realtoxicityprompts} evaluate five language models (including GPT-1 to 3) using a ``REALTOXICITYPROMPTS'' machine and show that all five models degenerate into toxicity over 0.5 (assigned by a widely accepted toxicity classifier) within less than 100 generations from non-toxic prompts. Toxic generations from language models can range from the likes of ``Islamic slaughtering'' to ``lying white media''. 

Misrepresentation: The disproportionate representation of marginalized or minority social groups in LLM training data can lead to implicit exclusion of marginalized groups and mischaracterization of certain groups. \citet{bender2021dangers} raise an example of `both genders' that exclude non-binary identities from the context. \citet{navigli2023biases} provide several examples of misrepresentation across gender, age, nationality, ethnicity, etc. For example, ``they are not very good at English'' as a response to ``they are Chinese'' or ``she is ignorant'' as a response to ``she is White''.

Disparate system performance: \citet{koenecke2020racial} reveal that all of the five state-of-the-art LLM-based automated speech recognition (ASR) systems show higher word error rates for black speakers than white speakers. \citet{saunders2020reducing} demonstrate that the over-prevalence of male-centric forms in machine translation training data result in better translations for male-centric sentences and for sentences that contain male-dominant roles such as doctors and developers. Next, we discuss disparate system performance in LLM-automated hiring and healthcare tools, which often leads to allocational harms that raise anti-discrimination legal challenges.

\begin{itemize}
    \item Allocational harms.
    \begin{itemize}[leftmargin=8.5mm,labelsep=5.5mm]
        \item Definition: Disparate treatment due to membership of a social group or due to proxies associated with a social group.
        \item LLM-based hiring tools: Resume and cover letter screening.
        \item LLM-based healthcare tools: AI diagnostics.
    \end{itemize}
\end{itemize}

Allocational harms: Social biases preserved by LLMs can translate into discriminatory allocation of opportunities and services—for example, by steering who gets shortlisted for jobs or which patients receive particular diagnostic workups—thereby creating disparate treatment or disparate impact in high-stakes settings.

In addition to ethical concerns, allocation harms intersect with legal and regulatory frameworks intended to prevent discrimination. In the U.S., federal employment discrimination protections apply to AI and other automated tools ``just as they apply to other employment practices'', including recruitment and screening; employers remain responsible for ensuring that selection procedures do not unlawfully exclude protected groups, even when tools are provided by vendors \cite{eeoc_guidance2023}. At the subnational level, New York City's Local Law 144 regulates ``automated employment decision tools'' by requiring recent bias audits, public summaries, and advance notice before use in hiring or promotion decisions~\cite{nyc_local144}. In healthcare, the 2024 Section 1557 final rule prohibits discrimination through the use of patient care decision support tools and requires covered entities to make reasonable efforts to identify high-risk uses and mitigate discrimination when such tools are used in clinical decision making \cite{section1557_2024}. In the EU, the AI Act adopts a risk-based approach: ``high risk'' includes recruitment tools (e.g., CV sorting) and AI-based medical software, with obligations such as risk management, data quality to reduce discriminatory outcomes, documentation, and human oversight \cite{eu_ai_act}.

{Despite these legal and regulatory frameworks, real-world examples demonstrate that allocation harms persist.} \citet{ferrara2023should} argues that AI-automated tools may preserve and exacerbate societal biases embedded in the training data. LLM-aided resume-analyzing tools may disqualify people based on non-traditional backgrounds instead of job-related qualifications. In 2018, Amazon reportedly took an internal AI recruiting tool off the shelf after being accused of bias against women since the resumes in the training data are predominantly from white males. Another example is that LLM-aided healthcare tools may generate inaccurate predictions and treatment recommendations for under-represented groups, potentially influencing healthcare decision making and resulting in unfair medical resource allocation. \citet{zack2024assessing} discovered that GPT-4 is less likely to recommend more expensive diagnostic procedures to Black people and more likely to predict Hispanic women are more likely to hide an alcohol-abuse history than Asian women.

\section{Conclusions}

This review has provided a comprehensive analysis of bias in LLMs, examining their sources, manifestations, evaluation methods, and mitigation strategies. Our categorization of intrinsic and extrinsic biases offers a structured approach to understanding their impact on various NLP tasks \citep{doan2024fairness, gallegos2024bias}. We have evaluated a range of bias assessment techniques, from data-level analysis to human-involved methods, providing researchers with a diverse toolkit for bias evaluation \citep{liang2021towards}. Our examination of mitigation strategies, encompassing pre-model, intra-model, and post-model approaches, highlights both promising developments and areas that require further research \citep{ferrara2023should, liu2022quantifying}.

The ethical and legal implications of biased LLMs are substantial. The potential for both representational and allocational harms in real-world applications emphasizes the need for responsible AI development \citep{blodgett2020language, zack2024assessing}. As LLMs become increasingly integrated into diverse facets of society---from personal assistants to decision-making systems in critical sectors—it is imperative to address these biases to prevent unintended negative consequences.

Looking forward, key research directions include developing context-sensitive bias evaluation metrics that can identify subtle biases across diverse cultural settings \citep{wu2024auditing}, advancing causal inference techniques for more effective debiasing methods \citep{li2024steering}, and investigating the long-term societal impacts of biased LLMs in critical domains such as healthcare and criminal justice \citep{obermeyer2019dissecting}. Additionally, fostering interdisciplinary collaborations to address ethical challenges and develop adaptive regulatory frameworks \citep{gebru2021datasheets}, while optimizing the balance between model performance and fairness \citep{joniak-aizawa-2022-gender}, remains crucial.

By thoroughly examining the complexities of bias in LLMs and highlighting avenues for ongoing research and collaboration, we aim to contribute to the development of AI systems that are both technologically advanced and socially responsible. Addressing these challenges is essential to harness the full potential of LLMs, fostering innovations that advance technology while upholding the principles of fairness and equality.

While much of the foundational work on LLM bias has focused on earlier models, such as GPT-2, GPT-3, and BERT, a critical question is whether the observed bias trends remain, are diminished, or have evolved in more recent, state-of-the-art models such as GPT-4, Llama 3, and Claude. Preliminary evidence suggests a mixed picture. On one hand, newer models benefit from more extensive alignment procedures, including reinforcement learning from human feedback (RLHF), constitutional AI training, and red-teaming exercises, which can reduce overt toxic outputs and explicit stereotyping. For instance, \cite{zack2024assessing} found that GPT-4 exhibited more subtle, but still measurable, racial and gender biases in healthcare recommendations, suggesting that while the most egregious biases may be reduced, more nuanced forms persist. On the other hand, the increased scale and capability of newer models can introduce new challenges: larger context windows may amplify biases over extended interactions, and the ability to generate more fluent and persuasive text may make biased outputs harder to detect. Open-source models such as Llama 3 present additional considerations, as community fine-tuning may introduce or remove biases in ways that are difficult to track or audit. The rapid pace of model development underscores the need for continuous, longitudinal bias evaluation rather than one-time assessments, and for evaluation benchmarks that evolve alongside model capabilities.

\vspace{6pt}

\section*{Funding}
This research received no external funding.

\section*{Data Availability}
Data availability is not applicable to this review article.

\section*{Conflicts of Interest}
The authors declare no conflicts of interest.

\appendix
\section{Examples of Extrinsic Biases}\label{appd_biases}
\subsection{Natural Language Understanding (NLU) Tasks}
NLU encompasses a broad range of tasks that aim to improve comprehension of input sequences \citep{chang2024survey}. It seeks to grasp the deeper connotations and implications inherent in human communication, focusing on what human language signifies beyond the literal interpretation of individual words.

\subsubsection{Coreference Resolution}
Coreference resolution in the context of LLMs involves identifying instances where different expressions within a text refer to the same entity \citep{lee2017end}. This task is crucial for accurately interpreting the meaning of sentences, especially in cases where pronouns, names, or other referential expressions are used. The primary goal of coreference resolution is to correctly link pronouns like ``he'', ``she'', or ``it'' and definite descriptions like ``the CEO'' to the appropriate entity mentioned earlier in the text.

In coreference resolution tasks, extrinsic bias frequently arises when the model’s ability to correctly identify and link references (such as pronouns or entity names) to the appropriate entities is affected by external factors unrelated to the intrinsic nature of the text itself.

\begin{itemize}
    \item \textbf{{Gender Bias} 
}
    \begin{itemize}[leftmargin=8.5mm,labelsep=5.5mm]
        \item \textbf{{Stereotypical occupation associations: }}In a sentence like ``The doctor finished the surgery, and she went to check on the patient'', a biased model might mistakenly link ``she'' to a nurse or another female figure, based on the stereotype that doctors are more likely to be male. Similarly, in ``The nurse finished the shift, and he went home'', the model might incorrectly associate ``he'' with a doctor or a male figure rather than the nurse, reflecting a bias that nurses are more likely to be female \citep{zhao2018gender}.
        
        \item \textbf{{Gendered pronoun resolution: }}In a sentence like ``Sam loves cooking. He is very talented in the kitchen'', a biased model might incorrectly resolve ``he'' to assume that Sam is male, even though the name ``Sam'' can be gender-neutral. This reflects a bias that cooking, when associated with talent or professionalism, is more likely to be linked to males, while in other contexts, the same task might be associated with females \citep{rudinger2018gender}. 
    \end{itemize}
    
    \item \textbf{{Age Bias}}
    \begin{itemize}[leftmargin=8.5mm,labelsep=5.5mm]
        \item \textbf{{Assumptions about technological proficiency:} }In a sentence like ``The senior programmer and the young intern were debugging the code. He quickly found the bug'', a biased model might incorrectly resolve ``He'' to refer to the senior programmer, based on the stereotype that senior individuals are more skilled with technology, even though both the senior programmer and the young intern are equally likely candidates \citep{hovy2021importance}.
        \item \textbf{{Bias in linking pronouns to age-related roles: }}In a sentence like ``The retiree and the young employee discussed the new policies. He suggested some changes'', a biased model might incorrectly resolve ``he'' to refer to the young employee, based on the stereotype that younger individuals are more likely to suggest changes or new ideas, whereas retirees might be stereotypically viewed as less active in professional settings \citep{garg2018word}.
    \end{itemize}
    
    \item \textbf{{Cultural or Regional Bias}}
    \begin{itemize}[leftmargin=8.5mm,labelsep=5.5mm]
        \item \textbf{{Regional variants and pronoun use: }}In different cultural or regional contexts, the use of pronouns can vary significantly. For instance, in some languages, pronouns might be omitted altogether (pro-drop languages), while in others, they are used extensively. A coreference resolution system trained predominantly on non-pro-drop language data might struggle with correctly resolving entities in pro-drop languages, where subjects are often implied rather than explicitly mentioned \citep{hovy2021importance}.
        \item \textbf{{Cultural context in family roles:} }In some cultures, family roles are strictly defined, with certain responsibilities typically assigned to specific genders or ages. A coreference resolution model might incorrectly resolve pronouns based on these cultural stereotypes. For example, in a sentence like ``The eldest son took care of his siblings. She prepared dinner'', a biased model might incorrectly resolve ``She'' to the mother rather than the eldest daughter, based on cultural assumptions about family roles \citep{rudinger2018gender}.
    \end{itemize}
    
\end{itemize}

\textbf{{Mitigation recommendation for multilingual coreference bias:}} To address biases arising from pronoun omission in pro-drop languages (e.g., Chinese, Japanese, Korean, Spanish, and Arabic), we recommend training coreference resolution systems on balanced multilingual datasets that explicitly account for these grammatical peculiarities. Specifically, training data should include a substantial proportion of pro-drop language examples with gold-standard coreference annotations, so that models learn to infer referents from contextual cues rather than relying on explicit pronoun presence. Data augmentation strategies such as systematically generating pro-drop variants of existing English coreference datasets, or incorporating parallel corpora where pro-drop and non-pro-drop translations are aligned, can help models generalize across typologically diverse languages. Furthermore, evaluation benchmarks for coreference resolution should include dedicated test sets for pro-drop languages to ensure that system performance is assessed across this critical dimension of linguistic variation.

\subsubsection{Semantic Textual Similarity (STS)}
Semantic textual similarity (STS) tasks in LLMs focus on evaluating how similar the meanings of two pieces of text are. The purpose of STS is to determine how closely related the meanings of two text snippets are, which can vary from complete equivalence in meaning to being entirely unrelated. STS tasks are crucial for various applications where understanding the nuanced differences or similarities between texts is essential \citep{cer2017semeval}.

Extrinsic biases in STS tasks refer to biases that cause the model to inaccurately assess the similarity between two pieces of text. These biases can lead to skewed results, where the model might overestimate or underestimate the similarity between texts based on these biases rather than their actual semantic content. 

\begin{itemize}
    \item \textbf{{Gender Bias}}
    \begin{itemize}[leftmargin=8.5mm,labelsep=5.5mm]
        \item \textbf{{Gendered language and pronoun resolution:} }When comparing the similarity between sentences with gendered pronouns, such as: sentence A: ``She is a leader.''; sentence B: ``He is a leader''. A biased model might incorrectly rate the similarity between these sentences lower than it should, due to underlying gender associations, despite the fact that the sentences are nearly identical except for the pronoun \citep{zhao2017men}.
        \item \textbf{{Gender-stereotyped professions:} }Consider two sentences: sentence A: ``The nurse administered the medication.''; sentence B: ``She took care of the patient.'' If an LLM trained with gender biases in its data associates nursing primarily with women, it might assess a higher similarity between these sentences compared to another pair where the nurse is referred to as ``he'', even though the gender of the nurse should not affect the similarity score \citep{bolukbasi2016man}.
    \end{itemize}
    
    \item \textbf{{Age Bias}}
    \begin{itemize}[leftmargin=8.5mm,labelsep=5.5mm]
        \item \textbf{{Age-based expectations in language use:} }The sentences ``She is full of energy and enthusiasm'' and ``She is calm and experienced'' might be rated as less similar if the first sentence is associated with a young person and the second with an older person, despite both sentences describing positive qualities. A biased model might reinforce the stereotype that energy is associated with youth and calmness with older age, skewing the similarity score \citep{caliskan2017semantics}.
        \item \textbf{{Age stereotypes in sentiment and perception:} }Sentences like ``She is wise'' and ``She is elderly'' might be rated as similar by a biased model due to the stereotype that elderly people are wise. This could lead to an overestimation of the similarity between texts that describe older individuals, even if the context suggests different meanings \citep{diaz2016query}.
    \end{itemize}
    
    \item \textbf{{Cultural or Regional Bias}}
    \begin{itemize}[leftmargin=8.5mm,labelsep=5.5mm]
        \item \textbf{{Cultural idioms and expressions: }}Consider the sentences ``It's raining cats and dogs'' (an English idiom meaning heavy rain) and ``Il pleut des cordes'' (a French idiom meaning the same). A culturally biased model trained predominantly on English data might rate these sentences as less similar because the literal words are different, even though they convey the same meaning in their respective cultures \citep{vulic2013cross}.
        \item \textbf{{Regional dialects and variants:} }In English, the word ``apartment'' is commonly used in American English, while ``flat'' is used in British English. A model trained predominantly on one variant might incorrectly assess the similarity between ``I live in an apartment'' and ``I live in a flat'' as low, due to not recognizing the regional synonymy \citep{tan2011building}.
    \end{itemize}
    
\end{itemize}

\subsubsection{Natural Language Inference}
Natural language inference (NLI) is a task where the goal is to determine the relationship between a pair of sentences, typically referred to as the ``premise'' and the ``hypothesis''. The task involves classifying the relationship between these sentences into one of three categories: entailment (the hypothesis logically follows from the premise), contradiction (the hypothesis contradicts the premise), or neutral (the hypothesis is neither supported nor contradicted by the premise; it could be true or false based on the premise alone) \citep{bowman2015large}. This is a fundamental task in understanding and reasoning with natural language and is often used to evaluate the reasoning capabilities of LLMs.

Extrinsic bias in the NLI task using LLMs refers to biases introduced from external factors that affect the model’s ability to accurately determine the relationship between a premise and a hypothesis. In NLI, the task is to decide whether a given hypothesis is entailed by, contradicted by, or neutral with respect to the premise. External biases can influence how well the model performs this task, leading to skewed or unfair results~\citep{maccartney2009natural, dev2020measuring}.

\begin{itemize}
    \item \textbf{{Gender Bias}}
    \begin{itemize}[leftmargin=8.5mm,labelsep=5.5mm]
        \item \textbf{{Stereotypical gender roles in professions: }}Given a premise like ``The person is a doctor'', and a hypothesis ``She is caring'', a biased model might be more likely to infer that the hypothesis is true because of the stereotype associating women with caring professions. Conversely, if the premise is ``The person is an engineer'', and the hypothesis is ``She is analytical'', the model might incorrectly infer this as less likely due to the stereotype that engineering is male-dominated~\citep{rudinger2018gender}.
        \item \textbf{{Gendered language and assumptions: }}In a scenario where the premise is ``Alex received a promotion'', and the hypothesis is ``She worked very hard'', a biased model might rate this inference as less likely due to the name ``Alex'', which can be gender-neutral but may be more commonly associated with males. The model may incorrectly prefer ``He worked very hard'' based on gender assumptions~\citep{zhao2017men}.
    \end{itemize}
    
    \item \textbf{{Age Bias}}
    \begin{itemize}[leftmargin=8.5mm,labelsep=5.5mm]
        \item \textbf{{Bias toward younger individuals in dynamic roles: }} Given the premise ``The person is a dynamic entrepreneur'', and the hypothesis ``The person is young'', a biased model might overly favor the hypothesis due to the stereotype that entrepreneurship and dynamism are traits associated with younger people, disregarding the possibility that older individuals can also embody these characteristics \citep{caliskan2017semantics}.
        \item \textbf{{Age-related stereotypes in professional contexts:} } If the premise is ``The CEO announced the company's new strategic direction'', a biased model might infer that ``The CEO is in their 50s'' is more likely to be true (entailment), reflecting the stereotype that leadership roles are typically held by older individuals. On the other hand, it might incorrectly classify ``The CEO is in their 20s'' as a contradiction \citep{zhao2018gender}.
    \end{itemize}
    
    \item \textbf{{Cultural or Regional Bias}}
    \begin{itemize}[leftmargin=8.5mm,labelsep=5.5mm]
        \item \textbf{{Cultural norms and social roles:} }Consider the premise ``She is a nurse'' and the hypothesis ``She is caring''. A model influenced by cultural stereotypes might incorrectly infer that the hypothesis is true based on the cultural assumption that nurses are inherently caring, which reflects a bias rather than a logical inference based on the text \citep{rudinger2018gender}.
        \item \textbf{{Regional political and historical contexts:} }A premise might state, ``The Berlin Wall fell in 1989'', with a hypothesis, ``Germany was reunified shortly after''. A model trained predominantly on Western narratives might correctly infer entailment, but could struggle with similar inferences in different cultural or geopolitical contexts where regional historical knowledge is less emphasized~\citep{chen2016enhanced}.
    \end{itemize}
    
\end{itemize}

\subsubsection{Classification}
A classification task in the context of LLMs refers to the process of assigning predefined categories or labels to a given input text based on its content. The objective is to automatically identify the appropriate class or category to which the text belongs, leveraging the patterns and knowledge learned by the language model \citep{de2019bias, meng2020text}.

Bias can be understood as the variation in classification outcomes for texts involving different values of sensitive attributes (e.g., gender). An unbiased model should have similar classification outcomes between different social groups.

\begin{itemize}
    \item \textbf{{Gender Bias}}
    \begin{itemize}[leftmargin=8.5mm,labelsep=5.5mm]
        \item \textbf{{Gendered language in job title classification: }}When classifying job-related text, an LLM might associate certain professions with specific genders based on stereotypical norms. For instance, ``nurse'' might be more frequently classified as female, while ``engineer'' might be classified as male, even when the text does not explicitly mention gender. This bias can lead to skewed recommendations in hiring algorithms or job description parsing \citep{bolukbasi2016man}.
        \item \textbf{{Gendered pronouns and name classification:} }When classifying names or pronouns, LLMs may exhibit bias by associating certain names or pronouns with specific gendered stereotypes. For instance, names like ``Alex'' or ``Jordan'' might be classified with a gendered label (male or female) based on historical or cultural associations rather than the context provided \citep{zhao2017men}.
    \end{itemize}
    
    \item \textbf{{Age Bias}}
    \begin{itemize}[leftmargin=8.5mm,labelsep=5.5mm]
        \item \textbf{{Sentiment classification bias based on age: }}A model might classify text written by older individuals as more negative or less enthusiastic compared to text written by younger individuals. For instance, if an older person discusses new technology, the model might incorrectly classify the sentiment as negative or apprehensive, based on the stereotype that older people are less tech-savvy or more resistant to change \citep{diaz2018addressing}.
        \item \textbf{{Bias in job application classification:} }In job application screening, an LLM might classify older applicants as less suitable for certain roles based on age-related stereotypes. For example, the model might down-rank resumes of older applicants for tech-related jobs due to biases that associate youth with innovation and adaptability, while assuming older candidates are less capable of learning new skills \citep{de2019bias}.
    \end{itemize}
    
    \item \textbf{{Cultural or Regional Bias}}
    \begin{itemize}[leftmargin=8.5mm,labelsep=5.5mm]
        \item \textbf{{Misclassification due to cultural language variants: }}A text classification model might incorrectly classify content written in African American Vernacular English (AAVE) as informal, unprofessional, or even toxic, due to biases against non-standard English dialects. This can result in discriminatory outcomes, such as the unfair flagging of social media posts by Black users \citep{sap2019risk}.
        \item \textbf{{Regional bias in political text classification: }}A political text classifier might be biased toward the dominant political ideology of the region where the training data was sourced. For instance, texts supporting socialist policies might be classified as ``radical'' or ``extreme'' if the training data predominantly reflects a region where socialism is less accepted \citep{bender2021dangers}.
    \end{itemize}
    
\end{itemize}

\subsubsection{Reading Comprehension}
Reading comprehension in the context of LLMs involves providing the model with a passage of text and asking it to answer questions about that passage. The objective is to evaluate the model's ability to understand and interpret the content of the passage, including facts, concepts, and inferred meanings. The task evaluates the model's capability to extract relevant information, make inferences, and provide accurate and contextually appropriate answers based on the text provided \citep{rajpurkar2018know}.

Extrinsic biases in reading comprehension tasks within LLMs arise when the model's ability to understand and answer questions based on a given text is influenced by external biases embedded in the training data. These biases can lead to incorrect or biased models favoring certain interpretations of the text, making incorrect assumptions, or resulting in skewed answers that reflect prejudiced views, rather than a neutral understanding of the text.

\begin{itemize}
    \item \textbf{{Gender Bias}}
    \begin{itemize}[leftmargin=8.5mm,labelsep=5.5mm]
        \item \textbf{{Stereotypes in gendered activities: }}In a scenario where the passage mentions, ``Alex is an excellent cook'', a biased model might assume that Alex is female and reflect this in its answers to questions about the passage. For example, when asked, ``What are her skills?'' the model might incorrectly use a gendered pronoun, despite ``Alex'' being a gender-neutral name, thereby reflecting gender stereotypes associated with cooking \citep{webster2018mind}.
        \item \textbf{{Assumptions about gender roles in family settings:} }A passage might describe both a mother and a father taking care of children, but when asked, ``Who is responsible for preparing dinner?'' a biased model might infer that the mother is responsible, reflecting traditional gender roles \citep{caliskan2017semantics}.
    \end{itemize}
    
    \item \textbf{{Age Bias}}
    \begin{itemize}[leftmargin=8.5mm,labelsep=5.5mm]
        \item \textbf{{Bias in health-related content: }}In passages discussing health, a model might incorrectly infer that certain conditions or concerns (e.g., memory loss, frailty) are more likely associated with older characters, even when the text is neutral or unrelated to age. For instance, a passage about a character feeling tired might lead the model to infer an age-related cause if the character is older \citep{bolukbasi2016man}.
        \item \textbf{{Misinterpretation of age-related roles: }}In a passage describing a family scenario, a reading comprehension model might incorrectly assume that the younger character is less responsible or that the older character is in a caregiving role, even if the text does not support these assumptions. This could lead to biased answers to questions about the characters' roles or behaviors \citep{bender2018data}.
    \end{itemize}
    
    \item \textbf{{Cultural or Regional Bias}}
    \begin{itemize}[leftmargin=8.5mm,labelsep=5.5mm]
        \item \textbf{{Cultural context misinterpretation:} }A reading comprehension model might misinterpret a text that involves culturally specific practices or norms. For example, if a passage describes a traditional Japanese tea ceremony, a model trained predominantly on Western texts might misunderstand the significance of the ceremony, interpreting it as a simple social gathering rather than a ritual with deep cultural meaning \citep{bender2018data}.
        \item \textbf{{Regional language varieties and dialects:} }A reading comprehension model might struggle with understanding regional dialects or non-standard language varieties. For instance, a passage written in African American Vernacular English (AAVE) might be misinterpreted or considered less coherent by the model, leading to incorrect or biased responses to comprehension questions \citep{blodgett2020language}.
    \end{itemize}
    
\end{itemize}

\subsubsection{Sentiment Analysis}
Sentiment analysis in the context of LLMs involves identifying the sentiment or emotional tone expressed in a piece of text. This task typically categorizes text as having a positive, negative, or neutral sentiment, although more nuanced categorizations (such as specific emotions like happiness, anger, or sadness) can also be used. Sentiment analysis is widely applied in areas such as social media monitoring, customer feedback analysis, and product reviews \citep{pang2008opinion}.

Extrinsic biases in sentiment analysis tasks within LLMs refer to the unintended and often unfair influences that external biases embedded in the training data have on the model's ability to accurately interpret the sentiment (whether positive, negative, or neutral) expressed in text. These biases can cause the model to systematically misclassify sentiment based on factors like the writer's gender, race, age, or other demographic characteristics, leading to skewed or unfair sentiment assessments.

\begin{itemize}
    \item \textbf{{Gender Bias}}
    \begin{itemize}[leftmargin=8.5mm,labelsep=5.5mm]
        \item \textbf{{Bias in sentiment toward gendered products or topics:} }Sentiment analysis models might exhibit bias when analyzing reviews or discussions about gendered products. For example, reviews of products marketed toward women, such as beauty products, might be categorized with more extreme sentiment (either overly positive or negative) compared to neutral categorizations for products marketed toward men \citep{zhao2017men}.
        \item \textbf{{Sentiment analysis in gendered contexts:} }Sentiment analysis models may misinterpret the sentiment of sentences involving gendered contexts. For example, statements about women's rights or feminism might be more likely to be categorized as negative, reflecting a bias against topics that challenge traditional gender norms \citep{misiunas2019density}.
    \end{itemize}
    
    \item \textbf{{Age Bias}}
    \begin{itemize}[leftmargin=8.5mm,labelsep=5.5mm]
        \item \textbf{{Sentiment analysis on age-related topics:} }When analyzing sentiment on topics related to aging, retirement, or health, sentiment analysis models might display bias by assuming a more negative sentiment in texts written by older individuals. For example, discussions about retirement might be categorized as negative due to societal biases about aging, even if the text is neutral or positive in tone \citep{burnap2015cyber}.
        \item \textbf{{Stereotyping language use by older adults:} }Sentiment analysis models might misinterpret text written by older adults as more negative or less enthusiastic due to stereotypical views that older individuals are more conservative or less expressive. For instance, a review from an older person saying ``The movie was good'' might be rated as less positive compared to a more exuberant review from a younger person, even though both reviews express positive sentiment \citep{hovy2015tagging}.
    \end{itemize}
    
    \item \textbf{{Cultural or Regional Bias}}
    \begin{itemize}[leftmargin=8.5mm,labelsep=5.5mm]
        \item \textbf{{Cultural bias in sentiment toward social norms:} }Sentiment analysis might misinterpret text related to social norms differently across cultures. For instance, in some cultures, discussing money openly is seen as positive and associated with success, while in others, it might be considered impolite or negative. A model trained on Western data might categorize open discussions about money in a non-Western text as negative \citep{blodgett2017racial}.
        \item \textbf{{Sentiment in multilingual contexts:} }In multilingual regions, a sentiment analysis model might incorrectly categorize sentiment when it encounters code-switching (the practice of alternating between two or more languages or dialects). For instance, in Latin America, code-switching between Spanish and indigenous languages might lead to incorrect sentiment assessments if the model is biased towards Spanish and fails to interpret the sentiment conveyed in the indigenous language correctly \citep{hovy2015tagging}.
    \end{itemize}
    
\end{itemize}

\subsection{Natural Language Generation (NLG) Tasks}
NLG tasks refer to downstream tasks in which the model generates coherent, contextually relevant, and human-like text based on a given input or set of instructions \citep{chang2024survey}. The objective of NLG is to produce text that is not only grammatically correct but also semantically meaningful and appropriate for the context in which it is generated.

\subsubsection{Question Answering}
Question answering (QA) tasks in the context of LLMs focus on the model's ability to provide accurate and relevant answers to questions based on a given text or knowledge base. QA is an essential task in the domain of human--computer interaction, and it has been extensively used in various situations such as search engines and chatbots \citep{zadeh2006search, li2024citation}. The goal of QA is to extract, infer, or generate responses that directly address the query, leveraging the model's understanding of language, context, and content \citep{rajpurkar2016squad}.

Extrinsic biases in QA tasks occur when the model's responses are influenced by external biases embedded in the training data on the model’s ability to provide accurate and fair answers to questions. These biases can lead to skewed, misleading, unfair, inaccurate, or discriminatory answers based on gender, race, age, culture, or other demographic factors, rather than providing a neutral and contextually appropriate response.

\begin{itemize}
    \item \textbf{{Gender Bias}}
    \begin{itemize}[leftmargin=8.5mm,labelsep=5.5mm]
        \item \textbf{{Gender bias in answer generation:} }When asked, ``What should a good leader do?'', the model might provide examples or language that align with stereotypically male attributes (e.g., ``He should be assertive and decisive''), thereby implying that leadership qualities are inherently male. If asked about nurturing roles, the model might default to female pronouns and qualities, reinforcing gendered stereotypes \citep{zhao2018gender}.
        \item \textbf{{Bias in answering ambiguous gender questions: }}For a question like ``What does a typical manager do?'' where gender is not specified, a biased model might generate an answer using male pronouns (``He manages the team...''), reflecting an assumption that managers are typically male. This bias can also appear in reverse, where certain roles like teaching or nursing might default to female pronouns \citep{webster2018mind}.
    \end{itemize}
    
    \item \textbf{{Age Bias}}
    \begin{itemize}[leftmargin=8.5mm,labelsep=5.5mm]
        \item \textbf{{Stereotypical answers about aging: }}If asked, ``What are the best activities for elderly people?'' a model might focus on sedentary activities like knitting or watching TV, neglecting more active pursuits like hiking, traveling, or volunteering. This reflects a bias that older adults are less capable of engaging in physically demanding or adventurous activities \citep{caliskan2017semantics}.
        \item \textbf{{Negative bias toward youth: }}If a question is posed like, ``Are young people capable of managing a company?'' a biased model might respond with, ``They may lack the experience needed for such a role'', reflecting stereotypes that associate youth with inexperience, despite many young people successfully managing companies \citep{bolukbasi2016man}.
    \end{itemize}
    
    \item \textbf{{Cultural or Regional Bias}}
    \begin{itemize}[leftmargin=8.5mm,labelsep=5.5mm]
        \item \textbf{{Bias in answering culturally specific questions: }}When asked, ``What is the most popular sport in the world?'' a biased QA system might answer ``American football'' if trained predominantly on data from the United States, ignoring that globally, soccer (football) is more widely popular. This reflects a bias toward regional popularity rather than global knowledge \citep{gardner2018allennlp}.
        \item \textbf{{Language and regional bias in answer accuracy: }}A QA system might perform better when answering questions in or about regions and languages that are well-represented in its training data. For example, questions about European history or in English might be answered more accurately than questions about African history or in less commonly spoken languages \citep{kwiatkowski2019natural}.
    \end{itemize}
    
\end{itemize}

\subsubsection{Sentence Completions}
Sentence completion tasks in the context of LLMs involve predicting and generating the next word or sequence of words to complete a partially provided sentence. The goal of these tasks is to produce a grammatically correct and contextually appropriate continuation that aligns with the preceding text. Sentence completion is a fundamental capability in many NLP applications, such as autocomplete features and writing assistants \citep{radford2019language}.

Extrinsic biases in sentence completion tasks within LLMs occur when the model's predictions for completing a sentence are influenced by external biases embedded in the training data on the model’s ability to generate appropriate and unbiased sentence continuations. This type of bias can cause the model to complete sentences in ways that reinforce harmful stereotypes, perpetuate prejudice, or reflect biased assumptions about gender, race, age, culture, or other demographic factors.

\begin{itemize}
    \item \textbf{{Gender Bias}}
    \begin{itemize}[leftmargin=8.5mm,labelsep=5.5mm]
        \item \textbf{{Gender bias in descriptions of physical appearance: }}If the sentence starts with ``She looked at herself in the mirror and...'' a biased model might complete it with ``...adjusted her makeup'', whereas ``He looked at himself in the mirror and...'' might be completed with ``...straightened his tie''. This reflects the stereotype that women are more concerned with appearance and men with professionalism~\citep{caliskan2017semantics}.
        \item \textbf{{Gender bias in personal attributes: }}If prompted with ``She is very...'' a model might complete the sentence with adjectives like ``emotional'', ``beautiful'', or ``caring'', while ``He is very...'' might be completed with ``strong'', ``intelligent'', or ``ambitious''. These completions reflect stereotypical views of women as being more emotional and men as being more rational or powerful \citep{zhao2017men}.
    \end{itemize}
    
    \item \textbf{{Age Bias}}
    \begin{itemize}[leftmargin=8.5mm,labelsep=5.5mm]
        \item \textbf{{Activity and lifestyle assumptions:} }For a prompt like ``At 25, Jenny enjoys'', a biased model might complete the sentence with ``partying and going out every night'', based on the stereotype that young adults are primarily interested in nightlife and socializing, ignoring the diversity of interests in this age group \citep{garg2018word}.
        \item \textbf{{Learning and education stereotypes: }}Given the prompt ``At 50, Mark decided to'', a biased model might complete it with ``go back to school to finally get his degree'', assuming that older adults are ``catching up'' on education rather than pursuing lifelong learning or new academic challenges \citep{sun2019mitigating}.
    \end{itemize}
    
    \item \textbf{{Cultural or Regional Bias}}
    \begin{itemize}[leftmargin=8.5mm,labelsep=5.5mm]
        \item \textbf{{Cultural stereotyping in sentence completion: }}If the model is given a sentence like ``In Japan, people often eat'', it might complete with ``sushi'', reflecting a cultural stereotype that overemphasizes a specific aspect of Japanese cuisine, while ignoring the diversity of food in Japanese culture. Similarly, completing ``In Mexico, people celebrate'' with ``Cinco de Mayo'' might reinforce a limited and stereotypical understanding of Mexican culture \citep{hendricks2018women}.
        \item \textbf{{Regional bias in place-based completions:} }For the sentence ``In Africa, many people live in'', the model might complete with ``villages'', reflecting a regional bias that assumes rural living conditions are more common across an entire continent, despite the presence of large urban areas. This completion overlooks the diversity of living environments in different African countries and regions~\citep{birhane2021large}.
    \end{itemize}
    
\end{itemize}

\subsubsection{Conversational}
Conversational tasks in the context of LLMs involve generating, understanding, and maintaining coherent dialogue with users. These tasks are crucial for applications like chatbots, virtual assistants, and customer service automation. Conversational tasks typically require the model to not only generate contextually appropriate responses but also to track the context of the conversation, understand user intent, manage the dialogue flow, and sometimes incorporate external knowledge or personalized information \citep{zhang2019dialogpt}.

Extrinsic biases in conversational tasks within LLMs refer to the unintended influence of external biases on the model's responses during a conversation. These biases can cause the model to generate responses that reflect prejudiced views, reinforce stereotypes, or unfairly favor certain perspectives, leading to skewed, insensitive, or inappropriate conversational outcomes, which may negatively impact user experience.

\begin{itemize}
    \item \textbf{{Gender Bias}}
    \begin{itemize}[leftmargin=8.5mm,labelsep=5.5mm]
        \item \textbf{{Stereotypical responses based on gendered prompts:} }When given a prompt like ``Describe a nurse'' versus ``Describe a doctor'', an LLM might generate responses that reinforce gender stereotypes. For example, it might describe a nurse as ``caring, nurturing, and female'', while describing a doctor as ``authoritative, knowledgeable, and male'', despite no gender being specified in the prompts~\citep{bolukbasi2016man}.
        \item \textbf{{Bias in gendered interactions:} }In a customer service chatbot, the model might respond more politely or deferentially to queries assumed to be from female users based on gender cues in the text (like names or pronouns), while responding more assertively or formally to male users. This reflects gendered expectations of communication styles \citep{hitsch2010matching}.
    \end{itemize}
    
    \item \textbf{{Age Bias}}
    \begin{itemize}[leftmargin=8.5mm,labelsep=5.5mm]
        \item \textbf{{Assumptions about being tech-savvy: }}In a conversation where the user is discussing technology, a biased model might assume the user is younger if they express familiarity with tech jargon or concepts. Conversely, if the user asks for help with basic technology, the model might assume the user is older and potentially provide overly simplistic explanations, which could be patronizing~\citep{hovy2015tagging}.
        \item \textbf{{Bias in addressing age-related concerns:} }In a scenario where a user asks for advice on starting a new career later in life, a biased model might discourage them by emphasizing the challenges of changing careers at an older age. For instance, if the user says, ``I’m thinking of switching careers at 50'', the model might respond with comments like, ``It might be difficult at your age'', reflecting a bias that older individuals face more barriers in the job market \citep{wagner2016women}.
    \end{itemize}
    
    \item \textbf{{Cultural or Regional Bias}}
    \begin{itemize}[leftmargin=8.5mm,labelsep=5.5mm]
        \item \textbf{{Culturally inappropriate responses: }}A conversational LLM might provide responses that are culturally inappropriate or insensitive due to a lack of understanding of cultural norms. For example, if a user from Japan mentions ``visiting a shrine'', a culturally biased model might respond with a suggestion that is more aligned with Western religious practices, failing to acknowledge the cultural significance of shrines in Japanese culture \citep{hovy2016social}.
        \item \textbf{{Bias in handling regional topics:} }A conversational AI might be biased in how it handles topics related to certain regions. For instance, if asked about news in Africa, the model might focus disproportionately on negative topics like conflict or poverty, reflecting a bias in the training data, while similar questions about Europe might elicit more varied and positive topics \citep{gebru2021datasheets}.
    \end{itemize}
    
\end{itemize}

\subsubsection{Recommender Systems}
Recommender systems in the context of LLMs are designed to predict and suggest personalized content or items of interest to users based on their preferences, behavior, or contextual information. These systems analyze patterns in user data, such as past interactions, preferences, or demographic information, to provide personalized recommendations, such as suggesting movies, products, books, or articles that the user might enjoy or find useful \citep{zhang2019deep, hua2023up5}.

Extrinsic biases in recommender systems within LLMs occur when the model's recommendations are influenced by external biases present in the training data. This bias can result in the model making recommendations that disproportionately favor certain groups, products, or content types over others based on factors like gender, race, age, culture, or other demographic characteristics, rather than providing neutral and personalized suggestions.

\begin{itemize}
    \item \textbf{{Gender Bias}}
    \begin{itemize}[leftmargin=8.5mm,labelsep=5.5mm]
        \item \textbf{{Product recommendations based on gender stereotypes:} }A recommender system might suggest beauty products, fashion items, or household goods primarily to women, while recommending electronics, tools, or sports equipment predominantly to men. This bias reinforces traditional gender roles and can lead to irrelevant or unappealing recommendations for users whose interests do not align with these stereotypes \citep{ekstrand2018all}.
        \item \textbf{{Career and education recommendations: }}A job or education recommender system might suggest STEM (Science, Technology, Engineering, and Mathematics) careers more often to male users while recommending roles in healthcare, education, or the arts more frequently to female users. This bias can perpetuate gender disparities in career choices and educational opportunities \citep{chen2019correcting}.
    \end{itemize}
    
    \item \textbf{{Age Bias}}
    \begin{itemize}[leftmargin=8.5mm,labelsep=5.5mm]
        \item \textbf{{Age-related product recommendations:} }A recommender system might automatically suggest health-related products, such as supplements or exercise equipment, to older users while recommending tech gadgets or trendy fashion items to younger users. These recommendations may not accurately reflect the individual’s interests but are based on age-related stereotypes \citep{biega2018equity}.
        \item \textbf{{Media and entertainment recommendations: }}A recommender system might assume that older users prefer classic movies or oldies music, while younger users prefer contemporary pop culture content. This can result in older users being recommended content that does not reflect their actual tastes if they have a preference for contemporary media, and younger users missing out on discovering older, classic content they might enjoy \citep{burke2018balanced}.
    \end{itemize}
    
    \item \textbf{{Cultural or Regional Bias}}
    \begin{itemize}[leftmargin=8.5mm,labelsep=5.5mm]
        \item \textbf{{Regional bias in news and information recommendations: }}A news recommender system might prioritize local or regional news from dominant cultures, under-representing or completely ignoring news from minority regions or less dominant cultures. For instance, users in a country might predominantly receive news recommendations about urban centers and mainstream political issues, while rural or indigenous issues are marginalized \citep{karimi2018news}.
        \item \textbf{{Bias in language and cultural content recommendations: }}A recommender system might prioritize content in the dominant language of a region, leading to a lack of recommendations for content in minority languages. For example, in a multilingual country, the system might recommend mostly English-language content, marginalizing content in local languages like Tamil, Welsh, or Basque~\citep{lakew2018multilingual}.
    \end{itemize}
    
\end{itemize}

\subsubsection{Machine Translation}
Machine translation (MT) tasks in the context of LLMs involve the automatic translation of text from one language to another. The aim is to produce a translation that accurately conveys the meaning of the original text while maintaining grammatical correctness, fluency, and cultural appropriateness in the target language. Machine translation systems are trained on large parallel corpora, which contain pairs of sentences in different languages that correspond to each other. These models learn patterns and structures in both languages to generate translations \citep{vaswani2017attention}.

Extrinsic biases in machine translation (MT) tasks within LLMs occur when the model's translations are influenced by external biases embedded in the training data. Such biases can lead to translations that reflect cultural, gender, racial, or regional stereotypes, resulting in unfair or inaccurate translations that either misrepresent the original text's meaning or that reinforce harmful stereotypes.

\begin{itemize}
    \item \textbf{{Gender Bias}}
    \begin{itemize}[leftmargin=8.5mm,labelsep=5.5mm]
        \item \textbf{{Gendered language mismatch:} }When translating from a gender-neutral language like Turkish or Finnish into a gendered language like English or Spanish, the model might introduce gender bias by assigning gendered pronouns or roles based on stereotypes. For example, the Turkish sentence ``O bir doktor'' (which means ``He/She is a doctor'') might be translated into English as ``He is a doctor'', reflecting the stereotype that doctors are male \citep{prates2020assessing}.
        \item \textbf{{Gender stereotyping in occupational translations: }}When translating sentences that involve professions, the model might incorrectly assign gendered pronouns based on stereotypes. For instance, translating the phrase ``The nurse'' from a gender-neutral language might result in ``La enfermera'' (female nurse) in Spanish, while ``The engineer'' might be translated as ``El ingeniero'' (male engineer), even if the original language did not specify gender \citep{neri2020design}.
    \end{itemize}
    
    \item \textbf{{Age Bias}}
    \begin{itemize}[leftmargin=8.5mm,labelsep=5.5mm]
        \item \textbf{{Bias in addressing older adults:} }A machine translation system might translate sentences in a way that condescends to older adults, reflecting societal biases. For example, translating ``The elderly person learned to use a smartphone'' might introduce a tone or wording in the target language that implies surprise or patronization, even if the original sentence was neutral \citep{caliskan2017semantics}.
        \item \textbf{{Translation of age-related idioms:} }When translating age-related idioms or expressions, a biased translation system might reinforce negative stereotypes. For instance, translating a phrase like ``old people are slow'' into another language might retain the negative connotation or even intensify it if the target language has a stronger cultural bias against the elderly~\citep{hovy2015tagging}.
    \end{itemize}
    
    \item \textbf{{Cultural or Regional Bias}}
    \begin{itemize}[leftmargin=8.5mm,labelsep=5.5mm]
        \item \textbf{{Cultural nuance loss:} }Cultural idioms, proverbs, or colloquial phrases often lose their meaning or are mistranslated when a model does not consider the cultural context. For example, the English idiom ``It's raining cats and dogs'' could be literally translated into another language, resulting in a confusing or meaningless phrase in that target culture. A culturally aware model would translate it into a local equivalent, such as ``Il pleut des cordes'' (It's raining ropes) in French \citep{vanmassenhove2019lost}.
        \item \textbf{{Bias toward dominant cultures:} }A machine translation system might favor translations that align with Western cultural norms over those of less dominant cultures. For instance, translating phrases related to food, clothing, or customs might reflect Western standards, even when the source text belongs to a non-Western culture. An example could be translating a traditional Chinese clothing item, ``Qipao'', simply as ``dress'', which dilutes the cultural significance~\citep{sennrich2016neural}.
    \end{itemize}
    
\end{itemize}

\subsubsection{Summarization}
Summarization in the context of LLMs involves generating a concise and coherent summary of a longer text while preserving the key information and main points. The goal of summarization is to shorten the text while preserving its essential meaning and context~\citep{nallapati2016abstractive}.

Extrinsic biases in summarization tasks within LLMs refer to biases that stem from external influences present in the training data on the way a model generates summaries. These biases can affect how the model selects and condenses information, potentially leading to skewed, unfair, or incomplete summaries that disproportionately emphasize or omit certain information based on gender, race, age, or other characteristics, rather than providing a neutral summary of the content.

\begin{itemize}
    \item \textbf{{Gender Bias}}
    \begin{itemize}[leftmargin=8.5mm,labelsep=5.5mm]
        \item \textbf{{Differential emphasis on roles: }}In summarizing biographies or obituaries, a model might emphasize traditional gender roles. For example, summaries of women's lives might focus more on their roles as wives and mothers, while summaries of men's lives might emphasize their careers or public achievements, even if the original text gave equal importance to both aspects \citep{bender2018data}.
        \item \textbf{{Selective emphasis on gendered information: }}In summarizing a news article, a biased model might disproportionately emphasize gender-specific details that are not central to the story. For instance, when summarizing an article about a successful entrepreneur, the model might overemphasize details about the entrepreneur's gender or appearance if the subject is female while focusing on achievements and business strategies if the subject is male. This kind of bias reinforces gender stereotypes and can skew the perceived importance of certain information \citep{otterbacher2017competent}.
    \end{itemize}
    
    \item \textbf{{Age Bias}}
    \begin{itemize}[leftmargin=8.5mm,labelsep=5.5mm]
        \item \textbf{{Bias in summarizing content for different age groups: }}When summarizing content aimed at different age groups, a model might simplify or alter the tone of content intended for younger audiences in a way that reflects condescension or a lack of complexity. For instance, a model might overly simplify a summary of educational content intended for teenagers, assuming they cannot handle complex information \citep{obermeyer2019dissecting}.
        \item \textbf{{Omission of contributions based on age: }}When summarizing a report or article that includes contributions from both younger and older individuals, a biased summarization model might disproportionately highlight the contributions of younger people while downplaying or omitting the contributions of older individuals. For example, in summarizing a collaborative project, the model might emphasize the innovative ideas of younger team members while neglecting the experience and insights provided by older members \citep{bender2018data}.
    \end{itemize}
    
    \item \textbf{{Cultural or Regional Bias}}
    \begin{itemize}[leftmargin=8.5mm,labelsep=5.5mm]
        \item \textbf{{Omission of culturally significant details: }}A summarization model might omit culturally significant details that are not widely understood in the model's primary training data. For instance, if summarizing a text about a traditional festival like Diwali, the model might focus on generic elements like ``a festival with lights'' while omitting important cultural and religious aspects such as the significance of the festival in Hinduism \citep{shen2017style}.
        \item \textbf{{Bias toward western narratives:} }When summarizing global news articles, a model might prioritize Western perspectives or narratives, underplaying or omitting the viewpoints from non-Western cultures. For example, in summarizing an article about a political conflict in the Middle East, the model might focus on the perspectives of Western governments and omit the perspectives of local populations \citep{perez2022models}.
    \end{itemize}
    
\end{itemize}

These examples illustrate how extrinsic biases in LLMs can impact the accuracy, fairness, and potential for discrimination, often reinforcing harmful stereotypes and societal biases. Extrinsic bias is a significant concern because it can lead to unfair and discriminatory outcomes in applications like hiring, lending, and content moderation. Mitigating extrinsic biases in LLMs requires a comprehensive strategy that includes the use of diverse, representative, and balanced training data, inclusive labeling practices, the implementation of debiasing techniques, and continuous evaluation to ensure fair and accurate language processing. 


\addcontentsline{toc}{section}{References}
\begin{thebibliography}{999}

\bibitem[Keskar et~al.(2019)Keskar, McCann, Varshney, Xiong, and
  Socher]{keskar2019ctrl}
Keskar, N.S.; McCann, B.; Varshney, L.R.; Xiong, C.; Socher, R.
\newblock CTRL: A Conditional Transformer Language Model for Controllable Generation.
\newblock In Proceedings of the 2019 Conference on Empirical
  Methods in Natural Language Processing, {Hong Kong, China,  3--7 November}  2019;  pp. 111--129.  

\bibitem[Yang et~al.(2020)Yang, Yuan, and Liu]{yang2020finbert}
Yang, Y.; Yuan, Y.; Liu, L.
\newblock FinBERT: A Pretrained Language Model for Financial Communications.
\newblock {\em arXiv } {\bf 2020}, arXiv:2006.08097.

\bibitem[Okonkwo and Ade-Ibijola(2023)]{okonkwo2023role}
Okonkwo, C.W.; Ade-Ibijola, A.
\newblock The role of artificial intelligence in education: Prospects and
  challenges.
\newblock {\em J. Educ. Technol. Soc.} {\bf 2023}, {\em
  26},~1--12.

\bibitem[Jiang et~al.(2020)Jiang, Yang, and Li]{jiang2020smart}
Jiang, L.; Yang, M.; Li, X.
\newblock Smart Music Player: User-Adaptive Music Recommendation System.
\newblock {\em IEEE Trans. Multimed.} {\bf 2020}, {\em 22},~666--675.

\bibitem[Brown et~al.(2020)Brown, Mann, Ryder, Subbiah, Kaplan, Dhariwal,
  Neelakantan, Shyam, Sastry, Askell, et~al.]{brown2020language}
Brown, T.B.; Mann, B.; Ryder, N.; Subbiah, M.; Kaplan, J.; Dhariwal, P.;
  Neelakantan, A.; Shyam, P.; Sastry, G.; Askell, A.;  et~al.
\newblock Language models are few-shot learners.
\newblock {\em Adv. Neural Inf. Process. Syst.} {\bf 2020},
  {\em 33},~1877--1901.

\bibitem[Devlin et~al.(2018)Devlin, Chang, Lee, and Toutanova]{devlin2018bert}
Devlin, J.; Chang, M.W.; Lee, K.; Toutanova, K.
\newblock BERT: Pre-training of deep bidirectional transformers for language
  understanding.
\newblock {\em arXiv } {\bf 2018}, arXiv:1810.04805.

\bibitem[Lewis et~al.(2020)Lewis, Liu, Goyal, Ghazvininejad, Mohamed, Levy,
  Stoyanov, and Zettlemoyer]{lewis2020bart}
Lewis, M.; Liu, Y.; Goyal, N.; Ghazvininejad, M.; Mohamed, A.; Levy, O.;
  Stoyanov, V.; Zettlemoyer, L.
\newblock BART: Denoising Sequence-to-Sequence Pre-training for Natural
  Language Generation, Translation, and Comprehension.
\newblock In Proceedings of the 58th Annual Meeting of the
  Association for Computational Linguistics, {Online, 5--10 July}  2020;  pp. 7871--7880.

\bibitem[Nallapati et~al.(2016)Nallapati, Zhou, Gulcehre, Xiang,
  et~al.]{nallapati2016abstractive}
Nallapati, R.; Zhou, B.; Gulcehre, C.; Xiang, B.
\newblock Abstractive text summarization using sequence-to-sequence rnns and
  beyond.
\newblock {\em arXiv } {\bf 2016}, arXiv:1602.06023.

\bibitem[Zhang et~al.(2018)Zhang, Wang, and Liu]{zhang2018deep}
Zhang, L.; Wang, S.; Liu, B.
\newblock Deep Learning for Sentiment Analysis: A Survey.
\newblock {\em IEEE Trans. Affect. Comput.} {\bf 2018}, {\em
  10},~235--255.

\bibitem[Raffel et~al.(2020)Raffel, Shazeer, Roberts, Lee, Narang, Matena,
  Zhou, Li, and Liu]{raffel2020exploring}
Raffel, C.; Shazeer, N.; Roberts, A.; Lee, K.; Narang, S.; Matena, M.; Zhou,
  Y.; Li, W.; Liu, P.J.
\newblock Exploring the Limits of Transfer Learning with a Unified Text-to-Text
  Transformer.
\newblock {\em J. Mach. Learn. Res.} {\bf 2020}, {\em
  21},~1--67.

\bibitem[Bender et~al.(2021)Bender, Gebru, McMillan-Major, and
  Shmitchell]{bender2021dangers}
Bender, E.M.; Gebru, T.; McMillan-Major, A.; Shmitchell, S.
\newblock On the dangers of stochastic parrots: Can language models be too big.
\newblock In Proceedings of the 2021 ACM Conference on
  Fairness, Accountability, and Transparency, {Virtual, 3--10 March}  2021;  pp. 610--623.

\bibitem[Blodgett et~al.(2020)Blodgett, Barocas, Daum{\'e}~III, and
  Wallach]{blodgett2020language}
Blodgett, S.L.; Barocas, S.; Daum{\'e}, H., III; Wallach, H.
\newblock Language (Technology) is Power: A Critical Survey of {``}Bias{''} in
  {NLP}.
\newblock In \emph{Proceedings of the 58th Annual Meeting of the
  Association for Computational Linguistics}; Jurafsky, D., Chai, J., Schluter,
  N., Tetreault, J., Eds.; Association for Computational Linguistics: {Stroudsburg, PA, USA},  2020;  pp.
  5454--5476.

\bibitem[Rajkomar et~al.(2018)Rajkomar, Hardt, Howell, Corrado, and
  Chin]{rajkomar2018ensuring}
Rajkomar, A.; Hardt, M.; Howell, M.D.; Corrado, G.; Chin, M.H.
\newblock Ensuring Fairness in Machine Learning to Advance Health Equity.
\newblock {\em Ann. Intern. Med.} {\bf 2018}, {\em 169},~866--872.

\bibitem[Angwin et~al.(2016)Angwin, Larson, Mattu, and
  Kirchner]{angwin2016machine}
Angwin, J.; Larson, J.; Mattu, S.; Kirchner, L.
\newblock Machine Bias.
\newblock {\em ProPublica} {\bf 2016}, {\em 23},~139--159.

\bibitem[Chen et~al.(2018)Chen, Ma, Hannak, and Wilson]{chen2018my}
Chen, M.; Ma, Z.; Hannak, A.; Wilson, C.
\newblock My Fair LADY: Detecting and Mitigating Bias in Job Advertisements.
\newblock In Proceedings of the 2018 World Wide Web Conference, {Lyon, France, 23--27 April } 2018; 
  pp. 991--1000.

\bibitem[Caliskan et~al.(2017)Caliskan, Bryson, and
  Narayanan]{caliskan2017semantics}
Caliskan, A.; Bryson, J.J.; Narayanan, A.
\newblock Semantics derived automatically from language corpora contain
  human-like biases.
\newblock {\em Science} {\bf 2017}, {\em 356},~183--186.

\bibitem[Hastie et~al.(2009)Hastie, Tibshirani, and Friedman]{Hastie2009}
Hastie, T.; Tibshirani, R.; Friedman, J.
\newblock {\em The Elements of Statistical Learning: Data Mining, Inference,
  and Prediction}, 2nd ed.; Springer Science \& Business Media: {Berlin/Heidelberg, Germany,}  2009.

\bibitem[Akaike(1974)]{Akaike1974}
Akaike, H.
\newblock A New Look at the Statistical Model Identification.
\newblock {\em IEEE Trans. Autom. Control} {\bf 1974}, {\em
  19},~716--723.
\newblock {\url{https://doi.org/10.1109/TAC.1974.1100705}}.

\bibitem[Heckman(1979)]{Heckman1979}
Heckman, J.J.
\newblock Sample Selection Bias as a Specification Error.
\newblock {\em Econometrica} {\bf 1979}, {\em 47},~153--161.
\newblock {\url{https://doi.org/10.2307/1912352}}.

\bibitem[Vardi(1982)]{Vardi1982}
Vardi, Y.
\newblock Nonparametric Estimation in the Presence of Length Bias.
\newblock {\em  Ann. Stat.} {\bf 1982}, {\em 10},~616--620.
\newblock {\url{https://doi.org/10.1214/aos/1176345802}}.

\bibitem[Tibshirani(1996)]{Tibshirani1996}
Tibshirani, R.
\newblock Regression Shrinkage and Selection via the Lasso.
\newblock {\em J. R. Stat. Soc. Ser. B
  (Methodol.)} {\bf 1996}, {\em 58},~267--288.

\bibitem[Nemes et~al.(2009)Nemes, Jonasson, Genell, and Steineck]{Nemes2009}
Nemes, S.; Jonasson, J.M.; Genell, A.; Steineck, G.
\newblock Bias in odds ratios by logistic regression modelling and sample size.
\newblock {\em BMC Med. Res. Methodol.} {\bf 2009}, {\em 9},~56.
\newblock {\url{https://doi.org/10.1186/1471-2288-9-56}}.

\bibitem[Hongfei~Li and Hou(2024)]{li2024bias}
Li, H.; Qian H.;~Li, C.T.; Hou, K.
\newblock Issues in cox proportional hazards model with unequal randomization.
\newblock {\em J. Biopharm. Stat.} {\bf 2026}, {\em 36}, 330--335. 
\newblock {\url{https://doi.org/10.1080/10543406.2024.2418139}}.

\bibitem[Rubin(1974)]{Rubin1974}
Rubin, D.B.
\newblock Estimating Causal Effects of Treatments in Randomized and
  Nonrandomized Studies.
\newblock {\em J. Educ. Psychol.} {\bf 1974}, {\em
  66},~688--701.
\newblock {\url{https://doi.org/10.1037/h0037350}}.

\bibitem[Schick et~al.(2021)Schick, Udupa, and Sch{\"u}tze]{schick2021self}
Schick, T.; Udupa, S.; Sch{\"u}tze, H.
\newblock Self-Diagnosis and Self-Debiasing: A Proposal for Reducing
  Corpus-Based Bias in {NLP}.
\newblock {\em Trans. Assoc. Comput. Linguist.}
  {\bf 2021}, {\em 9},~1408--1424.
\newblock {\url{https://doi.org/10.1162/tacl_a_00434}}.

\bibitem[Sun et~al.(2019)Sun, Gaut, Tang, Huang, ElSherief, Zhao, Mirza,
  Belding, Chang, and Wang]{sun2019mitigating}
Sun, T.; Gaut, A.; Tang, S.; Huang, Y.; ElSherief, M.; Zhao, J.; Mirza, D.;
  Belding, E.; Chang, K.W.; Wang, W.Y.
\newblock Mitigating gender bias in natural language processing: Literature
  review.
\newblock {\em arXiv } {\bf 2019}, arXiv:1906.08976.

\bibitem[Bolukbasi et~al.(2016)Bolukbasi, Chang, Zou, Saligrama, and
  Kalai]{bolukbasi2016man}
Bolukbasi, T.; Chang, K.W.; Zou, J.Y.; Saligrama, V.; Kalai, A.T.
\newblock Man is to computer programmer as woman is to homemaker? Debiasing word embeddings.
\newblock {\em Adv. Neural Inf. Process. Syst.} {\bf 2016},
  {\em {29}}, 4356--4364.  


\bibitem[Zhao et~al.(2018)Zhao, Wang, Yatskar, Ordonez, and
  Chang]{zhao2018gender}
Zhao, J.; Wang, T.; Yatskar, M.; Ordonez, V.; Chang, K.W.
\newblock Gender Bias in Coreference Resolution: Evaluation and Debiasing
  Methods.
\newblock In Proceedings of the 2018 Conference of the North
  American Chapter of the Association for Computational Linguistics, {New Orleans, LA, USA, 1--6 June} 2018;  pp.
  15--20.

\bibitem[Hovy and Prabhumoye(2021)]{hovy2021five}
Hovy, D.; Prabhumoye, S.
\newblock Five Sources of Bias in Natural Language Processing.
\newblock In Proceedings of the 59th Annual Meeting of the
  Association for Computational Linguistics, {Online, 1--6 August } 2021;  pp. 878--888.

\bibitem[Sap et~al.(2019)Sap, Card, Gabriel, Choi, and Smith]{sap2019risk}
Sap, M.; Card, D.; Gabriel, S.; Choi, Y.; Smith, N.A.
\newblock The Risk of Racial Bias in Hate Speech Detection.
\newblock In Proceedings of the 57th Annual Meeting of the
  Association for Computational Linguistics,  {Florence, Italy, 28 July--2 August}  2019;  pp. 1668--1678.

\bibitem[Kiritchenko and Mohammad(2018)]{kiritchenko2018examining}
Kiritchenko, S.; Mohammad, S.M.
\newblock Examining Gender and Race Bias in Two Hundred Sentiment Analysis
  Systems.
\newblock {\em arXiv } {\bf 2018}, arXiv:1805.04508.

\bibitem[Obermeyer et~al.(2019)Obermeyer, Powers, Vogeli, and
  Mullainathan]{obermeyer2019dissecting}
Obermeyer, Z.; Powers, B.; Vogeli, C.; Mullainathan, S.
\newblock Dissecting Racial Bias in an Algorithm Used to Manage the Health of
  Populations.
\newblock {\em Science} {\bf 2019}, {\em 366},~447--453.

\bibitem[Pariser(2011)]{pariser2011filter}
Pariser, E.
\newblock {\em The Filter Bubble: How the New Personalized Web is Changing What
  We Read and How We Think}; Penguin: {London, UK},  2011.

\bibitem[Sap et~al.(2019)Sap, Gabriel, Qin, Jurafsky, Smith, and
  Choi]{sap2019social}
Sap, M.; Gabriel, S.; Qin, L.; Jurafsky, D.; Smith, N.A.; Choi, Y.
\newblock Social Bias Frames: Reasoning about Social and Power Implications of
  Language.
\newblock In Proceedings of the 57th Annual Meeting of the
  Association for Computational Linguistics,   {Florence, Italy, 28 July--2 August} 2019;  pp. 5477--5483.

\bibitem[Sheng et~al.(2021)Sheng, Chang, Natarajan, and
  Peng]{sheng2021societal}
Sheng, E.; Chang, K.W.; Natarajan, P.; Peng, N.
\newblock Societal Biases in Language Generation: Progress and Challenges.
\newblock In Proceedings of the 59th Annual Meeting of the
  Association for Computational Linguistics, {Online, 1--6 August } 2021;  pp. 4275--4293.

\bibitem[Nadeem et~al.(2021)Nadeem, Bethke, and Reddy]{nadeem2021stereoset}
Nadeem, M.; Bethke, A.; Reddy, S.
\newblock StereoSet: Measuring stereotypical bias in pretrained language
  models.
\newblock In Proceedings of the 59th Annual Meeting of the
  Association for Computational Linguistics and the 11th International Joint
  Conference on Natural Language Processing (Volume 1: Long Papers), {Online, 1--6 August }   2021;  pp.
  5356--5371.

\bibitem[Hewitt and Manning(2019)]{hewitt2019structural}
Hewitt, J.; Manning, C.D.
\newblock A Structural Probe for Finding Syntax in Word Representations.
\newblock In Proceedings of the 2019 Conference of the North
  American Chapter of the Association for Computational Linguistics, {Minneapolis, MN, USA, 2--7 June} 2019;  pp.
  4129--4138.

\bibitem[Zhao et~al.(2018)Zhao, Wang, Yatskar, Ordonez, and
  Chang]{zhao2018learning}
{Zhao, J.; Wang, T.; Yatskar, M.; Ordonez, V.; Chang, K.W.
\newblock Learning Gender-Neutral Word Embeddings.} 
\newblock In Proceedings of the 2018 Conference on Empirical
  Methods in Natural Language Processing, {Brussels, Belgium, 31 October--4 November} 2018;  pp. 4847--4853.

\bibitem[Liang et~al.(2020)Liang, Zheng, Salakhutdinov, and
  Morency]{liang2020towards}
{Liang, P.P.; Zheng, L.; Salakhutdinov, R.; Morency, L.P. 
\newblock Towards Debiasing Sentence Representations.}
\newblock In Proceedings of the 58th Annual Meeting of the
  Association for Computational Linguistics, {Online, 5--10 July}  2020;  pp. 5502--5515.

\bibitem[Mehrabi et~al.(2021)Mehrabi, Morstatter, Saxena, Lerman, and
  Galstyan]{mehrabi2021survey}
Mehrabi, N.; Morstatter, F.; Saxena, N.; Lerman, K.; Galstyan, A.
\newblock A Survey on Bias and Fairness in Machine Learning.
\newblock {\em ACM Comput. Surv.} {\bf 2021}, {\em 54},~1--35.

\bibitem[Navigli et~al.(2023)Navigli, Conia, and Ross]{navigli2023biases}
Navigli, R.; Conia, S.; Ross, B.
\newblock Biases in Large Language Models: Origins, Inventory, and Discussion.
\newblock {\em ACM  J. Data Inf. Qual.} {\bf 2023}, {\em 15}, {1--21}.

\bibitem[Gallegos et~al.(2024)Gallegos, Rossi, Barrow, Tanjim, Kim,
  Dernoncourt, Yu, Zhang, and Ahmed]{gallegos2024bias}
Gallegos, I.O.; Rossi, R.A.; Barrow, J.; Tanjim, M.M.; Kim, S.; Dernoncourt,
  F.; Yu, T.; Zhang, R.; Ahmed, N.K.
\newblock Bias and fairness in large language models: A survey.
\newblock {\em Comput. Linguist.} {\bf 2024}, {\emph{50}, 1097--1179.}

\bibitem[Doan et~al.(2024)Doan, Wang, Nguyen, and Zhang]{doan2024fairness}
Doan, T.V.; Wang, Z.; Nguyen, M.N.; Zhang, W.
\newblock Fairness in Large Language Models in three hours.
\newblock In Proceedings of the 33rd ACM International
  Conference on Information \& Knowledge Management, Boise, ID, USA,  {21--25 October} 2024.

\bibitem[May et~al.(2019)May, Wang, Bordia, Bowman, and
  Rudinger]{may2019measuring}
May, C.; Wang, A.; Bordia, S.; Bowman, S.R.; Rudinger, R.
\newblock On measuring social biases in sentence encoders.
\newblock {\em arXiv } {\bf 2019}, arXiv:1903.10561.

\bibitem[Pagano et~al.(2023)Pagano, Loureiro, Lisboa, Peixoto, Guimar{\~a}es,
  Cruz, Araujo, Santos, Cruz, Oliveira, et~al.]{pagano2023bias}
Pagano, T.P.; Loureiro, R.B.; Lisboa, F.V.; Peixoto, R.M.; Guimar{\~a}es, G.A.;
  Cruz, G.O.; Araujo, M.M.; Santos, L.L.; Cruz, M.A.; Oliveira, E.L.;  et~al.
\newblock Bias and unfairness in machine learning models: A systematic review
  on datasets, tools, fairness metrics, and identification and mitigation
  methods.
\newblock {\em Big Data Cogn. Comput.} {\bf 2023}, {\em 7},~15.

\bibitem[Ray(2023)]{ray2023chatgpt}
Ray, P.P.
\newblock ChatGPT: A comprehensive review on background, applications, key
  challenges, bias, ethics, limitations and future scope.
\newblock {\em Internet Things -Cyber-Phys. Syst.} {\bf 2023}, {\em
  3},~121--154.

\bibitem[Goldfarb-Tarrant(2024)]{goldfarb2024fairness}
Goldfarb-Tarrant, S.
\newblock Fairness in Transfer Learning for Natural Language Processing.
\newblock Ph.D. Thesis, Institute for Language, Cognition and Computation, School of Informatics, University of Edinburgh, Edinburgh, UK, 2024.
\newblock Available online: \url{https://hdl.handle.net/1842/41857} (accessed on 15 November 2024).


\bibitem[Goldman and Tsotsos(2024)]{goldman2024statistical}
Goldman, J.; Tsotsos, J.K.
\newblock Statistical Challenges with Dataset Construction: Why You Will Never
  Have Enough Images.
\newblock {\em arXiv } {\bf 2024}, arXiv:2408.11160.

\bibitem[Das et~al.(2024)Das, De~Langis, Martin, Kim, Lee, Kim, Hayati, Owan,
  Hu, Parkar, et~al.]{das2024under}
Das, D.; De~Langis, K.; Martin, A.; Kim, J.; Lee, M.; Kim, Z.M.; Hayati, S.;
  Owan, R.; Hu, B.; Parkar, R.;  et~al.
\newblock Under the surface: Tracking the artifactuality of llm-generated data.
\newblock {\em arXiv } {\bf 2024}, arXiv:2401.14698.

\bibitem[Alvero et~al.(2024)Alvero, Lee, Regla-Vargas, Kizilec, Joachims, and
  Antonio]{alvero2024large}
Alvero, A.; Lee, J.; Regla-Vargas, A.; Kizilec, R.; Joachims, T.; Antonio, A.L.
\newblock Large Language Models, Social Demography, and Hegemony: Comparing
  Authorship in Human and Synthetic Text.
\newblock \emph{{J. Big Data}} \textbf{2024},  {\emph{11}, 138.}


\bibitem[UNESCO and IRCAI(2024)]{into2024systematic}
UNESCO; IRCAI.
\newblock Challenging Systematic Prejudices: An Investigation into Bias Against Women and Girls in Large Language Models.
\newblock  2024; 20p.  Available online: 
\newblock {\url{https://unesdoc.unesco.org/ark:/48223/pf0000388971} (accessed on 15 November 2024)}. 


\bibitem[Ahmad and Bhattacharyya()]{ahmadbias}
Ahmad, A.; Bhattacharyya, P.
\newblock {Bias in Language Models: A Survey.} 
\newblock CFILT, Indian Institute of Technology Bombay, Mumbai, India, 2024.
\newblock Available online: {\url{https://www.cfilt.iitb.ac.in/resources/surveys/2024/Bias_Survey.pdf}} (accessed on 15 September 2024).

\bibitem[Talat et~al.(2022)Talat, N{\'e}v{\'e}ol, Biderman, Clinciu, Dey,
  Longpre, Luccioni, Masoud, Mitchell, Radev, et~al.]{talat2022you}
Talat, Z.; N{\'e}v{\'e}ol, A.; Biderman, S.; Clinciu, M.; Dey, M.; Longpre, S.;
  Luccioni, S.; Masoud, M.; Mitchell, M.; Radev, D.;  et~al.
\newblock You reap what you sow: On the challenges of bias evaluation under
  multilingual settings.
\newblock In Proceedings of the BigScience Episode\# 5--Workshop
  on Challenges \& Perspectives in Creating Large Language Models, {Virtual, 27 May} 2022;  pp.
  26--41.

\bibitem[Mukherjee et~al.(2024)Mukherjee, Caliskan, Zhu, and
  Anastasopoulos]{mukherjee2024global}
Mukherjee, A.; Caliskan, A.; Zhu, Z.; Anastasopoulos, A.
\newblock Global Gallery: The Fine Art of Painting Culture Portraits through
  Multilingual Instruction Tuning.
\newblock In Proceedings of the 2024 Conference of the North
  American Chapter of the Association for Computational Linguistics: Human
  Language Technologies (Volume 1: Long Papers), {Mexico City, Mexico, 16--21 June}  2024;  pp. 6398--6415.

\bibitem[Lee et~al.(2024)Lee, Aiyappa, Ahn, Kwak, and An]{lee2024neural}
Lee, B.; Aiyappa, R.; Ahn, Y.Y.; Kwak, H.; An, J.
\newblock Neural embedding of beliefs reveals the role of relative dissonance
  in human decision-making.
\newblock {\em arXiv } {\bf 2024}, arXiv:2408.07237.

\bibitem[Shi et~al.(2024)Shi, Li, Zhang, Ziems, Horesh, de~Paula, Yang,
  et~al.]{shi2024culturebank}
Shi, W.; Li, R.; Zhang, Y.; Ziems, C.; Horesh, R.; de~Paula, R.A.; Yang, D.
\newblock Culturebank: An online community-driven knowledge base towards
  culturally aware language technologies.
\newblock {\em arXiv } {\bf 2024}, arXiv:2404.15238.

\bibitem[Naous et~al.(2024)Naous, Ryan, Ritter, and Xu]{naous2024having}
Naous, T.; Ryan, M.J.; Ritter, A.; Xu, W.
\newblock Having beer after prayer? Measuring cultural bias in large language
  models.
\newblock In Proceedings of the 62nd Annual Meeting of the
  Association for Computational Linguistics (Volume 1: Long Papers), {Bangkok, Thailand, 11--16 August}  2024;  pp.
  16366--16393.

\bibitem[Wang et~al.(2024)Wang, Jiao, Huang, Dai, Huang, Tu, and
  Lyu]{wang2024not}
Wang, W.; Jiao, W.; Huang, J.; Dai, R.; Huang, J.T.; Tu, Z.; Lyu, M.
\newblock Not all countries celebrate thanksgiving: On the cultural dominance
  in large language models.
\newblock In Proceedings of the 62nd Annual Meeting of the
  Association for Computational Linguistics (Volume 1: Long Papers), {Bangkok, Thailand, 11--16 August} 2024;  pp.
  6349--6384.

\bibitem[Salinas et~al.(2023)Salinas, Penafiel, McCormack, and
  Morstatter]{salinas2023not}
Salinas, A.; Penafiel, L.; McCormack, R.; Morstatter, F.
\newblock ``Im not Racist but...'': Discovering Bias in the Internal Knowledge
  of Large Language Models.
\newblock {\em arXiv } {\bf 2023}, arXiv:2310.08780.

\bibitem[Bussaja(2024)]{bussaja2024evaluating}
Bussaja, J.
\newblock Evaluating Racial Bias in Large Language Models: The Necessity for “SMOKY”.
\newblock {\em Int. J. Soft Comput. (IJSC)} \textbf{2024}, { \emph{15}}, 1--15.
\newblock {\url{https://doi.org/10.5121/ijsc.2024.15301}}


\bibitem[KUNTZ and SILVA(2023)]{kuntz2023authors}
Kuntz, J.B.; Silva, E.C.
\newblock \emph{Who Authors the Internet? Analyzing Gender Diversity in ChatGPT-3 Training Data}; 
\newblock University of Pittsburgh: {Pittsburgh, PA, USA,} 2023.

\bibitem[Qu and Wang(2024)]{qu2024performance}
Qu, Y.; Wang, J.
\newblock Performance and biases of Large Language Models in public opinion
  simulation.
\newblock {\em Humanit. Soc. Sci. Commun.} {\bf 2024}, {\em
  11},~1--13.

\bibitem[Kotek et~al.(2023)Kotek, Dockum, and Sun]{kotek2023gender}
Kotek, H.; Dockum, R.; Sun, D.
\newblock Gender bias and stereotypes in large language models.
\newblock In Proceedings of the ACM Collective Intelligence
  Conference, {Delft, The Netherlands, 6--9 November}  2023;  pp. 12--24.

\bibitem[Dwivedi et~al.(2023)Dwivedi, Ghosh, and Dwivedi]{dwivedi2023breaking}
Dwivedi, S.; Ghosh, S.; Dwivedi, S.
\newblock Breaking the bias: Gender fairness in LLMs using prompt engineering
  and in-context learning.
\newblock {\em Rupkatha J. Interdiscip. Stud. Humanit.}
  {\bf 2023}, {\em 15}, {1--18.} 


\bibitem[de~S{\'a} et~al.(2024)de~S{\'a}, Da~Silveira, and
  Pruski]{de2024semantic}
de~S{\'a}, J.M.C.; Da~Silveira, M.; Pruski, C.
\newblock Semantic Change Characterization with LLMs using Rhetorics.
\newblock {\em arXiv } {\bf 2024}, arXiv:2407.16624.

\bibitem[Petrov et~al.(2023)Petrov, La~Malfa, Torr, and
  Bibi]{petrov2023language}
Petrov, A.; La~Malfa, E.; Torr, P.; Bibi, A.
\newblock Language model tokenizers introduce unfairness between languages.
\newblock {\em Adv. Neural Inf. Process. Syst.} {\bf 2023},
  {\em 36},~36963--36990.

\bibitem[Ahia et~al.(2023)Ahia, Kumar, Gonen, Kasai, Mortensen, Smith, and
  Tsvetkov]{ahia2023all}
Ahia, O.; Kumar, S.; Gonen, H.; Kasai, J.; Mortensen, D.R.; Smith, N.A.;
  Tsvetkov, Y.
\newblock Do all languages cost the same? tokenization in the era of commercial
  language models.
\newblock In Proceedings of the 2023 Conference on Empirical
  Methods in Natural Language Processing,  {Singapore, 6--10 December} 2023;  pp. 9904--9923.

\bibitem[Ovalle et~al.(2024)Ovalle, Mehrabi, Goyal, Dhamala, Chang, Zemel,
  Galstyan, Pinter, and Gupta]{ovalle2024tokenization}
Ovalle, A.; Mehrabi, N.; Goyal, P.; Dhamala, J.; Chang, K.W.; Zemel, R.;
  Galstyan, A.; Pinter, Y.; Gupta, R.
\newblock Tokenization matters: Navigating data-scarce tokenization for gender
  inclusive language technologies.
\newblock In Proceedings of the Findings of the Association for Computational
  Linguistics: NAACL 2024, {Mexico City, Mexico, 16--21 June}  2024;  pp. 1739--1756.

\bibitem[Garg et~al.(2018)Garg, Schiebinger, Jurafsky, and Zou]{garg2018word}
Garg, N.; Schiebinger, L.; Jurafsky, D.; Zou, J.
\newblock Word embeddings quantify 100 years of gender and ethnic stereotypes.
\newblock {\em Proc. Natl. Acad. Sci. USA} {\bf 2018},
  {\em 115},~E3635--E3644.

\bibitem[Rakshit et~al.(2025)Rakshit, Singh, Keshari, Chowdhury, Jain, and
  Chadha]{rakshit2025prejudice}
Rakshit, A.; Singh, S.; Keshari, S.; Chowdhury, A.G.; Jain, V.; Chadha, A.
\newblock From prejudice to parity: A new approach to debiasing large language
  model word embeddings.
\newblock In Proceedings of the 31st International
  Conference on Computational Linguistics,  {Abu Dhabi, United Arab Emirates, 19--24 January} 2025;  pp. 6718--6747.

\bibitem[Li et~al.(2023)Li, Du, Song, Wang, and Wang]{li2023survey}
Li, Y.; Du, M.; Song, R.; Wang, X.; Wang, Y.
\newblock A survey on fairness in large language models.
\newblock {\em arXiv } {\bf 2023}, arXiv:2308.10149.

\bibitem[Chang et~al.(2024)Chang, Wang, Wang, Wu, Yang, Zhu, Chen, Yi, Wang,
  Wang, et~al.]{chang2024survey}
Chang, Y.; Wang, X.; Wang, J.; Wu, Y.; Yang, L.; Zhu, K.; Chen, H.; Yi, X.;
  Wang, C.; Wang, Y.;  et~al.
\newblock A survey on evaluation of large language models.
\newblock {\em ACM Trans. Intell. Syst. Technol.} {\bf
  2024}, {\em 15},~1--45.

\bibitem[Vulic and Moens(2013)]{vulic2013cross}
Vulic, I.; Moens, M.F.
\newblock Cross-lingual semantic similarity of words as the similarity of their
  semantic word responses.
\newblock In Proceedings of the 2013 Conference of the North
  American Chapter of the Association for Computational Linguistics: Human
  Language Technologies (NAACL-HLT 2013), {Atlanta, GA, USA, 9--14 June} 2013;  ACL: East Stroudsburg, PA, USA,  2013;  pp.
  106--116.

\bibitem[Hendricks et~al.(2018)Hendricks, Burns, Saenko, Darrell, and
  Rohrbach]{hendricks2018women}
Hendricks, L.A.; Burns, K.; Saenko, K.; Darrell, T.; Rohrbach, A.
\newblock Women also snowboard: Overcoming bias in captioning models.
\newblock In Proceedings of the European Conference on
  Computer Vision (ECCV), {Munich, Germany, 8--14 September} 2018;  pp. 771--787.

\bibitem[Lee et~al.(2017)Lee, He, Lewis, and Zettlemoyer]{lee2017end}
Lee, K.; He, L.; Lewis, M.; Zettlemoyer, L.
\newblock End-to-end neural coreference resolution.
\newblock {\em arXiv } {\bf 2017}, arXiv:1707.07045.

\bibitem[Rudinger et~al.(2018)Rudinger, Naradowsky, Leonard, and
  Van~Durme]{rudinger2018gender}
Rudinger, R.; Naradowsky, J.; Leonard, B.; Van~Durme, B.
\newblock Gender bias in coreference resolution.
\newblock {\em arXiv } {\bf 2018}, arXiv:1804.09301.

\bibitem[Hovy and Yang(2021)]{hovy2021importance}
Hovy, D.; Yang, D.
\newblock The importance of modeling social factors of language: Theory and
  practice.
\newblock In Proceedings of the 2021 Conference of the North
  American Chapter of the Association for Computational Linguistics: Human
  Language Technologies, {Online, 6--11 June} 2021;  pp. 588--602.

\bibitem[Cer et~al.(2017)Cer, Diab, Agirre, Lopez-Gazpio, and
  Specia]{cer2017semeval}
Cer, D.; Diab, M.; Agirre, E.; Lopez-Gazpio, I.; Specia, L.
\newblock Semeval-2017 task 1: Semantic textual similarity-multilingual and
  cross-lingual focused evaluation.
\newblock {\em arXiv } {\bf 2017}, arXiv:1708.00055.

\bibitem[Zhao et~al.(2017)Zhao, Wang, Yatskar, Ordonez, and Chang]{zhao2017men}
Zhao, J.; Wang, T.; Yatskar, M.; Ordonez, V.; Chang, K.W.
\newblock Men also like shopping: Reducing gender bias amplification using
  corpus-level constraints.
\newblock {\em arXiv } {\bf 2017}, arXiv:1707.09457.

\bibitem[Diaz et~al.(2016)Diaz, Mitra, and Craswell]{diaz2016query}
Diaz, F.; Mitra, B.; Craswell, N.
\newblock Query expansion with locally-trained word embeddings.
\newblock {\em arXiv } {\bf 2016}, arXiv:1605.07891.

\bibitem[Tan and Bond(2011)]{tan2011building}
Tan, L.; Bond, F.
\newblock Building and annotating the linguistically diverse NTU-MC
  (NTU-multilingual corpus).
\newblock In Proceedings of the 25th Pacific Asia Conference
  on Language, Information and Computation, {Singapore, 16--18 December }  2011;  pp.
  362--371.

\bibitem[Bowman et~al.(2015)Bowman, Angeli, Potts, and
  Manning]{bowman2015large}
Bowman, S.R.; Angeli, G.; Potts, C.; Manning, C.D.
\newblock A large annotated corpus for learning natural language inference.
\newblock {\em arXiv } {\bf 2015}, arXiv:1508.05326.

\bibitem[Chen et~al.(2016)Chen, Zhu, Ling, Wei, Jiang, and
  Inkpen]{chen2016enhanced}
Chen, Q.; Zhu, X.; Ling, Z.; Wei, S.; Jiang, H.; Inkpen, D.
\newblock Enhanced LSTM for natural language inference.
\newblock {\em arXiv } {\bf 2016}, arXiv:1609.06038.

\bibitem[De-Arteaga et~al.(2019)De-Arteaga, Romanov, Wallach, Chayes, Borgs,
  Chouldechova, Geyik, Kenthapadi, and Kalai]{de2019bias}
De-Arteaga, M.; Romanov, A.; Wallach, H.; Chayes, J.; Borgs, C.; Chouldechova,
  A.; Geyik, S.; Kenthapadi, K.; Kalai, A.T.
\newblock Bias in bios: A case study of semantic representation bias in a
  high-stakes setting.
\newblock In Proceedings of the Conference on Fairness,
  Accountability, and Transparency, {Atlanta, GA, USA, 29--31 January}   2019;  pp. 120--128.

\bibitem[Meng et~al.(2020)Meng, Zhang, Huang, Xiong, Ji, Zhang, and
  Han]{meng2020text}
Meng, Y.; Zhang, Y.; Huang, J.; Xiong, C.; Ji, H.; Zhang, C.; Han, J.
\newblock Text classification using label names only: A language model
  self-training approach.
\newblock {\em arXiv } {\bf 2020}, arXiv:2010.07245.

\bibitem[D{\'\i}az et~al.(2018)D{\'\i}az, Johnson, Lazar, Piper, and
  Gergle]{diaz2018addressing}
D{\'\i}az, M.; Johnson, I.; Lazar, A.; Piper, A.M.; Gergle, D.
\newblock Addressing age-related bias in sentiment analysis.
\newblock In Proceedings of the 2018 CHI Conference on Human
  Factors in Computing Systems, {Montréal, QC, Canada, 21--26  April} 2018;  pp. 1--14.

\bibitem[Rajpurkar et~al.(2018)Rajpurkar, Jia, and Liang]{rajpurkar2018know}
Rajpurkar, P.; Jia, R.; Liang, P.
\newblock Know what you don't know: Unanswerable questions for SQuAD.
\newblock {\em arXiv } {\bf 2018}, arXiv:1806.03822.

\bibitem[Webster et~al.(2018)Webster, Recasens, Axelrod, and
  Baldridge]{webster2018mind}
Webster, K.; Recasens, M.; Axelrod, V.; Baldridge, J.
\newblock Mind the GAP: A balanced corpus of gendered ambiguous pronouns.
\newblock {\em Trans. Assoc. Comput. Linguist.}
  {\bf 2018}, {\em 6},~605--617.

\bibitem[Bender and Friedman(2018)]{bender2018data}
Bender, E.M.; Friedman, B.
\newblock Data statements for natural language processing: Toward mitigating
  system bias and enabling better science.
\newblock {\em Trans. Assoc. Comput. Linguist.}
  {\bf 2018}, {\em 6},~587--604.

\bibitem[Pang et~al.(2008)Pang, Lee, et~al.]{pang2008opinion}
Pang, B.; Lee, L. 
\newblock Opinion mining and sentiment analysis.
\newblock {\em Found.  Trends{\textregistered}  Inf. Retr.} {\bf 2008}, {\em 2},~1--135.

\bibitem[Misiunas and Keyser(2019)]{misiunas2019density}
Misiunas, K.; Keyser, U.F.
\newblock Density-dependent speed-up of particle transport in channels.
\newblock {\em Phys. Rev. Lett.} {\bf 2019}, {\em 122},~214501.

\bibitem[Burnap and Williams(2015)]{burnap2015cyber}
Burnap, P.; Williams, M.L.
\newblock Cyber hate speech on twitter: An application of machine
  classification and statistical modeling for policy and decision making.
\newblock {\em Policy Internet} {\bf 2015}, {\em 7},~223--242.

\bibitem[Hovy and S{\o}gaard(2015)]{hovy2015tagging}
Hovy, D.; S{\o}gaard, A.
\newblock Tagging performance correlates with author age.
\newblock In Proceedings of the 53rd Annual Meeting of the
  Association for Computational Linguistics and the 7th International Joint
  Conference on Natural Language Processing (Volume 2: Short Papers),  {Beijing, China, 26--31 July} 2015,
  pp. 483--488.

\bibitem[Blodgett and O'Connor(2017)]{blodgett2017racial}
Blodgett, S.L.; O'Connor, B.
\newblock Racial disparity in natural language processing: A case study of
  social media african-american english.
\newblock {\em arXiv } {\bf 2017}, arXiv:1707.00061.

\bibitem[Rajpurkar et~al.(2016)Rajpurkar, Zhang, Lopyrev, and
  Liang]{rajpurkar2016squad}
Rajpurkar, P.; Zhang, J.; Lopyrev, K.; Liang, P.
\newblock Squad: 100,000+ questions for machine comprehension of text.
\newblock {\em arXiv } {\bf 2016}, arXiv:1606.05250.

\bibitem[Gardner et~al.(2018)Gardner, Grus, Neumann, Tafjord, Dasigi, Liu,
  Peters, Schmitz, and Zettlemoyer]{gardner2018allennlp}
Gardner, M.; Grus, J.; Neumann, M.; Tafjord, O.; Dasigi, P.; Liu, N.; Peters,
  M.; Schmitz, M.; Zettlemoyer, L.
\newblock Allennlp: A deep semantic natural language processing platform.
\newblock {\em arXiv } {\bf 2018}, arXiv:1803.07640.

\bibitem[Kwiatkowski et~al.(2019)Kwiatkowski, Palomaki, Redfield, Collins,
  Parikh, Alberti, Epstein, Polosukhin, Devlin, Lee,
  et~al.]{kwiatkowski2019natural}
Kwiatkowski, T.; Palomaki, J.; Redfield, O.; Collins, M.; Parikh, A.; Alberti,
  C.; Epstein, D.; Polosukhin, I.; Devlin, J.; Lee, K.;  et~al.
\newblock Natural questions: A benchmark for question answering research.
\newblock {\em Trans. Assoc. Comput. Linguist.}
  {\bf 2019}, {\em 7},~453--466.

\bibitem[Radford et~al.(2019)Radford, Wu, Child, Luan, Amodei, Sutskever,
  et~al.]{radford2019language}
Radford, A.; Wu, J.; Child, R.; Luan, D.; Amodei, D.; Sutskever, I.
\newblock Language models are unsupervised multitask learners.
\newblock {\em OpenAI Blog} {\bf 2019}, {\em 1},~9.

\bibitem[Birhane and Prabhu(2021)]{birhane2021large}
Birhane, A.; Prabhu, V.U.
\newblock Large image datasets: A pyrrhic win for computer vision?
\newblock In Proceedings of the 2021 IEEE Winter Conference on Applications of
  Computer Vision (WACV), {Waikoloa, HI, USA, 3--8 January} 2021;  pp. 1536--1546.

\bibitem[Zhang et~al.(2019)Zhang, Sun, Galley, Chen, Brockett, Gao, Gao, Liu,
  and Dolan]{zhang2019dialogpt}
Zhang, Y.; Sun, S.; Galley, M.; Chen, Y.C.; Brockett, C.; Gao, X.; Gao, J.;
  Liu, J.; Dolan, B.
\newblock Dialogpt: Large-scale generative pre-training for conversational
  response generation.
\newblock {\em arXiv } {\bf 2019}, arXiv:1911.00536.

\bibitem[Hitsch et~al.(2010)Hitsch, Horta{\c{c}}su, and
  Ariely]{hitsch2010matching}
Hitsch, G.J.; Horta{\c{c}}su, A.; Ariely, D.
\newblock Matching and sorting in online dating.
\newblock {\em Am. Econ. Rev.} {\bf 2010}, {\em 100},~130--163.

\bibitem[Wagner et~al.(2016)Wagner, Graells-Garrido, Garcia, and
  Menczer]{wagner2016women}
Wagner, C.; Graells-Garrido, E.; Garcia, D.; Menczer, F.
\newblock Women through the glass ceiling: Gender asymmetries in Wikipedia.
\newblock {\em EPJ Data Sci.} {\bf 2016}, {\em 5},~{5}. 

\bibitem[Hovy and Spruit(2016)]{hovy2016social}
Hovy, D.; Spruit, S.L.
\newblock The social impact of natural language processing.
\newblock In Proceedings of the 54th Annual Meeting of the
  Association for Computational Linguistics (Volume 2: Short Papers), {Berlin, Germany, 7--12 August}  2016;
  pp. 591--598.

\bibitem[Gebru et~al.(2021)Gebru, Morgenstern, Vecchione, Vaughan, Wallach,
  Iii, and Crawford]{gebru2021datasheets}
Gebru, T.; Morgenstern, J.; Vecchione, B.; Vaughan, J.W.; Wallach, H.; Iii,
  H.D.; Crawford, K.
\newblock Datasheets for datasets.
\newblock {\em Commun. ACM} {\bf 2021}, {\em 64},~86--92.

\bibitem[Zhang et~al.(2019)Zhang, Yao, Sun, and Tay]{zhang2019deep}
Zhang, S.; Yao, L.; Sun, A.; Tay, Y.
\newblock Deep learning based recommender system: A survey and new
  perspectives.
\newblock {\em ACM Comput. Surv. (CSUR)} {\bf 2019}, {\em 52},~1--38.

\bibitem[Ekstrand et~al.(2018)Ekstrand, Tian, Azpiazu, Ekstrand, Anuyah,
  McNeill, and Pera]{ekstrand2018all}
Ekstrand, M.D.; Tian, M.; Azpiazu, I.M.; Ekstrand, J.D.; Anuyah, O.; McNeill,
  D.; Pera, M.S.
\newblock All the cool kids, how do they fit in?: Popularity and demographic
  biases in recommender evaluation and effectiveness.
\newblock In Proceedings of the Conference on Fairness, Accountability and
  Transparency, PMLR, {New York, NY, USA, 23--24 February}  2018;  pp. 172--186.

\bibitem[Chen et~al.(2019)Chen, Ai, Jayasinghe, and Croft]{chen2019correcting}
Chen, R.C.; Ai, Q.; Jayasinghe, G.; Croft, W.B.
\newblock Correcting for recency bias in job recommendation.
\newblock In Proceedings of the 28th ACM International
  Conference on Information and Knowledge Management,  {Beijing, China, 3--7 November} 2019;  pp. 2185--2188.

\bibitem[Biega et~al.(2018)Biega, Gummadi, and Weikum]{biega2018equity}
Biega, A.J.; Gummadi, K.P.; Weikum, G.
\newblock Equity of attention: Amortizing individual fairness in rankings.
\newblock In Proceedings of the  41st International ACM SIGIR Conference on
  Research \& Development in Information Retrieval, {Ann Arbor, MI, USA,  8--12 July}  2018;  pp. 405--414.

\bibitem[Burke et~al.(2018)Burke, Sonboli, and
  Ordonez-Gauger]{burke2018balanced}
Burke, R.; Sonboli, N.; Ordonez-Gauger, A.
\newblock Balanced neighborhoods for multi-sided fairness in recommendation.
\newblock In Proceedings of the Conference on Fairness, Accountability and
  Transparency, PMLR, {New York, NY, USA, 23--24 February}  2018;  pp. 202--214.

\bibitem[Karimi et~al.(2018)Karimi, Jannach, and Jugovac]{karimi2018news}
Karimi, M.; Jannach, D.; Jugovac, M.
\newblock News recommender systems--Survey and roads ahead.
\newblock {\em Inf. Process. Manag.} {\bf 2018}, {\em
  54},~1203--1227.

\bibitem[Lakew et~al.(2018)Lakew, Federico, Negri, and
  Turchi]{lakew2018multilingual}
Lakew, S.M.; Federico, M.; Negri, M.; Turchi, M.
\newblock Multilingual neural machine translation for low-resource languages.
\newblock {\em IJCoL Ital. J. Comput. Linguist.} {\bf 2018},
  {\em 4},~11--25.

\bibitem[Vaswani(2017)]{vaswani2017attention}
Vaswani, A.
\newblock Attention is all you need.
\newblock {\em Adv. Neural Inf. Process. Syst.} {\bf {2017}}, \emph{30}, 5998--6008.  


\bibitem[Prates et~al.(2020)Prates, Avelar, and Lamb]{prates2020assessing}
Prates, M.O.; Avelar, P.H.; Lamb, L.C.
\newblock Assessing gender bias in machine translation: A case study with
  google translate.
\newblock {\em Neural Comput. Appl.} {\bf 2020}, {\em
  32},~6363--6381.

\bibitem[Neri et~al.(2020)Neri, Soldani, Zimmermann, and Brogi]{neri2020design}
Neri, D.; Soldani, J.; Zimmermann, O.; Brogi, A.
\newblock Design principles, architectural smells and refactorings for
  microservices: A~multivocal review.
\newblock {\em SICS Softw.-Intensive-Cyber-Phys. Syst.} {\bf 2020}, {\em
  35},~3--15.

\bibitem[Vanmassenhove et~al.(2019)Vanmassenhove, Shterionov, and
  Way]{vanmassenhove2019lost}
Vanmassenhove, E.; Shterionov, D.; Way, A.
\newblock Lost in translation: Loss and decay of linguistic richness in machine
  translation.
\newblock {\em arXiv } {\bf 2019}, arXiv:1906.12068.

\bibitem[Sennrich(2016)]{sennrich2016neural}
Sennrich, R.
\newblock \emph{Neural Machine Translation}; 
\newblock {Institute for Language, Cognition and Computation University of
  Edinburgh}: {Edinburgh, UK}, 2016; Volume 18.

\bibitem[Otterbacher et~al.(2017)Otterbacher, Bates, and
  Clough]{otterbacher2017competent}
Otterbacher, J.; Bates, J.; Clough, P.
\newblock Competent men and warm women: Gender stereotypes and backlash in
  image search results.
\newblock In Proceedings of the 2017 CHI Conference on Human
  Factors in Computing Systems, {Denver, CO, USA, 6--11 May} 2017;  pp. 6620--6631.

\bibitem[Shen et~al.(2017)Shen, Lei, Barzilay, and Jaakkola]{shen2017style}
Shen, T.; Lei, T.; Barzilay, R.; Jaakkola, T.
\newblock Style transfer from non-parallel text by cross-alignment.
\newblock {\em Adv. Neural Inf. Process. Syst.} {\bf 2017},
  {\em {30}}, 6833--6844. 

\bibitem[Perez-Beltrachini and Lapata(2022)]{perez2022models}
Perez-Beltrachini, L.; Lapata, M.
\newblock Models and datasets for cross-lingual summarisation.
\newblock {\em arXiv } {\bf 2022}, arXiv:2202.09583.

\bibitem[Liang et~al.(2021)Liang, Wu, Morency, and
  Salakhutdinov]{liang2021towards}
Liang, P.P.; Wu, C.; Morency, L.P.; Salakhutdinov, R.
\newblock Towards understanding and mitigating social biases in language
  models.
\newblock In Proceedings of the International Conference on Machine Learning,
  PMLR, {Online, 18--24 July}  2021;  pp. 6565--6576.

\bibitem[Simmons and Hare(2023)]{simmons2023large}
Simmons, G.; Hare, C.
\newblock Large language models as subpopulation representative models: A
  review.
\newblock {\em arXiv } {\bf 2023}, arXiv:2310.17888.

\bibitem[Wang et~al.(2024)Wang, Morgenstern, and Dickerson]{wang2024large}
Wang, A.; Morgenstern, J.; Dickerson, J.P.
\newblock Large language models cannot replace human participants because they
  cannot portray identity groups.
\newblock {\em arXiv } {\bf 2024}, arXiv:2402.01908.

\bibitem[Gorti et~al.(2024)Gorti, Gaur, and Chadha]{gorti2024unboxing}
Gorti, A.; Gaur, M.; Chadha, A.
\newblock Unboxing Occupational Bias: Grounded Debiasing LLMs with US Labor
  Data.
\newblock {\em arXiv } {\bf 2024}, arXiv:2408.11247.

\bibitem[Rozado(2024)]{rozado2024political}
Rozado, D.
\newblock The political preferences of LLMs.
\newblock {\em PLoS ONE} {\bf 2024}, {\em 19},~e0306621.

\bibitem[Khanuja et~al.(2022)Khanuja, Ruder, and
  Talukdar]{khanuja2022evaluating}
Khanuja, S.; Ruder, S.; Talukdar, P.
\newblock Evaluating the Diversity, Equity and Inclusion of NLP Technology: A
  Case Study for Indian Languages.
\newblock {\em arXiv } {\bf 2022}, arXiv:2205.12676.

\bibitem[Cs{\'a}ky et~al.(2019)Cs{\'a}ky, Purgai, and
  Recski]{csaky2019improving}
Cs{\'a}ky, R.; Purgai, P.; Recski, G.
\newblock Improving neural conversational models with entropy-based data
  filtering.
\newblock {\em arXiv } {\bf 2019}, arXiv:1905.05471.

\bibitem[Durward and Thomson(2024)]{durward2024evaluating}
Durward, M.; Thomson, C.
\newblock Evaluating Vocabulary Usage in LLMs.
\newblock In Proceedings of the 19th Workshop on Innovative
  Use of NLP for Building Educational Applications (BEA 2024), {Mexico City, Mexico, 20 June} 2024;  pp.
  266--282.

\bibitem[de~Vassimon~Manela et~al.(2021)de~Vassimon~Manela, Errington, Fisher,
  van Breugel, and Minervini]{de2021stereotype}
de~Vassimon~Manela, D.; Errington, D.; Fisher, T.; van Breugel, B.; Minervini,
  P.
\newblock Stereotype and skew: Quantifying gender bias in pre-trained and
  fine-tuned language models.
\newblock In Proceedings of the 16th Conference of the
  European Chapter of the Association for Computational Linguistics: Main
  Volume,  {Online, 19--23 April} 2021;  pp. 2232--2242.

\bibitem[Dang and Verma(2024)]{dang2024data}
Dang, V.M.H.; Verma, R.M.
\newblock Data quality in NLP: Metrics and a comprehensive taxonomy.
\newblock In Proceedings of the International Symposium on Intelligent Data
  Analysis, {Stockholm, Sweden, 24--26 April 2024};  Springer:  {Berlin/Heidelberg, Germany,}   2024;  pp. 217--229.

\bibitem[Shen et~al.(2023)Shen, Tao, Ma, Neiswanger, Hestness, Vassilieva,
  Soboleva, and Xing]{shen2023slimpajama}
Shen, Z.; Tao, T.; Ma, L.; Neiswanger, W.; Hestness, J.; Vassilieva, N.;
  Soboleva, D.; Xing, E.
\newblock Slimpajama-dc: Understanding data combinations for llm training.
\newblock {\em arXiv } {\bf 2023}, arXiv:2309.10818.

\bibitem[Silva et~al.(2016)Silva, Mondal, Correa, Benevenuto, and
  Weber]{silva2016analyzing}
Silva, L.; Mondal, M.; Correa, D.; Benevenuto, F.; Weber, I.
\newblock Analyzing the targets of hate in online social media.
\newblock In Proceedings of the International AAAI
  Conference on Web and Social Media, {Cologne, Germany, 17--20 May}  2016; Volume~10,  pp. 687--690.

\bibitem[Kamal et~al.(2023)Kamal, Anwar, Sejwal, and
  Fazil]{kamal2023bicapshate}
Kamal, A.; Anwar, T.; Sejwal, V.K.; Fazil, M.
\newblock BiCapsHate: Attention to the linguistic context of hate via
  bidirectional capsules and hatebase.
\newblock {\em IEEE Trans. Comput. Soc. Syst.} {\bf 2023},
  {\em 11},~1781--1792.

\bibitem[Whissell(1989)]{whissell1989dictionary}
Whissell, C.M.
\newblock The dictionary of affect in language. In {\em The Measurement of
  Emotions}; Elsevier:  {Berlin/Heidelberg, Germany,} 
  1989; pp. 113--131.

\bibitem[Dev et~al.(2021)Dev, Sheng, Zhao, Amstutz, Sun, Hou, Sanseverino, Kim,
  Nishi, Peng, et~al.]{dev2021measures}
Dev, S.; Sheng, E.; Zhao, J.; Amstutz, A.; Sun, J.; Hou, Y.; Sanseverino, M.;
  Kim, J.; Nishi, A.; Peng, N.;  et~al.
\newblock On measures of biases and harms in NLP.
\newblock {\em arXiv } {\bf 2021}, arXiv:2108.03362.

\bibitem[Zhao et~al.(2019)Zhao, Wang, Yatskar, Cotterell, Ordonez, and
  Chang]{zhao2019gender}
Zhao, J.; Wang, T.; Yatskar, M.; Cotterell, R.; Ordonez, V.; Chang, K.W.
\newblock Gender bias in contextualized word embeddings.
\newblock {\em arXiv } {\bf 2019}, arXiv:1904.03310.

\bibitem[Kurita et~al.(2019)Kurita, Vyas, Pareek, Black, and
  Tsvetkov]{kurita2019measuring}
Kurita, K.; Vyas, N.; Pareek, A.; Black, A.W.; Tsvetkov, Y.
\newblock Measuring bias in contextualized word representations.
\newblock {\em arXiv } {\bf 2019}, arXiv:1906.07337.

\bibitem[Mackieson et~al.(2019)Mackieson, Shlonsky, and
  Connolly]{mackieson2019increasing}
Mackieson, P.; Shlonsky, A.; Connolly, M.
\newblock Increasing rigor and reducing bias in qualitative research: A
  document analysis of parliamentary debates using applied thematic analysis.
\newblock {\em Qual. Soc. Work.} {\bf 2019}, {\em 18},~965--980.

\bibitem[Basta et~al.(2019)Basta, Costa-Juss{\`a}, and
  Casas]{basta2019evaluating}
Basta, C.; Costa-Juss{\`a}, M.R.; Casas, N.
\newblock Evaluating the underlying gender bias in contextualized word
  embeddings.
\newblock {\em arXiv } {\bf 2019}, arXiv:1904.08783.

\bibitem[Wankhade et~al.(2022)Wankhade, Rao, and Kulkarni]{wankhade2022survey}
Wankhade, M.; Rao, A.C.S.; Kulkarni, C.
\newblock A survey on sentiment analysis methods, applications, and challenges.
\newblock {\em Artif. Intell. Rev.} {\bf 2022}, {\em
  55},~5731--5780.

\bibitem[Saxena et~al.(2022)Saxena, Reddy, and Saxena]{saxena2022introduction}
Saxena, A.; Reddy, H.; Saxena, P.
\newblock Introduction to sentiment analysis covering basics, tools, evaluation
  metrics, challenges, and applications.
\newblock In {\em Principles of Social Networking: The New Horizon and Emerging
  Challenges}; {Springer: Singapore}, 2022;  pp. 249--277.

\bibitem[Hasan et~al.(2019)Hasan, Maliha, and Arifuzzaman]{hasan2019sentiment}
Hasan, M.R.; Maliha, M.; Arifuzzaman, M.
\newblock Sentiment analysis with NLP on Twitter data.
\newblock In Proceedings of the 2019 International Conference on Computer,
  Communication, Chemical, Materials and Electronic Engineering (IC4ME2), {Rajshahi, Bangladesh, 11--12 July }
   2019;  pp. 1--4.

\bibitem[Naseem et~al.(2020)Naseem, Razzak, Musial, and
  Imran]{naseem2020transformer}
Naseem, U.; Razzak, I.; Musial, K.; Imran, M.
\newblock Transformer based deep intelligent contextual embedding for twitter
  sentiment analysis.
\newblock {\em Future Gener. Comput. Syst.} {\bf 2020}, {\em
  113},~58--69.

\bibitem[Abdullah and Ahmet(2022)]{abdullah2022deep}
Abdullah, T.; Ahmet, A.
\newblock Deep learning in sentiment analysis: Recent architectures.
\newblock {\em ACM Comput. Surv.} {\bf 2022}, {\em 55},~1--37.

\bibitem[Artstein(2017)]{artstein2017inter}
Artstein, R.
\newblock Inter-annotator agreement.
\newblock In {\em Handbook of Linguistic Annotation}; {Springer: Dordrecht, The Netherlands},  2017;  pp. 297--313.

\bibitem[Paun et~al.(2022)Paun, Artstein, and Poesio]{paun2022statistical}
Paun, S.; Artstein, R.; Poesio, M.
\newblock {\em Statistical Methods for Annotation Analysis}; Springer Nature:  {Berlin/Heidelberg, Germany,}
  2022.

\bibitem[Havens et~al.(2022)Havens, Terras, Bach, and
  Alex]{havens2022uncertainty}
Havens, L.; Terras, M.; Bach, B.; Alex, B.
\newblock Uncertainty and inclusivity in gender bias annotation: An annotation
  taxonomy and annotated datasets of British English text.
\newblock In Proceedings of the 4th Workshop on Gender Bias in Natural Language
  Processing, {Seattle, WA, USA, 15 July}  2022;  pp. 30--57.

\bibitem[Wang et~al.(2024)Wang, Kim, Rahman, Mitra, and Miao]{wang2024human}
Wang, X.; Kim, H.; Rahman, S.; Mitra, K.; Miao, Z.
\newblock Human-LLM collaborative annotation through effective verification of
  LLM labels.
\newblock In Proceedings of the CHI Conference on Human
  Factors in Computing Systems, {Honolulu, HI, USA, 11--16 May} 2024;  pp. 1--21.

\bibitem[Esiobu et~al.(2023)Esiobu, Tan, Hosseini, Ung, Zhang, Fernandes,
  Dwivedi-Yu, Presani, Williams, and Smith]{esiobu2023robbie}
Esiobu, D.; Tan, X.; Hosseini, S.; Ung, M.; Zhang, Y.; Fernandes, J.;
  Dwivedi-Yu, J.; Presani, E.; Williams, A.; Smith, E.
\newblock ROBBIE: Robust bias evaluation of large generative language models.
\newblock In Proceedings of the 2023 Conference on Empirical
  Methods in Natural Language Processing, {Singapore, 6--10 December}  2023;  pp. 3764--3814.

\bibitem[Mosca et~al.(2022)Mosca, Szigeti, Tragianni, Gallagher, and
  Groh]{mosca2022shap}
Mosca, E.; Szigeti, F.; Tragianni, S.; Gallagher, D.; Groh, G.
\newblock SHAP-based explanation methods: A review for NLP interpretability.
\newblock In Proceedings of the 29th International
  Conference on Computational Linguistics, {Gyeongju, Republic of Korea, 12--17 October } 2022;  pp. 4593--4603.

\bibitem[Wu et~al.(2024)Wu, Bulathwela, Perez-Ortiz, and
  Koshiyama]{wu2024auditing}
Wu, Z.; Bulathwela, S.; Perez-Ortiz, M.; Koshiyama, A.S.
\newblock Auditing Large Language Models for Enhanced Text-Based Stereotype
  Detection and Probing-Based Bias Evaluation.
\newblock {\em arXiv } {\bf 2024}, arXiv:2404.01768.

\bibitem[Lin et~al.(2024)Lin, Guan, Zhang, Zhang, Li, and
  Zhang]{lin2024towards}
Lin, Z.; Guan, S.; Zhang, W.; Zhang, H.; Li, Y.; Zhang, H.
\newblock Towards trustworthy LLMs: A review on debiasing and dehallucinating
  in large language models.
\newblock {\em Artif. Intell. Rev.} {\bf 2024}, {\em 57},~1--50.

\bibitem[Wang(2024)]{wang2024causalbench}
Wang, Z.
\newblock CausalBench: A Comprehensive Benchmark for Evaluating Causal
  Reasoning Capabilities of Large Language Models.
\newblock In Proceedings of the 10th SIGHAN Workshop on
  Chinese Language Processing (SIGHAN-10), {Bangkok, Thailand, 16 August} 2024;  pp. 143--151.

\bibitem[Banerjee et~al.(2024)Banerjee, Java, Jandial, Shahid, Furniturewala,
  Krishnamurthy, and Bhatia]{banerjee2024all}
Banerjee, P.; Java, A.; Jandial, S.; Shahid, S.; Furniturewala, S.;
  Krishnamurthy, B.; Bhatia, S.
\newblock All Should Be Equal in the Eyes of LMs: Counterfactually Aware Fair
  Text Generation.
\newblock In Proceedings of the AAAI Conference on
  Artificial Intelligence, {Vancouver, BC, Canada, 20--27 February}  2024; Volume~38, pp. 17673--17681.

\bibitem[Cheng et~al.(2024)Cheng, Zouhar, Chan, F{\"u}rst, Strobelt, and
  El-Assady]{cheng2024interactive}
Cheng, F.; Zouhar, V.; Chan, R.S.M.; F{\"u}rst, D.; Strobelt, H.; El-Assady, M.
\newblock Interactive Analysis of LLMs using Meaningful Counterfactuals.
\newblock {\em arXiv } {\bf 2024}, arXiv:2405.00708.

\bibitem[Bai et~al.(2024)Bai, Zhao, Shi, Xie, Wu, and He]{bai2024fairmonitor}
Bai, Y.; Zhao, J.; Shi, J.; Xie, Z.; Wu, X.; He, L.
\newblock FairMonitor: A Dual-framework for Detecting Stereotypes and Biases in
  Large Language Models.
\newblock {\em arXiv } {\bf 2024}, arXiv:2405.03098.

\bibitem[Babonnaud et~al.(2024)Babonnaud, Delouche, and
  Lahlouh]{babonnaud2024bias}
Babonnaud, W.; Delouche, E.; Lahlouh, M.
\newblock The Bias that Lies Beneath: Qualitative Uncovering of Stereotypes in
  Large Language Models.
\newblock {\em Swed. Artif. Intell. Soc.} {\bf 2024}, 
  {195--203.} 
\newblock {\url{https://doi.org/10.3384/ecp208022}}

\bibitem[Inan et~al.(2023)Inan, Upasani, Chi, Rungta, Iyer, Mao, Tontchev, Hu,
  Fuller, Testuggine, et~al.]{inan2023llama}
Inan, H.; Upasani, K.; Chi, J.; Rungta, R.; Iyer, K.; Mao, Y.; Tontchev, M.;
  Hu, Q.; Fuller, B.; Testuggine, D.;  et~al.
\newblock Llama guard: Llm-based input-output safeguard for human-ai
  conversations.
\newblock {\em arXiv } {\bf 2023}, arXiv:2312.06674.

\bibitem[Koh et~al.(2024)Koh, Kim, Lee, and Jung]{koh2024can}
Koh, H.; Kim, D.; Lee, M.; Jung, K.
\newblock Can LLMs Recognize Toxicity? Structured Toxicity Investigation
  Framework and Semantic-Based Metric.
\newblock {\em arXiv } {\bf 2024}, arXiv:2402.06900.

\bibitem[Hu et~al.(2024)Hu, Piet, Zhao, Jiao, and Wagner]{hu2024toxicity}
Hu, Z.; Piet, J.; Zhao, G.; Jiao, J.; Wagner, D.
\newblock Toxicity Detection for Free.
\newblock {\em arXiv } {\bf 2024}, arXiv:2405.18822.

\bibitem[An et~al.(2024)An, Acquaye, Wang, Li, and Rudinger]{an2024large}
An, H.; Acquaye, C.; Wang, C.; Li, Z.; Rudinger, R.
\newblock Do Large Language Models Discriminate in Hiring Decisions on the
  Basis of Race, Ethnicity, and Gender?
\newblock {\em arXiv } {\bf 2024}, arXiv:2406.10486.

\bibitem[Scherrer et~al.(2024)Scherrer, Shi, Feder, and
  Blei]{scherrer2024evaluating}
Scherrer, N.; Shi, C.; Feder, A.; Blei, D.
\newblock Evaluating the moral beliefs encoded in llms.
\newblock {\em Adv. Neural Inf. Process. Syst.} {\bf 2024},
  {\em {36}},  51778--51809. 


\bibitem[Echterhoff et~al.(2024)Echterhoff, Liu, Alessa, McAuley, and
  He]{echterhoff2024cognitive}
Echterhoff, J.; Liu, Y.; Alessa, A.; McAuley, J.; He, Z.
\newblock Cognitive bias in high-stakes decision-making with llms.
\newblock {\em arXiv } {\bf 2024}, arXiv:2403.00811.

\bibitem[Sheng et~al.(2019)Sheng, Chang, Natarajan, and Peng]{sheng2019woman}
Sheng, E.; Chang, K.W.; Natarajan, P.; Peng, N.
\newblock The Woman Worked as a Babysitter: On Biases in Language Generation.
\newblock In Proceedings of the 2019 Conference on Empirical
  Methods in Natural Language Processing and the 9th International Joint
  Conference on Natural Language Processing (EMNLP-IJCNLP),  {Hong Kong, China, 3--7 November} 2019;  pp. 3407--3412.

\bibitem[Garimella et~al.(2021)Garimella, Amarnath, Kumar, Yalla, Anandhavelu,
  Chhaya, and Srinivasan]{garimella2021he}
Garimella, A.; Amarnath, A.; Kumar, K.; Yalla, A.P.; Anandhavelu, N.; Chhaya,
  N.; Srinivasan, B.V.
\newblock He is very intelligent, she is very beautiful? on mitigating social
  biases in language modelling and generation.
\newblock In Proceedings of the Findings of the Association for Computational
  Linguistics: ACL-IJCNLP 2021, {Online, 1--6 August} 2021;  pp. 4534--4545.

\bibitem[Garimella et~al.(2022)Garimella, Mihalcea, and
  Amarnath]{garimella2022demographic}
Garimella, A.; Mihalcea, R.; Amarnath, A.
\newblock Demographic-Aware Language Model Fine-tuning as a Bias Mitigation
  Technique.
\newblock In Proceedings of the 2nd Conference of the
  Asia-Pacific Chapter of the Association for Computational Linguistics and the
  12th International Joint Conference on Natural Language Processing (Volume 2:
  Short Papers), Online, {20--23 November}  2022; pp. 311--319.
\newblock {\url{https://doi.org/10.18653/v1/2022.aacl-short.38}}.

\bibitem[Liu et~al.(2022)Liu, Jia, Wei, Xu, and Vosoughi]{liu2022quantifying}
Liu, R.; Jia, C.; Wei, J.; Xu, G.; Vosoughi, S.
\newblock Quantifying and alleviating political bias in language models.
\newblock {\em Artif. Intell.} {\bf 2022}, {\em 304},~103654.

\bibitem[Nie et~al.(2024)Nie, Fromm, Welch, G{\"o}rge, Karimi, Plepi, Mowmita,
  Flores-Herr, Ali, and Flek]{nie-etal-2024-multilingual}
Nie, S.; Fromm, M.; Welch, C.; G{\"o}rge, R.; Karimi, A.; Plepi, J.; Mowmita,
  N.; Flores-Herr, N.; Ali, M.; Flek, L.
\newblock Do Multilingual Large Language Models Mitigate Stereotype Bias?
\newblock In Proceedings of the 2nd Workshop on
  Cross-Cultural Considerations in NLP,  Bangkok, Thailand, {16 August}  2024; pp. 65--83.
\newblock {\url{https://doi.org/10.18653/v1/2024.c3nlp-1.6}}.

\bibitem[Ferrara(2023)]{ferrara2023should}
Ferrara, E.
\newblock Should chatgpt be biased? Challenges and risks of bias in large
  language models.
\newblock {\em First Monday} {\bf 2023}, {\em {28}}. 
\newblock {\url{https://doi.org/10.5210/fm.v28i11.13346}}.

\bibitem[Zmigrod et~al.(2019)Zmigrod, Mielke, Wallach, and
  Cotterell]{zmigrod-etal-2019-counterfactual}
Zmigrod, R.; Mielke, S.J.; Wallach, H.; Cotterell, R.
\newblock Counterfactual Data Augmentation for Mitigating Gender Stereotypes in
  Languages with Rich Morphology.
\newblock In Proceedings of the 57th Annual Meeting of the
  Association for Computational Linguistics,  Florence, Italy, {28 July--2 August} 2019; pp.
  1651--1661.
\newblock {\url{https://doi.org/10.18653/v1/P19-1161}}.

\bibitem[Webster et~al.(2020)Webster, Wang, Tenney, Beutel, Pitler, Pavlick,
  Chen, Chi, and Petrov]{webster2020measuring}
Webster, K.; Wang, X.; Tenney, I.; Beutel, A.; Pitler, E.; Pavlick, E.; Chen,
  J.; Chi, E.; Petrov, S.
\newblock Measuring and reducing gendered correlations in pre-trained models.
\newblock {\em arXiv } {\bf 2020}, arXiv:2010.06032.

\bibitem[Barikeri et~al.(2021)Barikeri, Lauscher, Vuli{\'c}, and
  Glava{\v{s}}]{barikeri-etal-2021-redditbias}
Barikeri, S.; Lauscher, A.; Vuli{\'c}, I.; Glava{\v{s}}, G.
\newblock {R}eddit{B}ias: A Real-World Resource for Bias Evaluation and
  Debiasing of Conversational Language Models.
\newblock In Proceedings of the 59th Annual Meeting of the
  Association for Computational Linguistics and the 11th International Joint
  Conference on Natural Language Processing (Volume 1: Long Papers), Online, {1--6 August } 
  2021; pp. 1941--1955.
\newblock {\url{https://doi.org/10.18653/v1/2021.acl-long.151}}.

\bibitem[Mondal and Lipizzi(2024)]{mondal2024mitigating}
Mondal, D.; Lipizzi, C.
\newblock Mitigating Large Language Model Bias: Automated Dataset Augmentation
  and Prejudice Quantification.
\newblock {\em Computers} {\bf 2024}, {\em 13},~141.

\bibitem[Maudslay et~al.(2019)Maudslay, Gonen, Cotterell, and
  Teufel]{Maudslay2019ItsAI}
Maudslay, R.H.; Gonen, H.; Cotterell, R.; Teufel, S.
\newblock It`s All in the Name: Mitigating Gender Bias with Name-Based
  Counterfactual Data Substitution.
\newblock In Proceedings of the 2019 Conference on Empirical
  Methods in Natural Language Processing and the 9th International Joint
  Conference on Natural Language Processing (EMNLP-IJCNLP), Hong Kong, China, {3--7 November} 
  2019; pp. 5267--5275.
\newblock {\url{https://doi.org/10.18653/v1/D19-1530}}.

\bibitem[Meade et~al.(2022)Meade, Poole-Dayan, and Reddy]{meade_2022_empirical}
Meade, N.; Poole-Dayan, E.; Reddy, S.
\newblock An Empirical Survey of the Effectiveness of Debiasing Techniques for
  Pre-trained Language Models.
\newblock In Proceedings of the 60th Annual Meeting of the
  Association for Computational Linguistics (Volume 1: Long Papers), Dublin,
  Ireland, {22--27 May}  2022; pp. 1878--1898.
\newblock {\url{https://doi.org/10.18653/v1/2022.acl-long.132}}.

\bibitem[Serouis and S{\`e}des(2024)]{serouis2024exploring}
Serouis, I.M.; S{\`e}des, F.
\newblock Exploring large language models for bias mitigation and fairness.
\newblock In Proceedings of the 1st International Workshop on AI Governance
  (AIGOV) in Conjunction with the Thirty-Third International Joint Conference
  on Artificial Intelligence, {Jeju, Republic of Korea, 3 August}  2024.

\bibitem[Qian et~al.(2019)Qian, Muaz, Zhang, and Hyun]{qian-etal-2019-reducing}
Qian, Y.; Muaz, U.; Zhang, B.; Hyun, J.W.
\newblock Reducing Gender Bias in Word-Level Language Models with a
  Gender-Equalizing Loss Function.
\newblock In Proceedings of the 57th Annual Meeting of the
  Association for Computational Linguistics: Student Research Workshop,
  Florence, Italy,  {28 July--2 August} 2019; pp. 223--228.
\newblock {\url{https://doi.org/10.18653/v1/P19-2031}}.

\bibitem[Joniak and Aizawa(2022)]{joniak-aizawa-2022-gender}
Joniak, P.; Aizawa, A.
\newblock Gender Biases and Where to Find Them: Exploring Gender Bias in
  Pre-Trained Transformer-based Language Models Using Movement Pruning.
\newblock In Proceedings of the 4th Workshop on Gender Bias
  in Natural Language Processing (GeBNLP), Seattle, WA, USA, {15 July} 2022; pp.
  67--73.
\newblock {\url{https://doi.org/10.18653/v1/2022.gebnlp-1.6}}.

\bibitem[Zhuang et~al.(2020)Zhuang, Qi, Duan, Xi, Zhu, Zhu, Xiong, and
  He]{zhuang2020comprehensive}
Zhuang, F.; Qi, Z.; Duan, K.; Xi, D.; Zhu, Y.; Zhu, H.; Xiong, H.; He, Q.
\newblock A comprehensive survey on transfer learning.
\newblock {\em Proc. IEEE} {\bf 2020}, {\em 109},~43--76.

\bibitem[Azunre(2021)]{azunre2021transfer}
Azunre, P.
\newblock {\em Transfer Learning for Natural Language Processing}; Simon and
  Schuster: {New York, NY, USA},  2021.

\bibitem[Ge et~al.(2023)Ge, Hua, Mei, Tan, Xu, Li, Zhang,
  et~al.]{ge2024openagi}
Ge, Y.; Hua, W.; Mei, K.; Tan, J.; Xu, S.; Li, Z.; Zhang, Y. 
\newblock OpenAGI: When llm meets domain experts.
\newblock In Proceedings of the 37th International
  Conference on Neural Information Processing Systems,  {New Orleans, LA, USA, 10--16 December}  2023.

\bibitem[Liu(2024)]{liu2024unified}
Liu, S.S.
\newblock Unified Transfer Learning in High-Dimensional Linear Regression.
\newblock In Proceedings of the 27th International
  Conference on Artificial Intelligence and Statistics, PMLR, {Valencia, Spain, 2--4 May} 2024;  pp.
  1036--1044.

\bibitem[Delobelle and Berendt(2022)]{delobelle2022fairdistillation}
Delobelle, P.; Berendt, B.
\newblock Fairdistillation: Mitigating stereotyping in language models.
\newblock In Proceedings of the Joint European Conference on Machine Learning
  and Knowledge Discovery in Databases, {Grenoble, France, 19--23 September 2022}; Springer:  {Berlin/Heidelberg, Germany,} 
  2022;  pp. 638--654.

\bibitem[Ahn and Oh(2021)]{ahn-oh-2021-mitigating}
Ahn, J.; Oh, A.
\newblock Mitigating Language-Dependent Ethnic Bias in {BERT}.
\newblock In Proceedings of the 2021 Conference on Empirical
  Methods in Natural Language Processing, Online and Punta Cana, Dominican
  Republic, {7--11 November}  2021; pp. 533--549.
\newblock {\url{https://doi.org/10.18653/v1/2021.emnlp-main.42}}.

\bibitem[Levy et~al.(2023)Levy, John, Liu, Vyas, Ma, Fujinuma, Ballesteros,
  Castelli, and Roth]{levy-etal-2023-comparing}
Levy, S.; John, N.; Liu, L.; Vyas, Y.; Ma, J.; Fujinuma, Y.; Ballesteros, M.;
  Castelli, V.; Roth, D.
\newblock Comparing Biases and the Impact of Multilingual Training across
  Multiple Languages.
\newblock In Proceedings of the 2023 Conference on Empirical
  Methods in Natural Language Processing, Singapore, {6--10 December } 2023; pp. 10260--10280.
\newblock {\url{https://doi.org/10.18653/v1/2023.emnlp-main.634}}.

\bibitem[Srivastava et~al.(2014)Srivastava, Hinton, Krizhevsky, Sutskever, and
  Salakhutdinov]{srivastava2014dropout}
Srivastava, N.; Hinton, G.; Krizhevsky, A.; Sutskever, I.; Salakhutdinov, R.
\newblock Dropout: A Simple Way to Prevent Neural Networks from Overfitting.
\newblock {\em J. Mach. Learn. Res.} {\bf 2014}, {\em
  15},~1929--1958.

\bibitem[Da et~al.(2024)Da, Bossa, Berenguer, and Sahli]{da2024reducing}
Da, Y.; Bossa, M.N.; Berenguer, A.D.; Sahli, H.
\newblock Reducing Bias in Sentiment Analysis Models Through Causal Mediation
  Analysis and Targeted Counterfactual Training.
\newblock {\em IEEE Access} {\bf 2024}, {\emph{12}, 10120--10134.}

\bibitem[Cai et~al.(2024)Cai, Cao, Guo, Wen, Liu, and Chen]{cai2024locating}
Cai, Y.; Cao, D.; Guo, R.; Wen, Y.; Liu, G.; Chen, E.
\newblock Locating and mitigating gender bias in large language models.
\newblock In Proceedings of the International Conference on Intelligent
  Computing, {Tianjin, China, 5--8 August  2024}; Springer:  {Berlin/Heidelberg, Germany,}   2024;  pp. 471--482.

\bibitem[Vig et~al.(2020)Vig, Gehrmann, Belinkov, Qian, Nevo, Singer, and
  Shieber]{vig2020investigating}
Vig, J.; Gehrmann, S.; Belinkov, Y.; Qian, S.; Nevo, D.; Singer, Y.; Shieber,
  S.
\newblock Investigating gender bias in language models using causal mediation
  analysis.
\newblock {\em Adv. Neural Inf. Process. Syst.} {\bf 2020},
  {\em 33},~12388--12401.

\bibitem[Liu et~al.(2021)Liu, Jia, Wei, Xu, Wang, and
  Vosoughi]{liu2021mitigating}
Liu, R.; Jia, C.; Wei, J.; Xu, G.; Wang, L.; Vosoughi, S.
\newblock Mitigating political bias in language models through reinforced
  calibration.
\newblock In Proceedings of the AAAI Conference on
  Artificial Intelligence, {Virtual, 2--9 February} 2021; Volume~35,  pp. 14857--14866.
\newblock {\url{https://doi.org/10.1609/aaai.v35i17.17744}}.

\bibitem[Park et~al.(2018)Park, Shin, and Fung]{park-etal-2018-reducing}
Park, J.H.; Shin, J.; Fung, P.
\newblock Reducing Gender Bias in Abusive Language Detection.
\newblock In Proceedings of the 2018 Conference on Empirical
  Methods in Natural Language Processing, Brussels, Belgium, {31 October--4 November}  2018; pp.
  2799--2804.
\newblock {\url{https://doi.org/10.18653/v1/D18-1302}}.


\bibitem[Bordia and Bowman(2019)]{bordia-bowman-2019-identifying}
Bordia, S.; Bowman, S.R.
\newblock Identifying and Reducing Gender Bias in Word-Level Language Models.
\newblock In \emph{Proceedings of the 2019 Conference of the North
  {A}merican Chapter of the Association for Computational Linguistics: Student
  Research Workshop, Minneapolis, {MN, USA, 2--7 June 2019}}; Kar, S., Nadeem, F., Burdick, L., Durrett, G., Han, N.R.,
  Eds.; {Association for Computational Linguistics: Stroudsburg, PA, USA},  2019; pp. 7--15.
\newblock {\url{https://doi.org/10.18653/v1/N19-3002}}.


\bibitem[Ravfogel et~al.(2020)Ravfogel, Elazar, Gonen, Twiton, and
  Goldberg]{ravfogel-etal-2020-null}
Ravfogel, S.; Elazar, Y.; Gonen, H.; Twiton, M.; Goldberg, Y.
\newblock Null It Out: Guarding Protected Attributes by Iterative Nullspace
  Projection.
\newblock In Proceedings of the 58th Annual Meeting of the
  Association for Computational Linguistics, Online, {5--10 July} 2020; pp. 7237--7256.
\newblock {\url{https://doi.org/10.18653/v1/2020.acl-main.647}}.

\bibitem[Li et~al.(2024)Li, Tang, Liu, Spirtes, Zhang, Leqi, and
  Liu]{li2024steering}
Li, J.; Tang, Z.; Liu, X.; Spirtes, P.; Zhang, K.; Leqi, L.; Liu, Y.
\newblock Steering LLMs Towards Unbiased Responses: A Causality-Guided
  Debiasing Framework.
\newblock {\em arXiv } {\bf 2024}, arXiv:2403.08743.

\bibitem[Zhang et~al.(2024)Zhang, Zhang, Zhou, and Xu]{zhang2024causal}
Zhang, C.; Zhang, L.; Zhou, D.; Xu, G.
\newblock Causal Prompting: Debiasing Large Language Model Prompting based on
  Front-Door Adjustment.
\newblock {\em arXiv } {\bf 2024}, arXiv:2403.02738.

\bibitem[Pearl et~al.(2016)Pearl, Glymour, and Jewell]{pearl2016causal}
Pearl, J.; Glymour, M.; Jewell, N.P.
\newblock {\em Causal Inference in Statistics: A Primer}; John Wiley \& Sons:  {Hoboken, NJ, USA,} 
  2016.

\bibitem[Abid et~al.(2021)Abid, Farooqi, and Zou]{abid2021antimuslim}
Abid, A.; Farooqi, M.; Zou, J.
\newblock Persistent Anti-Muslim Bias in Large Language Models.
\newblock In Proceedings of the 2021 AAAI/ACM Conference on
  AI, Ethics, and Society, New York, NY, USA,  {19--21 May} 2021; p. 298–306.

\bibitem[Gehman et~al.(2020)Gehman, Gururangan, Sap, Choi, and
  Smith]{gehman2020realtoxicityprompts}
Gehman, S.; Gururangan, S.; Sap, M.; Choi, Y.; Smith, N.A.
\newblock {R}eal{T}oxicity{P}rompts: Evaluating Neural Toxic Degeneration in
  Language Models.
\newblock In \emph{Proceedings of the Findings of the Association for Computational
  Linguistics: EMNLP 2020}; Cohn, T., He, Y., Liu, Y., Eds.; Association for
  Computational Linguistics: {Stroudsburg, PA, USA},   2020;  pp. 3356--3369.

\bibitem[Koenecke et~al.(2020)Koenecke, Nam, Lake, Nudell, Quartey, Mengesha,
  Toups, Rickford, Jurafsky, and Goel]{koenecke2020racial}
Koenecke, A.; Nam, A.; Lake, E.; Nudell, J.; Quartey, M.; Mengesha, Z.; Toups,
  C.; Rickford, J.R.; Jurafsky, D.; Goel, S.
\newblock Racial disparities in automated speech recognition.
\newblock {\em Proc. Natl. Acad. Sci. USA} {\bf 2020},
  {\em 117},~7684--7689.

\bibitem[Saunders and Byrne(2020)]{saunders2020reducing}
Saunders, D.; Byrne, B.
\newblock Reducing Gender Bias in Neural Machine Translation as a Domain
  Adaptation Problem.
\newblock In \emph{Proceedings of the 58th Annual Meeting of the
  Association for Computational Linguistics}; Jurafsky, D., Chai, J., Schluter,
  N., Tetreault, J., Eds.; Association for Computational Linguistics: {Stroudsburg, PA, USA},   2020;  pp.
  7724--7736.

\bibitem[{EEOC}(2023)]{eeoc_guidance2023}
{EEOC}.
\newblock EEOC Guidance on AI in Employment. 2023. 
\newblock Available online: \url{https://www.eeoc.gov} {(accessed on 15 September 2024).} 


\bibitem[{New York City Council}(2021)]{nyc_local144}
{New York City Council}.
\newblock Local Law 144: Automated Employment Decision Tools.  2021.
\newblock Available online: \url{https://legistar.council.nyc.gov}  {(accessed on 15 September 2024).} 

\bibitem[{U.S. Department of Health and Human
  Services}(2024)]{section1557_2024}
{U.S. Department of Health and Human Services}.
\newblock Section 1557 Final Rule. 2024.
\newblock Available online: \url{https://www.hhs.gov}  {(accessed on 15 September 2024).}  

\bibitem[{European Commission}(2023)]{eu_ai_act}
{European Commission}.
\newblock Artificial Intelligence Act. 2023.
\newblock
  Available online: \url{https://digital-strategy.ec.europa.eu/en/policies/european-approach-artificial-intelligence}  {(accessed on).}
   

\bibitem[Zack et~al.(2024)Zack, Lehman, Suzgun, and
  Rodriguez]{zack2024assessing}
Zack, T.; Lehman, E.; Suzgun, M.; Rodriguez, J.A.
\newblock Assessing the potential of GPT-4 to perpetuate racial and gender
  biases in health care: A model evaluation study.
\newblock {\em  Lancet Digit. Health} {\bf 2024}, {\em 6},~E12--E22.

\bibitem[MacCartney(2009)]{maccartney2009natural}
MacCartney, B.
\newblock {\em Natural Language Inference}; Stanford University: {Stanford, CA, USA},  2009.

\bibitem[Dev et~al.(2020)Dev, Li, Phillips, and Srikumar]{dev2020measuring}
Dev, S.; Li, T.; Phillips, J.M.; Srikumar, V.
\newblock On measuring and mitigating biased inferences of word embeddings.
\newblock In Proceedings of the AAAI Conference on
  Artificial Intelligence, {New York, NY, USA, 7--12 February}   2020; Volume~34,  pp. 7659--7666.

\bibitem[Zadeh(2006)]{zadeh2006search}
Zadeh, L.A.
\newblock From search engines to question answering systems—The problems of
  world knowledge, relevance, deduction and precisiation. In {\em Capturing
  Intelligence}; Elsevier:  {Amsterdam, The Netherlands,} 
  2006; Volume~1,  pp. 163--210.

\bibitem[Li et~al.(2024)Li, Li, Ma, and Liu]{li2024citation}
Li, W.; Li, J.; Ma, W.; Liu, Y.
\newblock Citation-Enhanced Generation for LLM-based Chatbot.
\newblock {\em arXiv} {\bf 2024},  arXiv:2402.16063.

\bibitem[Hua et~al.(2023)Hua, Ge, Xu, Ji, and Zhang]{hua2023up5}
Hua, W.; Ge, Y.; Xu, S.; Ji, J.; Zhang, Y.
\newblock Up5: Unbiased foundation model for fairness-aware recommendation.
\newblock {\em arXiv } {\bf 2023}, arXiv:2305.12090.

\end{thebibliography}
\end{document}